\pdfoutput=1
\documentclass{article}

\PassOptionsToPackage{round}{natbib}

\usepackage[main, final]{neurips_2026}

\usepackage[utf8]{inputenc}
\usepackage[T1]{fontenc}
\usepackage{hyperref}
\usepackage{url}
\usepackage{booktabs}
\usepackage{multirow}
\usepackage{amsmath}
\usepackage{amsfonts}
\usepackage{nicefrac}
\usepackage{microtype}
\usepackage{xcolor}
\usepackage{xspace}
\usepackage{graphicx}
\graphicspath{{figures/main/}{figures/appendix/}{./}}
\usepackage{float}
\usepackage{wrapfig}
\usepackage{subcaption}
\usepackage[most]{tcolorbox}
\usepackage{listings}

\newcommand{\codecontests}{CodeContests\xspace}
\newcommand{\codecontestsplus}{CodeContestsPlus\xspace}
\newtcolorbox{promptbox}{
  enhanced,
  breakable,
  colback=gray!4,
  colframe=black!65,
  boxrule=0.5pt,
  arc=2pt,
  left=6pt,
  right=6pt,
  top=6pt,
  bottom=6pt,
}
\newtcblisting{trajbox}[1][]{%
    listing options={
        breaklines=true,
        breakautoindent=false,
        breakindent=0ex,
        basicstyle=\ttfamily\scriptsize,
        aboveskip=1pt,
        belowskip=1pt,
        escapechar=|,
        upquote=true,
    },
    width=\textwidth,
    listing only,
    autoparskip,
    colback=blue!2!white,
    colbacktitle=gray!15,
    coltitle=black,
    fonttitle=\small\bfseries,
    title={#1},
    enhanced,
    breakable,
    boxrule=0.4pt,
    arc=2pt,
    left=4pt,
    right=4pt,
    top=4pt,
    bottom=4pt,
}

\defcitealias{faircodegenteam2025cwmopenweightsllmresearch}{FAIR CodeGen team, 2025}
\defcitealias{kimiteam2026kimik2openagentic}{Kimi team, 2026}
\defcitealias{minimax2025minimaxm1scalingtesttimecompute}{Minimax team, 2025}
\defcitealias{qwen2025qwen25technicalreport}{Qwen team, 2025}

\title{
Correctness vs Efficiency: Behavior Control with Weight-Space Extrapolation in Code RL
}

\author{%
  \textbf{Kunhao Zheng}$^{1,2}$ \quad
  \textbf{Pierre Chambon}$^{1,3}$ \quad
  \textbf{Juliette Decugis}$^{1,3}$ \quad
  \textbf{Jonas Gehring}$^{1}$ \\
  \textbf{Taco Cohen}$^{1}$ \quad
  \textbf{Benjamin Negrevergne}$^{2 \ \dagger}$ \quad
  \textbf{Gabriel Synnaeve}$^{1\ \dagger}$ \\[0.4em]
  \normalfont
  $^{1}$Meta FAIR \qquad
  $^{2}$Université Paris-Dauphine PSL \qquad
  $^{3}$Inria\\[0.2em]
  $^{\dagger}$Equal supervision. \\[0.2em]
  \texttt{\{kunhao,gab\}@meta.com} \quad
  \texttt{benjamin.negrevergne@dauphine.psl.eu}
}

\begin{document}

\maketitle

\begin{abstract}

A model trained with an imperfect reward in RL may learn to exploit the reward with undesired behavior.
We study a controlled form of this problem in competitive programming. A program may pass small tests while violating the intended efficiency requirement on larger inputs.
We construct a family of rewards that progressively relax the efficiency constraint, making the reward increasingly easy to exploit, and train separate policies. This sweep reveals a \emph{correctness–efficiency frontier}. Relaxing the reward leads to less efficient programs and more resource-limit failures. Stricter rewards suppress these failures but increase correctness failures near the model's capability boundary.
This frontier is also reflected in weight space. Interpolating between the resulting models reproduces the same behavioral change. More surprisingly, extrapolation in both directions further amplifies or suppresses it beyond the trained checkpoints. Extensive experiments show that this pattern holds across pure reasoning, tool use, and agentic coding, and at both 32B and 7B scales.
This suggests a simple intervention for this form of reward hacking: use a loose reward to amplify the undesired behavior, use the weight-space difference to a strict-reward model as its \emph{steering vector}, and then extrapolate past the strict model to suppress it.
\end{abstract}

\begin{figure}[h]
    \centering
    \includegraphics[width=\linewidth]{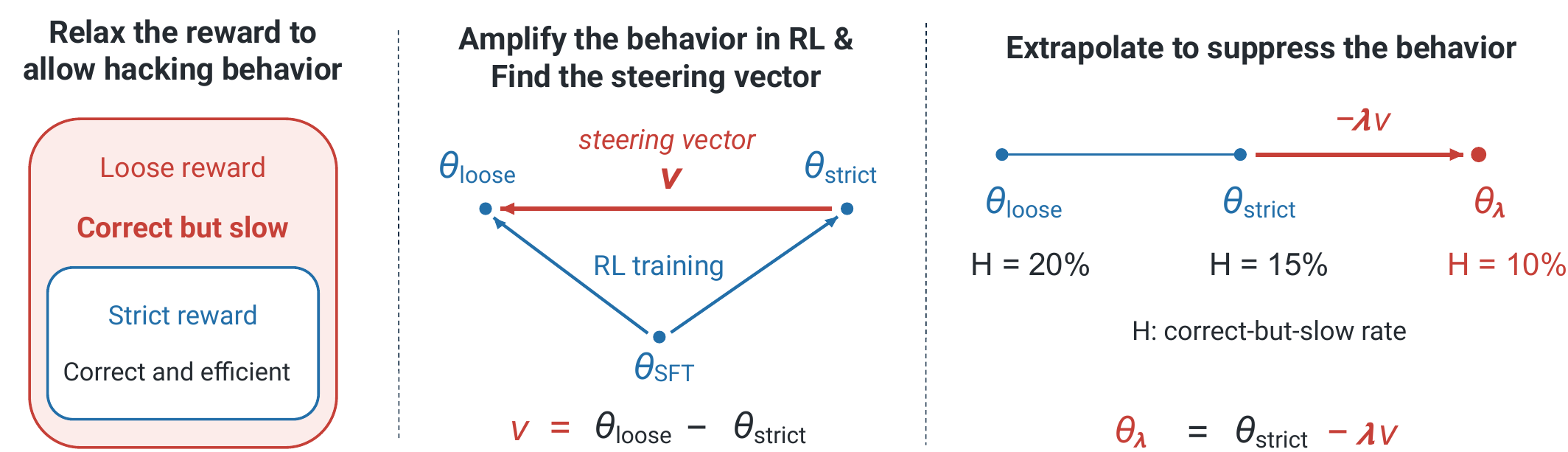}
    \caption{\textbf{Controlling reward hacking after RL.}
We use a loose reward to amplify the tendency to generate correct but slow
programs. We take the weight-space difference between the loose and strict models
as a steering vector and extrapolate past the strict model to suppress
this behavior.}
\label{fig:steering}
\end{figure}


\section{Introduction}

The reward used in reinforcement learning (RL) determines which trajectories are reinforced, but may only partially capture the intended behavior, allowing reward hacking~\citep{pan2022effects,skalse2022defining,gao2023scaling}.
With verifiable rewards, models may exploit what the verifier checks~\citep{baker2025monitoring,helff2026gaming}.
If relaxing a reward amplifies such undesired behavior, can we use the
resulting difference in model weights to suppress it after training?

Competitive programming provides a controlled setting for studying reward
exploitation, since generated programs can be tested automatically for
both correctness and efficiency.
Rewards based on only small tests can favor brute-force programs that larger stress tests reject.
We vary reward strictness through nested input-size thresholds and train
an independent policy under each reward, holding the SFT initialization
and task distribution fixed.
We take the weight-space difference between the loose and strict checkpoints,
$v = \theta_{\mathrm{loose}} - \theta_{\mathrm{strict}}$, as a \emph{steering
vector} for the tendency to generate correct but slow programs.
To test whether we can further amplify or suppress this behavior without additional
RL, we extrapolate from the both checkpoints along $v$ (Figure~\ref{fig:steering}).

\begin{figure}[t]
    \centering
    \includegraphics[width=\linewidth]{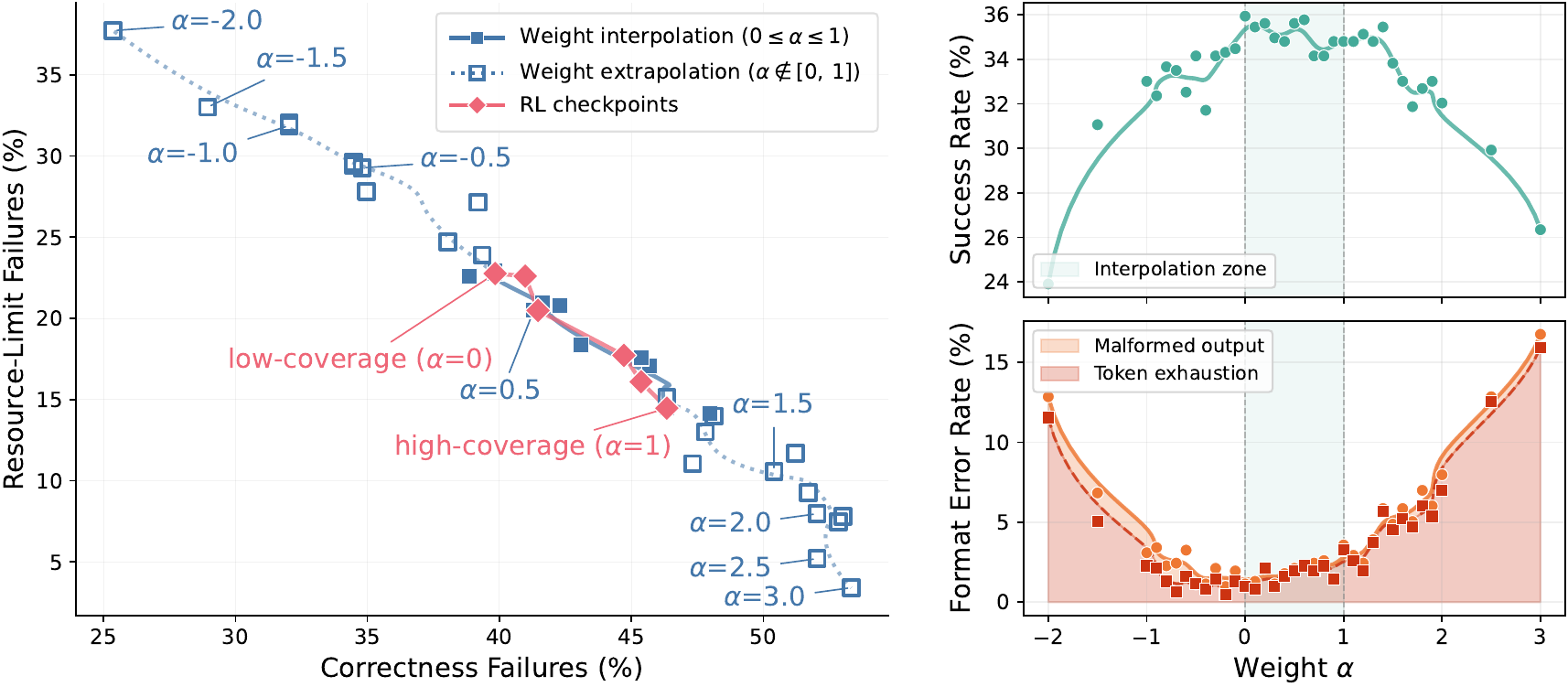}
    \caption{Linear interpolation and extrapolation between two
    RL checkpoints (trained under tool-use setting) on LCB/hard. \textbf{Left}:
    extrapolated checkpoints continue the same correctness-efficiency frontier traced by
    interpolated and independently trained RL checkpoints beyond any single RL training run produces.
    \textbf{Right}: solve rate is stable across the interpolation interval and a
    wide extrapolation regime, and degrades only at extreme $|\alpha|$ where
    format errors rise sharply through token exhaustion.}
    \label{fig:teaser}
\end{figure}

We evaluate all policies on the same full test suite.
Stricter rewards suppress resource-limit failures, but on hard problems, this comes at the cost of more correctness failures, leaving the overall solve rate nearly unchanged.
This trade-off defines a \emph{correctness-efficiency frontier}.
The frontier gives extrapolative weight averaging a concrete target: interpolation tests whether it can be recovered between trained checkpoints, while extrapolation tests whether the same behavioral change continues beyond either endpoint, further amplifying or suppressing resource-limit failures (Figure~\ref{fig:teaser}).

We organize the paper around three questions:

\begin{enumerate}
  \item \textbf{What frontier does nested unit-test coverage create?}
  Nested coverage gives a controlled axis along which stricter rewards suppress resource-limit failures but incur a correctness cost near the model's capability boundary, leaving solve rate nearly unchanged (Section~\ref{sec:results-frontier}).

  \item \textbf{Can extrapolative weight averaging go beyond trained checkpoints?}
  Interpolation between low- and high-coverage endpoints recovers the frontier traced by independently trained checkpoints, while extrapolation continues the same frontier beyond the RL endpoints.
  This continuation appears across pure reasoning, tool use, agentic coding, and the model scales we study (Sections~\ref{sec:results-interp}-\ref{sec:results-robustness}).

  \item \textbf{What are the benefits and costs of this control?}
  Extrapolation beyond the stricter-reward endpoint produces faster correct programs on commonly solved problems without requiring more tests, but stronger suppression can increase correctness failures.
  Extreme extrapolation can also introduce format errors. Moving along the frontier changes which problems are solved, providing complementary policies for inference-time scaling (Section~\ref{sec:anatomy}).
  \end{enumerate}

\section{Extrapolative Weight Averaging: Hypothesis and Setup}
\label{sec:weight-averaging}

We study linear weight averaging between two RL checkpoints $\theta_A$ and $\theta_B$,
\begin{equation}
\theta_\alpha \;=\; \alpha\,\theta_B \;+\; (1-\alpha)\,\theta_A,
\label{eq:wa}
\end{equation}
where $\alpha \in [0,1]$ gives interpolation and $\alpha \notin [0,1]$ gives extrapolation.
The first empirical check is interpolation: if this arithmetic is meaningful in our setting, interpolated checkpoints should recover sensible behavior between the trained endpoints, in line with prior results~\citep{rame2023rewarded}.
The main question is whether the same arithmetic remains meaningful outside the interval spanned by trained checkpoints.

This question is meaningful only when $\alpha$ corresponds to one controlled difference: they should share an initialization, differ along a single interpretable axis, and keep the rest of training fixed. Otherwise the averaging direction mixes several changes at once, and behavior outside the trained interval is hard to attribute to one underlying cause.

Verifier strictness in competitive programming provides such an axis. Relaxed tests can reward programs that work on smaller inputs, whereas stricter tests increasingly demand asymptotically efficient algorithms. This leads to a concrete hypothesis: RL checkpoints trained under progressively stricter verifiers may induce a correctness-efficiency frontier on hard problems. If they do, interpolation should recover that frontier and extrapolation may continue it beyond the checkpoints that RL itself produced.
We instantiate this axis as follows.

\paragraph{Training family.}
We instantiate the verifier-strictness sweep separately in three RL settings of increasing interaction capability:
\textbf{pure reasoning}, which generates code after single-turn reasoning; \textbf{tool use}, which interleaves reasoning and Python tool call in multi turns; and \textbf{agentic coding}, which uses a sandboxed terminal with file-system access.
In each setting, we train independent RL runs from a shared SFT checkpoint on \codecontestsplus~\citep{wang2025codecontestshighqualitytestcase} (1$\times$ variant), under a nested family of unit-test rewards indexed by the input-length threshold $10^k$:
\begin{align}
R_k(y) &= \mathbf{1}\!\left[y \text{ passes every test with input length} < 10^k\right],
\quad R_\infty(y) = \mathbf{1}\!\left[y \text{ passes every test}\right].
\label{eq:rk-def}
\end{align}
Lower $k$ accepts more submissions;
higher $k$ demands asymptotically efficient algorithms. We sweep
$k \in \{2, \ldots, 6, \infty\}$ to span the no-pressure to strict-pressure
endpoints, with this range chosen by the unit test distributions in \codecontestsplus (Appendix~\ref{sec:appendix-input-len}).

\paragraph{Evaluation.} We evaluate checkpoints sampled throughout training, including the final checkpoints, on the same held-out benchmark with full test coverage. The benchmark is LiveCodeBench (LCB) v5~\citep{jain2024livecodebench} (20240801-20250201), with easy, medium, and hard split.

\paragraph{Outcome taxonomy.} We sort each problem's tests by input size and stop at the first failing test. Each code submission receives one of four mutually exclusive labels: \textbf{correct} (all tests pass), \textbf{resource-limit failure} (a time or memory limit is exceeded after passing the preceding tests), \textbf{correctness failure} (a wrong answer or runtime exception within resource limits), or \textbf{format error} (unparsable output). These labels describe the observed execution outcome: resource-limit failure does not guarantee algorithmic correctness; a submission may contain a bug masked by the resource limit. To validate our design, Section~\ref{sec:results-frontier} presents a re-evaluation protocol and tests whether this ambiguity affects the observed trade-off.

\paragraph{Models.} We study at both 32B and 7B scale, in both cases we start from a pre-RL SFT checkpoint to ensure that we study the effect of RL itself rather than layering RL on an already RL-trained model. For 32B model, we use CWM-SFT~\citetext{\citetalias{faircodegenteam2025cwmopenweightsllmresearch}}. Specially for tool-use environment, we conduct two iterations of rejection sampling to teach the tool-call format and use the resulting checkpoint as the starting policy of our experiments (Appendix~\ref{sec:appendix-tool-training}). For 7B model,
since Qwen-2.5~7B~\citetext{\citetalias{qwen2025qwen25technicalreport}} does not provide a public pre-RL SFT reasoning checkpoint,
we fine-tune its base model ourselves on the public OpenCodeReasoning-2~\citep{ahmad2025opencodereasoning}
and
OpenMathReasoning~\citep{moshkov2025openmathreasoning}, giving us full control and transparency over the SFT datamix. Training details are in Appendices~\ref{sec:appendix-7b-sft-training} and~\ref{sec:appendix-tool-training}.

\section{Extrapolation Extends the Correctness-Efficiency Frontier}
\label{sec:results-extrapolation}
\label{sec:tradeoff}
\label{sec:geometry}
\label{sec:linear-interp}
\label{sec:main_frontier_result}

\begin{figure}[ht]
    \centering
    \includegraphics[width=\linewidth]{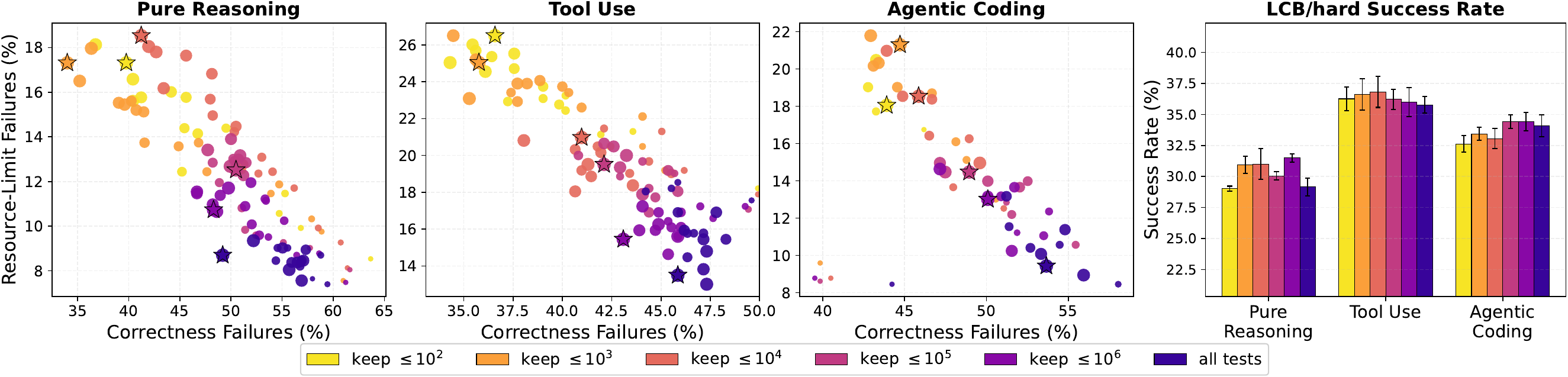}
    \caption{Correctness-efficiency frontier on LCB/hard across three
    inference settings. Each scatter panel shows checkpoints sampled throughout
    RL training under different coverage thresholds. Smaller points are earlier
    checkpoints, larger points are later checkpoints, and stars mark final
    checkpoints. In every setting, the training trajectories follow the same
    trade-off between optimization and correctness failures, defining the frontier we extrapolate along.
    \textbf{Right}: bars averaging the last four checkpoints at each threshold
    confirm that solve rate remains nearly unchanged across conditions in late
    training.}
    \label{fig:main-tradeoff}
\end{figure}

We first establish the prerequisite empirical structure for the paper: varying verifier strictness induces a frontier between resource-limit failures and correctness failures on hard problems (Section~\ref{sec:results-frontier}). We then test the hypothesis: linear
interpolation recovers this frontier (Section~\ref{sec:results-interp}), and linear
extrapolation continues it beyond what any single RL run reaches
(Section~\ref{sec:results-extrap}). We close by showing a strong generalization of both the frontier and its extension by extrapolative weight averaging across RL environment settings and model scales (Section~\ref{sec:results-robustness}).

\subsection{A Correctness-Efficiency Frontier from Verifier-Strictness Sweeps}
\label{sec:results-frontier}

\begin{figure}[ht]
    \centering
    \includegraphics[width=\linewidth]{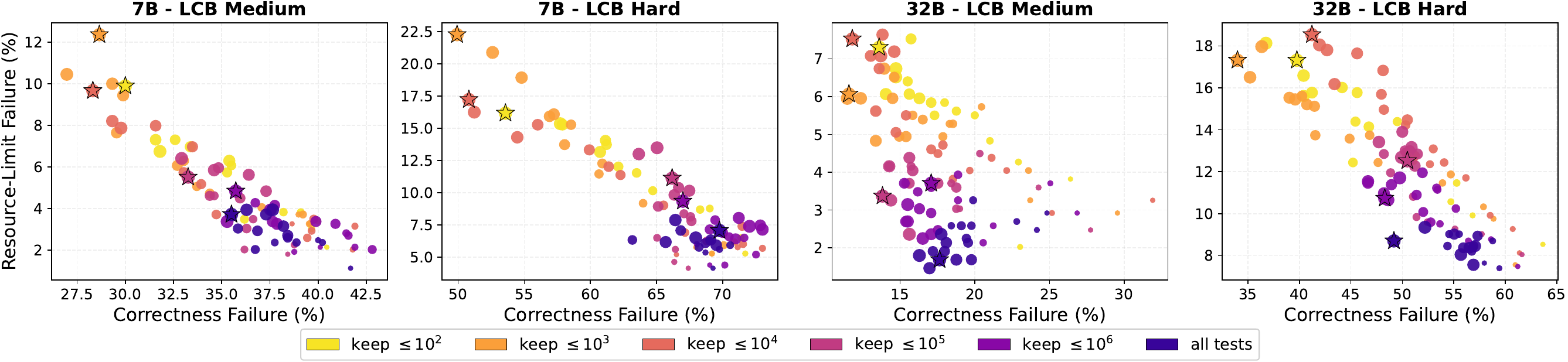}
    \caption{Pure-reasoning evaluation at 7B and 32B on medium and hard splits.
    The trade-off appears only on hard problems at 32B, but already on medium
    problems at 7B, showing that the frontier tracks model capability rather
    than a fixed difficulty label.}
    \label{fig:7b-vs-32b}
\end{figure}

On held-out LCB/hard problems, stricter verifier coverage does not steadily raise solve rate. Instead, it organizes independently trained RL checkpoints into a narrow frontier: relaxed coverage produces more
resource-limit failures, stricter coverage produces more correctness failures, and late-training solve rates remain close across thresholds. Figure~\ref{fig:main-tradeoff} shows this pattern throughout training across pure reasoning, tool use, and agentic coding. The trade-off also persists across random seeds at both model scales (Appendix~\ref{sec:appendix-seed-variance}).

Rather than attaching to one fixed benchmark label, the frontier appears wherever problems approach the model's capability limit. Figure~\ref{fig:7b-vs-32b} shows this shift across scales: at 32B the frontier is
clear on the hard split, whereas at 7B the same structure already appears on the medium split.
In both cases, the frontier emerges once problem difficulty is high enough that correctness and efficiency genuinely compete.

\begin{figure}[ht]
    \centering
    \includegraphics[width=0.9\linewidth]{%
      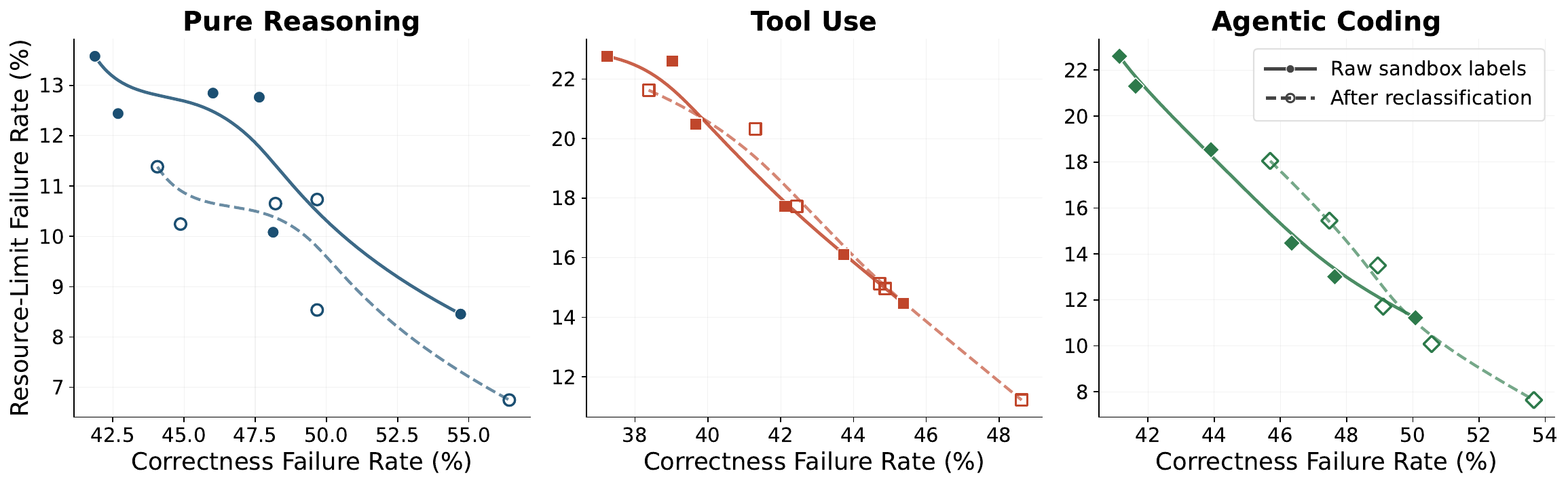}
    \caption{Trade-off curves under raw sandbox labels
      (solid) versus reclassified labels (dashed) for
      pure reasoning (left), tool use (center), and
      agentic coding (right).
      Each point is one training checkpoint under a
      fixed test-coverage threshold.
      The two labeling schemes produce nearly
      overlapping frontiers, validating the use of raw
      labels solely from execution.}
    \label{fig:reclassification-comparison}
\end{figure}

\paragraph{Error taxonomy validation with extended re-classification.} Resource-limit failures can mask correctness bug. We re-evaluate these submissions to distinguish correct but slow programs from programs with hidden bugs. With a 100$\times$ extended time budget, 16\%-40\% pass the full test suite. For the ones which still hit the resource limits, we use an LLM judge with majority voting to assess correctness and manual review the cases of disagreements (Appendix~\ref{sec:appendix-reclassification}). Across both stages, 5\% to 32\% of the original resource-limit failures are reclassified as correctness failures. The trade-off and the ordering of coverage levels persist (Figure~\ref{fig:reclassification-comparison}). Thus, the observed frontier is robust to this re-classification of the failure labels.

Taken together, these results give the rest of the section its starting geometry:
verifier strictness creates an ordered family of checkpoints along a single behavioral axis. The next two subsections show that weight averaging can follow that axis directly, first by recovering the frontier
through interpolation and then by extending it beyond the checkpoints reached by RL.

\subsection{Linear interpolation recovers the frontier}
\label{sec:results-interp}

Interpolation between a lower-coverage endpoint $\theta_{\mathrm{low}}$ and a higher-coverage endpoint $\theta_{\mathrm{high}}$, namely $\theta_\alpha = (1-\alpha)\theta_{\mathrm{low}} +
\alpha\theta_{\mathrm{high}}$ with $\alpha \in [0,1]$, recovers the frontier traced by the independently trained verifier sweep. As $\alpha$ moves from $0$ to $1$, solve rate stays nearly flat while resource-limit failures give way to correctness failures in a smooth progression (Figure~\ref{fig:teaser}). Two checkpoints therefore reconstruct the same trade-off that otherwise required a full family of RL runs. A single interpolation direction in weight space is sufficient to traverse that frontier.

Prior work shows that interpolation can navigate a multi-objective trade-off between checkpoints trained under different rewards~\citep{rame2023rewarded}.
In our case, the trade-off is not designed by specifying separate objectives since the rewards are nested and monotone improvement is plausible in theory. Empirically, the verifier-strictness sweep traces out a frontier across separate RL runs, and interpolation recovers that same frontier.

\subsection{Linear extrapolation extends the frontier beyond what RL reaches}
\label{sec:results-extrap}

\begin{figure}[ht]
    \centering
    \includegraphics[width=0.8\linewidth]{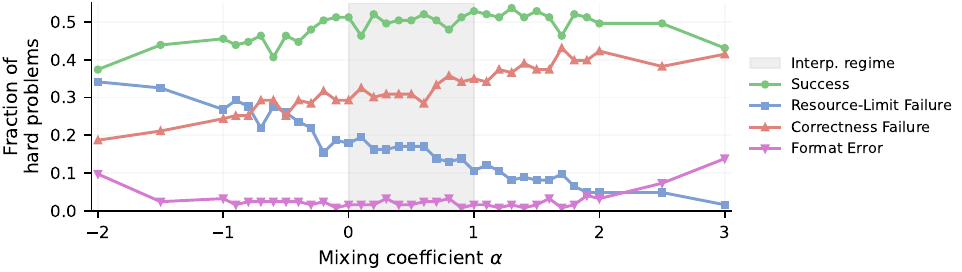}
    \caption{Outcome composition along the interpolation and extrapolation axis on LCB/hard for RL checkpoints (trained under tool-use setting). The interpolation coefficient $\alpha$ moves from the lower-coverage endpoint $\theta_{\mathrm{low}}$ at $\alpha = 0$ to the higher-coverage endpoint $\theta_{\mathrm{high}}$ at $\alpha = 1$. Resource-limit failures fall as correctness failures rise, while solve rate remains nearly flat across the interpolation regime and a broad extrapolation range. Format errors become substantial only at extreme $|\alpha|$. The grey band marks the interpolation interval $[0,1]$.}
    \label{fig:outcome-composition}
\end{figure}

\begin{figure}[ht]
    \centering
    \includegraphics[width=0.9\textwidth]{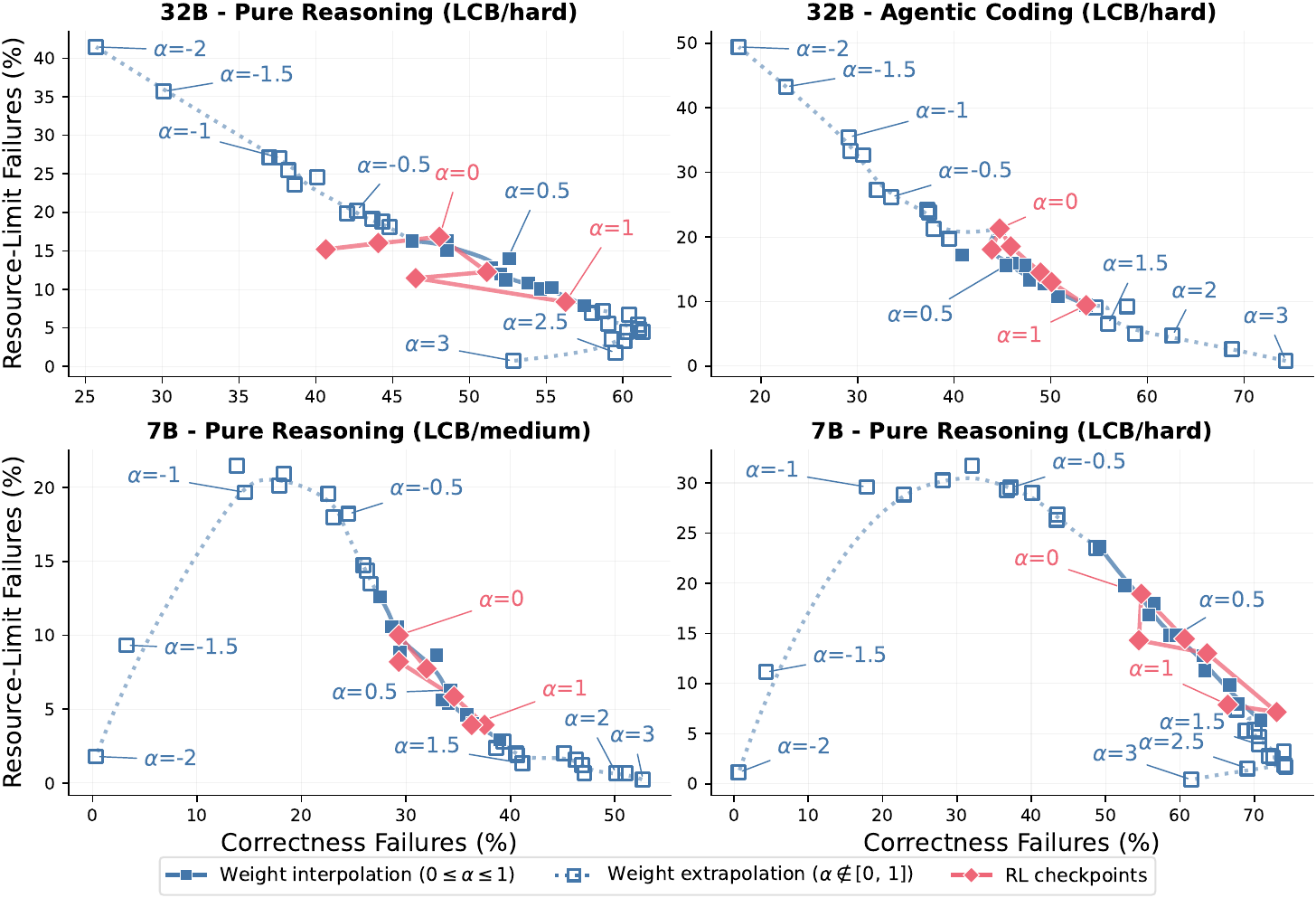}
    \caption{Extrapolative continuation across inference settings and model scale. Figure~\ref{fig:teaser}
    provides the 32B tool-use case. We show the two remaining 32B settings on LCB/hard, pure reasoning and agentic coding, and 7B pure reasoning on LCB/medium and LCB/hard.
    In each plot, interpolation recovers the correctness-efficiency frontier inside the trained interval, and extrapolation continues the same trade-off outside it, while the stable extrapolation range is narrower for 7B.}
    \label{fig:extrapolation-strip}
\end{figure}

Extrapolation continues the same frontier beyond the checkpoints reached by RL (Figure~\ref{fig:teaser}). For $\alpha > 1$, the extrapolated checkpoints move past the higher-coverage endpoint towards fewer resource-limit failures and more correctness failures than RL itself reaches. For $\alpha < 0$, they continue the same trade-off in the opposite direction beyond the lower-coverage endpoint.
Extrapolative weight averaging therefore does more than recover the verifier-strictness sweep. It extends that sweep to new checkpoints that no single RL run produces.

This continuation remains stable over a broad range of extrapolation coefficients. For $\alpha \in [-0.5, 1.5]$, solve rate stays nearly flat even though this interval already extends beyond the two trained endpoints, while resource-limit failures fall and correctness failures rise along the same pattern seen inside the trained interval (Figure~\ref{fig:outcome-composition}). At more extreme coefficients, the same pattern
remains visible, but solve rate drops because format errors rise sharply, mostly through token exhaustion. Extrapolative weight averaging therefore extends the verifier-strictness sweep beyond the range reached by RL, until model generations collapse.

\subsection{Extrapolative weight averaging generalizes across inference settings and model scales}
\label{sec:results-robustness}

Figure~\ref{fig:extrapolation-strip} extends the 32B tool-use case in Figure~\ref{fig:teaser} to the two remaining 32B inference settings. In both pure-reasoning and agentic coding, interpolation recovers the correctness-efficiency frontier inside the trained interval, and extrapolation continues the same frontier beyond it.
Agentic coding shows the cleanest continuation. Pure reasoning follows the same trend, but with visibly more scatter and earlier degradation at extreme coefficients.

The scale comparison follows the same capability-relative pattern as Section~\ref{sec:results-frontier}.
For the 7B model, Figure~\ref{fig:extrapolation-strip} shows both LCB/medium and LCB/hard, which shows the same frontier and extrapolative continuation.
The stable range for extrapolation is narrower at 7B, especially on the low-$\alpha$ side where format errors dominate once $\alpha < -1$ (Appendix~\ref{app:7b-hard-extrapolation}). Extrapolative weight averaging therefore generalizes across inference settings and across model scale once problem difficulty is matched to model
capability, while the width of the stable regime is setting dependent.

\section{Analysis and Perspectives}
\label{sec:implications}
\label{sec:breaking}

We next examine how extrapolation changes the programs the model generates. We first measure the runtime of correct programs and show that it makes them faster (Section~\ref{sec:runtime}). We also show that moving along the axis changes which hard problems are solved, which makes ensembling extrapolated checkpoints useful for inference-time scaling (Section~\ref{sec:anatomy}).

\subsection{Extrapolation makes correct programs faster}
\label{sec:runtime}

We next ask whether extrapolation also produces faster correct programs. We compare the model trained with the stricter reward against the model obtained by extrapolating beyond it ($\alpha = 2$). For each problem solved by both models, we measure the runtime on the 5 largest test cases, and take the median of multiple repetition after warmup.

\begin{figure}[ht]
    \centering
    \includegraphics[width=0.8\linewidth]{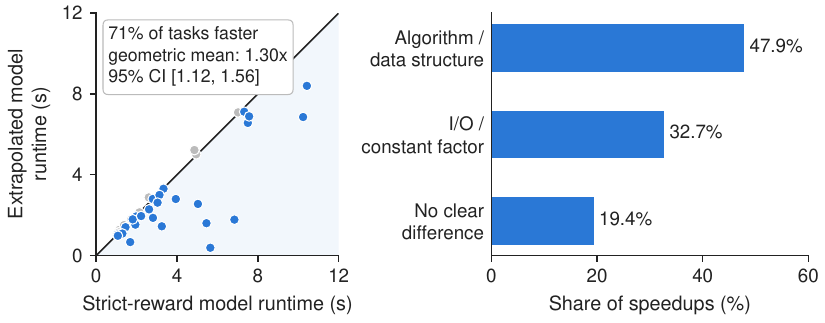}
    \caption{
\textbf{Left:} Correct-program runtime on commonly solved LCB/v5 problems where the strict-reward model produce solutions whose runtime >1s. Points below the diagonal mean that extrapolated model ($\alpha=2$) produces faster solutions.
\textbf{Right:} Code differences identified by an LLM judge among pairwise runtime speedups for the extrapolated model.}
\label{fig:speedup}
\end{figure}

Across all 141 problems solved by both models on LCB/v5 medium and hard split, extrapolation yields a $1.14\times$ geometric-mean speedup (95\% CI $[1.04, 1.27]$) compared to the strict-reward model. The gains are larger on problems in which solutions take longer time to run: for these problems where strict-reward model's solutions take more than 1s to finish, extrapolation is faster on 71\% of them and yields a $1.30\times$ speedup (Figure~\ref{fig:speedup} \textbf{Left}).

To examine what changes in the code, we compare pairs of correct programs from problems where extrapolation lowers runtime meaningfully (relative runtime delta > 5\%). We use LLM judge to assess the paired solutions. The judge attributes 47.9\% of improvement to algorithm or data-structure improvements and 32.7\% to I/O or constant-factor improvements (Figure~\ref{fig:speedup} \textbf{Right}).

\subsection{Per-problem turnover and inference-time ensembles}
\label{sec:anatomy}
\label{sec:ensembles}

\begin{figure}[ht]
    \centering
    \includegraphics[width=0.9\linewidth]{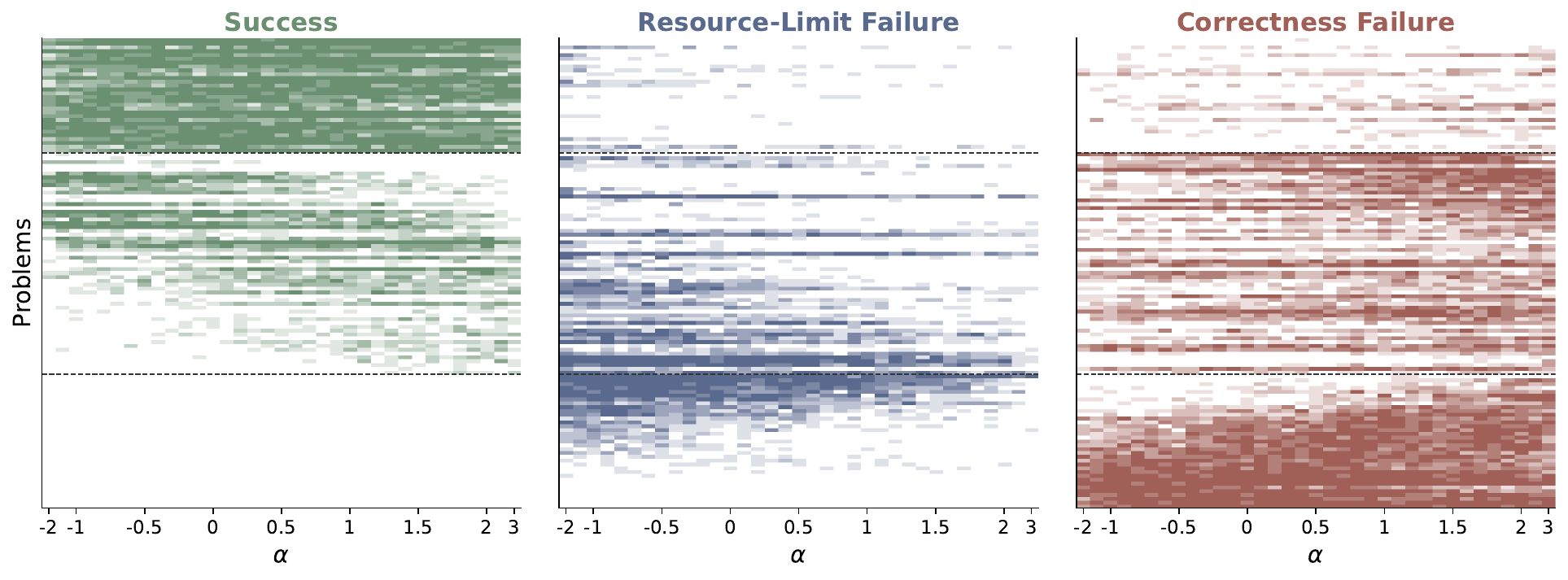}
    \caption{
    Per-problem outcomes across weight-averaging coefficients $\alpha$, measured with 10 samples per problem and sorted into always-solved (top), partially solved (middle), and never-solved (bottom). The middle group (47.2\%) accounts for the changes in which problems are solved. Failure types also continue to trade off among never-solved problems throughout the sweep.}
    \label{fig:anatomy}
\end{figure}

\begin{wrapfigure}{r}{0.4\linewidth}
    \vspace{-1\baselineskip}
    \centering
    \includegraphics[width=\linewidth]{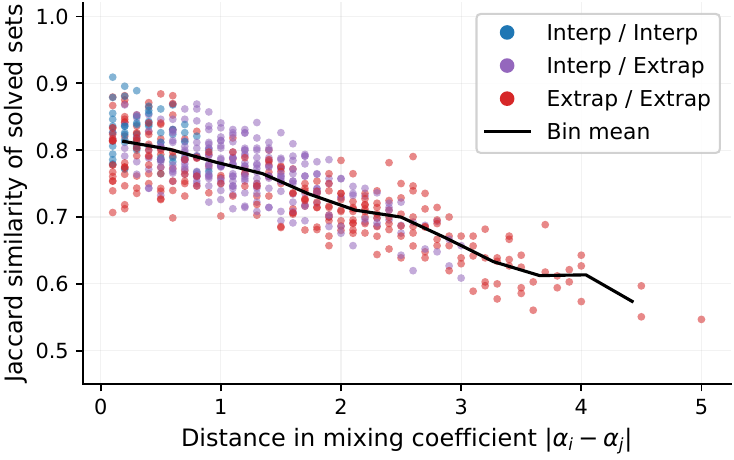}
    \caption{Jaccard similarity of solved sets as a function of $|\alpha_i - \alpha_j|$. Overlap falls steadily with distance in $\alpha$ for interpolation, extrapolation, and mixed pairs.}
    \label{fig:jaccard-alpha}
\end{wrapfigure}

The nearly flat solve rate along the extrapolation axis conceals substantial per-problem turnover. On LCB/hard, $24.4\%$ of problems are solved at every $\alpha$, $28.5\%$ are never solved, and the remaining $47.2\%$ are solved only on part of the axis (Figure~\ref{fig:anatomy}).  This middle band is the source of the solved-set diversity that ensembles can exploit. Figure~\ref{fig:jaccard-alpha} shows that the solve-set turnover is gradual: adjacent values of $\alpha$ share more than $80\%$ of solved problems, and the overlap falls steadily to about $55\%$ at the extremes.

\begin{wrapfigure}{r}{0.4\linewidth}
    \vspace{-1\baselineskip}
    \centering
    \includegraphics[width=\linewidth]{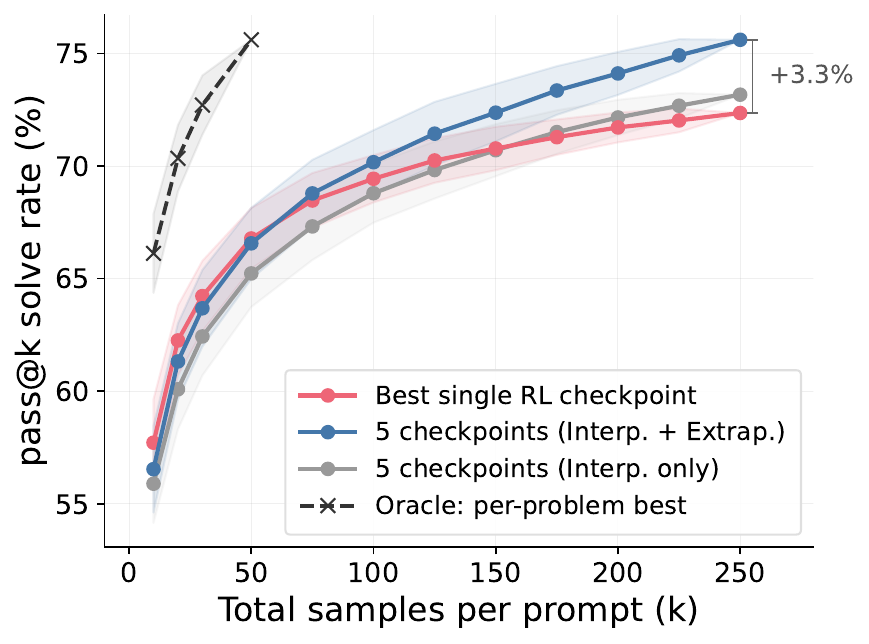}
    \caption{Matched-budget LCB/hard pass@$k$ of the best single checkpoint, the best interpolation-and-extrapolation ensemble, and the best interpolation-only ensemble.}
    \label{fig:ensemble-diversity}
    \vspace{-2\baselineskip}
\end{wrapfigure}

This turnover gives the frontier axis a direct use at inference time: we can spread a fixed sampling budget across checkpoints along the axis and aggregate their samples. We compare the best single checkpoint with five-checkpoint pools drawn from either the interpolation only or the larger interpolation-plus-extrapolation range.
At the same total sampling budget, the selected pool that includes
extrapolated checkpoints improves pass@250 by $3.3\%$ on LCB/v5 over
the best single checkpoint (Figure~\ref{fig:ensemble-diversity}). Since the pool is selected and evaluated on the same split, we treat this gain as an oracle-selected reference.

We apply public-test filtering~\citep{li2022codecontests,
gehring2025rlefgroundingcodellms, zheng2025makeslargelanguagemodels}
and report pass~$n$@$k$, the success rate when selecting up to $n$ submissions from $k$ candidates. Public-test filtering can carry part of the large-candidate-budget diversity gain to smaller submission budgets (Appendix~\ref{app:v5-filtered-pass-n}).

\begin{figure}[t]
    \centering
    \includegraphics[width=\linewidth]{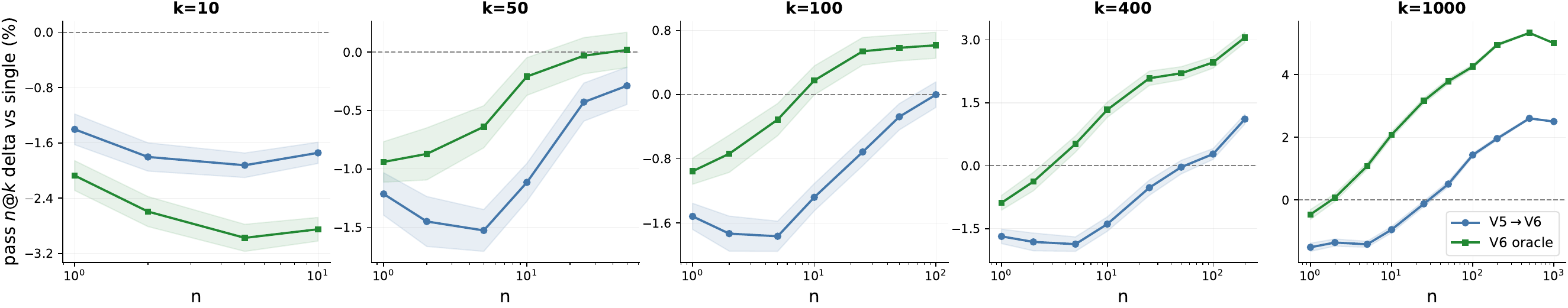}
    \caption{Transfer of ensemble selection from LCB/v5 to LCB/v6 on the hard split.
    Pools are selected on LCB/v5 (blue) or LCB/v6 (oracle, green). Both are evaluated on LCB/v6, with pass~$n$@$k$ gains measured relative to the best single checkpoint selected on LCB/v6.
    }
    \label{fig:transfer}
\end{figure}

A natural concern is that checkpoint-pool selection is tuned to LCB/v5. To test transfer, for each candidate budget $k$, we search for the best five-checkpoint pool on LCB/v5 and evaluate it on the newer LCB/v6 (20250201-20250501) hard split (Figure~\ref{fig:transfer}). The transferred pool becomes beneficial once both $k$ and $n$ are large enough. A pool selected directly on LCB/v6 still performs better, so extrapolation provides transferable diversity while better pool selection leads to potentially better performance.

\section{Related Work}

\paragraph{RL for code reasoning and test quality.}
Reinforcement learning with verifiable rewards has become a standard post-training recipe for code reasoning, from single-turn generation to tool-integrated and agentic
settings~\citetext{%
\citealp{Guo_2025,liu2025drgrpo,xue2025simpletirendtoendreinforcementlearning};
  \citetalias{minimax2025minimaxm1scalingtesttimecompute};
  \citetalias{faircodegenteam2025cwmopenweightsllmresearch};
  \citetalias{kimiteam2026kimik2openagentic}%
}.
Benchmarks such as \codecontests~\citep{li2022codecontests}, APPS~\citep{hendrycks2021measuringcodingchallengecompetence}, and TACO~\citep{li2023tacotopicsalgorithmiccode} established code generation under executable tests as a central evaluation setting.
Later work sharpened the role of test quality, including EvalPlus~\citep{liu2023codegeneratedchatgptreally}, LiveCodeBench~\citep{jain2024livecodebench}, and \codecontestsplus~\citep{wang2025codecontestshighqualitytestcase}.
At inference time, a parallel line studies how additional inference compute improves code generation through larger candidate pools, public-test filtering, and iterative candidate refinement~\citep{li2022codecontests,
zheng2025makeslargelanguagemodels, li2025stesttimescaling}.
Other methods use simulated execution for self-fixing~\citep{maimon2026selfexecutionsimulationimprovescoding} or trajectory reuse in agentic coding~\citep{kim2026scalingtesttimecomputeagentic}.
During training, other work moves beyond a single outcome reward by adding richer execution-level or process-level signals, through execution feedback~\citep{gehring2025rlefgroundingcodellms}, process-level rewards~\citep{li-etal-2025-codeprm, dai2025processsupervisionguidedpolicyoptimization}, and self-verification with generated tests~\citep{jin2025revealselfevolvingcodeagents}.

\paragraph{Linear mode connectivity and model merging.}
Mode-connectivity results identify low-loss paths between independently
trained networks~\citep{garipov2018loss}, while stochastic weight
averaging combines checkpoints along a single optimization trajectory
to improve generalization~\citep{izmailov2018averaging}.
In RL, \citet{nikishin2018improving} use the same idea to stabilize
training and reduce policy forgetting.
\citet{frankle2020linear} show that vision networks trained from
the same initialization can often be connected by low-loss linear paths.
Related approaches include model souping~\citep{wortsman2022model}, task vector editing~\citep{ilharco2023editing}, and layer-wise variants~\citep{yang2024adamergingadaptivemodelmerging, wang2024localizingtaskinformationimproved, lin2024mitigatingalignmenttaxrlhf}.
Recently, Neural Thickets~\citep{gan2026neuralthicketsdiversetask} provides a complementary view of weight-space diversity through random perturbation and ensembling.
Closest to our interpolation result is Rewarded Soups~\citep{rame2023rewarded}, which shows that interpolating checkpoints trained under explicitly different rewards traces a multi-objective Pareto front.
Follow-up work extends this idea to reward models and policy models~\citep{rame2024warmbenefitsweightaveraged, rame2024warpbenefitsweightaveraged}.
Our setting differs in that the checkpoints are trained under a nested family of stricter and looser verification regimes within the same binary pass-fail signal, rather than under explicitly different reward objectives.

\paragraph{Weight-space directions and extrapolation.}
Task arithmetic treats differences in model weights as editing directions~\citep{ilharco2023editing}.
Activation-steering methods instead construct directions in hidden
representations and intervene during inference~\citep{rimsky2024steering}.
For extrapolation, the nearest precedents are ExPO~\citep{zheng2025modelextrapolationexpeditesalignment}
and AlphaRL~\citep{cai2026predictabilityreinforcementlearningdynamics}, which extrapolate along a single post-training trajectory
(SFT-to-DPO or early-to-late RL).
In our setting, by contrast, the extrapolation is \emph{between two RL checkpoints}, and it continues the same frontier beyond what either training run reaches.

\paragraph{Correctness and efficiency in code.}
Recent work increasingly treats efficiency as an essential dimension of code quality alongside correctness.
Mercury~\citep{du2024mercurycodeefficiencybenchmark} introduces
efficiency-stratified evaluation to distinguish correct-but-slow solutions from efficient ones.
ECCO~\citep{waghjale2024eccoimprovemodelgeneratedcode} studies the tension between improving efficiency and preserving correctness in model-generated code.
Related benchmarks extend this concern to algorithmic efficiency~\citep{huang2025effibenchbenchmarkingefficiencyautomatically, chambon2025bigobenchllmsgenerate}, GPU kernels, and software engineering workloads~\citep{ ouyang2025kernelbench, ma2025swefficiencylanguagemodelsoptimize}.
Afterburner~\citep{du2025afterburnerreinforcementlearningfacilitates} uses a self-improving loop with RL to improve code efficiency. These works study settings where efficiency is made explicit, either in the benchmark or in the objective, leaving open what standard unit-test RL does when correctness and efficiency are folded into the same pass-fail supervision.

\section{Discussion and Limitations}
\label{sec:discussion}

\paragraph{Nested rewards need not yield nested policies.}
In our setting, the rewards are nested by construction.
However, empirically, the learned policies during on-policy RL are not ordered by dominance.
Their weight difference provides a direction for controlling behavior.
Even when the rewards differ along a controlled axis, the steering direction need not affect only one behavior. In our case, it changes both efficiency and correctness.
Appendices~\ref{app:reward-shaping} and~\ref{app:failure-conditioned} test whether reward shaping and failure-conditioned online fixing can shift this frontier during training.

\paragraph{Finer-grained weight-space analysis of behavioral control.}
Appendix~\ref{app:weight-geometry} shows that the weight-matrix differences are high-rank, with squared weight changes distributed nearly uniformly across most layers.
However, layer stitching (Appendix~\ref{app:layer-stitching}) suggests that the behavioral effects are concentrated in the middle layers. In our experiments, swapping the middle half of the layers reproduces the full shift of behavior (Figure~\ref{fig:block-swap}). This motivates testing whether constraining RL updates, for example with LoRA or frozen layers, can produce a more localized direction.

\paragraph{Beyond a single verifier frontier.}
As preliminary evidence, Appendix~\ref{app:auxiliary-frontier} averages two checkpoints optimized for different sampling behavior: one with stronger pass@1 and one with stronger pass@$k$. Interpolation traces a smooth Precision-Recall curve, and yields ensembles that outperform single-checkpoint sampling at larger budgets.
Extrapolation adds only modest gains in this setting.
A useful interpolation direction therefore does not necessarily support equally useful extrapolation.
For other behaviors, the key question is how to design rewards that bring out the behavior we want to control while limiting changes in other capabilities.

\paragraph{Limitations.}
Our experiments are confined to competitive programming domain, but they are not confined to a single RL environment. Our experiments span from single-turn generation to richer agentic training, suggesting that the effect is tied to verifier-induced optimization pressure rather than to one specific inference setting.
Applying the same analysis to other domains such as SWE tasks and kernel generation requires first identifying a comparable ordered axis, such as test-suite strength or performance tolerance.
The input-length thresholds are a coarse proxy for test-case scale and the optimization stress it induces, while our experiments show that it is a reliable proxy that consistently exposes the trade-off.
Our ensemble comparisons match code generation count rather than token or wall-clock cost.

\section{Conclusion}

Our results show that reward design can be used to construct weight-space directions for behavior control. In code RL, two runs from a shared initialization under different
reward strictness provide a direction that can amplify or suppress inefficient coding behavior after RL training. Extending this approach to other behaviors requires designing rewards that reliably induce the behavioral difference we want to control. The core challenge is to make that difference pronounced enough to support weight-space control while limiting changes in other capabilities.

\section*{Acknowledgement}
We especially thank Loïc Cabannes for his “Cool! Then what about extrapolation?” question, a question we could not resist testing; Matteo Pagliardini for his “Hey, I have an interpolation script!” moment and for pointing us to the task-vector literature; and Fabian Gloeckle for his enthusiastic “Yes, the result is so cool, I like this!” when we first showed him our interpolation and extrapolation results. We thank Ori Yoran, Gautier Izacard, Keyur Muzumdar, David Zhang, Rémi Munos, François Fleuret, Wassim Bouaziz, Krunoslav Lehman Pavasovic, Theophane Vallaeys and João Maria Janeiro for many fruitful discussions. We also thank all members of the FAIR CodeGen team for their support throughout the project.

\newpage
\bibliographystyle{plainnat}
\bibliography{references.bib}

@misc{wang2025codecontestshighqualitytestcase,
      title={CodeContests+: High-Quality Test Case Generation for Competitive Programming}, 
      author={Zihan Wang and Siyao Liu and Yang Sun and Hongyan Li and Kai Shen},
      year={2025},
      eprint={2506.05817},
      archivePrefix={arXiv},
      primaryClass={cs.SE},
      url={https://arxiv.org/abs/2506.05817}, 
}

@article{li2022codecontests,
   title={Competition-level code generation with AlphaCode},
   volume={378},
   ISSN={1095-9203},
   url={http://dx.doi.org/10.1126/science.abq1158},
   DOI={10.1126/science.abq1158},
   number={6624},
   journal={Science},
   publisher={American Association for the Advancement of Science (AAAS)},
   author={Li, Yujia and Choi, David and Chung, Junyoung and Kushman, Nate and Schrittwieser, Julian and Leblond, Rémi and Eccles, Tom and Keeling, James and Gimeno, Felix and Dal Lago, Agustin and Hubert, Thomas and Choy, Peter and de Masson d’Autume, Cyprien and Babuschkin, Igor and Chen, Xinyun and Huang, Po-Sen and Welbl, Johannes and Gowal, Sven and Cherepanov, Alexey and Molloy, James and Mankowitz, Daniel J. and Sutherland Robson, Esme and Kohli, Pushmeet and de Freitas, Nando and Kavukcuoglu, Koray and Vinyals, Oriol},
   year={2022},
   month=Dec, pages={1092–1097} }

@article{Guo_2025,
   title={DeepSeek-R1 incentivizes reasoning in LLMs through reinforcement learning},
   volume={645},
   ISSN={1476-4687},
   url={http://dx.doi.org/10.1038/s41586-025-09422-z},
   DOI={10.1038/s41586-025-09422-z},
   number={8081},
   journal={Nature},
   publisher={Springer Science and Business Media LLC},
   author={Guo, Daya and Yang, Dejian and Zhang, Haowei and Song, Junxiao and Wang, Peiyi and Zhu, Qihao and Xu, Runxin and Zhang, Ruoyu and Ma, Shirong and Bi, Xiao and Zhang, Xiaokang and Yu, Xingkai and Wu, Yu and Wu, Z. F. and Gou, Zhibin and Shao, Zhihong and Li, Zhuoshu and Gao, Ziyi and Liu, Aixin and Xue, Bing and Wang, Bingxuan and Wu, Bochao and Feng, Bei and Lu, Chengda and Zhao, Chenggang and Deng, Chengqi and Ruan, Chong and Dai, Damai and Chen, Deli and Ji, Dongjie and Li, Erhang and Lin, Fangyun and Dai, Fucong and Luo, Fuli and Hao, Guangbo and Chen, Guanting and Li, Guowei and Zhang, H. and Xu, Hanwei and Ding, Honghui and Gao, Huazuo and Qu, Hui and Li, Hui and Guo, Jianzhong and Li, Jiashi and Chen, Jingchang and Yuan, Jingyang and Tu, Jinhao and Qiu, Junjie and Li, Junlong and Cai, J. L. and Ni, Jiaqi and Liang, Jian and Chen, Jin and Dong, Kai and Hu, Kai and You, Kaichao and Gao, Kaige and Guan, Kang and Huang, Kexin and Yu, Kuai and Wang, Lean and Zhang, Lecong and Zhao, Liang and Wang, Litong and Zhang, Liyue and Xu, Lei and Xia, Leyi and Zhang, Mingchuan and Zhang, Minghua and Tang, Minghui and Zhou, Mingxu and Li, Meng and Wang, Miaojun and Li, Mingming and Tian, Ning and Huang, Panpan and Zhang, Peng and Wang, Qiancheng and Chen, Qinyu and Du, Qiushi and Ge, Ruiqi and Zhang, Ruisong and Pan, Ruizhe and Wang, Runji and Chen, R. J. and Jin, R. L. and Chen, Ruyi and Lu, Shanghao and Zhou, Shangyan and Chen, Shanhuang and Ye, Shengfeng and Wang, Shiyu and Yu, Shuiping and Zhou, Shunfeng and Pan, Shuting and Li, S. S. and Zhou, Shuang and Wu, Shaoqing and Yun, Tao and Pei, Tian and Sun, Tianyu and Wang, T. and Zeng, Wangding and Liu, Wen and Liang, Wenfeng and Gao, Wenjun and Yu, Wenqin and Zhang, Wentao and Xiao, W. L. and An, Wei and Liu, Xiaodong and Wang, Xiaohan and Chen, Xiaokang and Nie, Xiaotao and Cheng, Xin and Liu, Xin and Xie, Xin and Liu, Xingchao and Yang, Xinyu and Li, Xinyuan and Su, Xuecheng and Lin, Xuheng and Li, X. Q. and Jin, Xiangyue and Shen, Xiaojin and Chen, Xiaosha and Sun, Xiaowen and Wang, Xiaoxiang and Song, Xinnan and Zhou, Xinyi and Wang, Xianzu and Shan, Xinxia and Li, Y. K. and Wang, Y. Q. and Wei, Y. X. and Zhang, Yang and Xu, Yanhong and Li, Yao and Zhao, Yao and Sun, Yaofeng and Wang, Yaohui and Yu, Yi and Zhang, Yichao and Shi, Yifan and Xiong, Yiliang and He, Ying and Piao, Yishi and Wang, Yisong and Tan, Yixuan and Ma, Yiyang and Liu, Yiyuan and Guo, Yongqiang and Ou, Yuan and Wang, Yuduan and Gong, Yue and Zou, Yuheng and He, Yujia and Xiong, Yunfan and Luo, Yuxiang and You, Yuxiang and Liu, Yuxuan and Zhou, Yuyang and Zhu, Y. X. and Huang, Yanping and Li, Yaohui and Zheng, Yi and Zhu, Yuchen and Ma, Yunxian and Tang, Ying and Zha, Yukun and Yan, Yuting and Ren, Z. Z. and Ren, Zehui and Sha, Zhangli and Fu, Zhe and Xu, Zhean and Xie, Zhenda and Zhang, Zhengyan and Hao, Zhewen and Ma, Zhicheng and Yan, Zhigang and Wu, Zhiyu and Gu, Zihui and Zhu, Zijia and Liu, Zijun and Li, Zilin and Xie, Ziwei and Song, Ziyang and Pan, Zizheng and Huang, Zhen and Xu, Zhipeng and Zhang, Zhongyu and Zhang, Zhen},
   year={2025},
   month={Sept}, pages={633–638} }

@misc{qwen2025qwen25technicalreport,
      title={Qwen2.5 Technical Report}, 
      author={Qwen and : and An Yang and Baosong Yang and Beichen Zhang and Binyuan Hui and Bo Zheng and Bowen Yu and Chengyuan Li and Dayiheng Liu and Fei Huang and Haoran Wei and Huan Lin and Jian Yang and Jianhong Tu and Jianwei Zhang and Jianxin Yang and Jiaxi Yang and Jingren Zhou and Junyang Lin and Kai Dang and Keming Lu and Keqin Bao and Kexin Yang and Le Yu and Mei Li and Mingfeng Xue and Pei Zhang and Qin Zhu and Rui Men and Runji Lin and Tianhao Li and Tianyi Tang and Tingyu Xia and Xingzhang Ren and Xuancheng Ren and Yang Fan and Yang Su and Yichang Zhang and Yu Wan and Yuqiong Liu and Zeyu Cui and Zhenru Zhang and Zihan Qiu},
      year={2025},
      eprint={2412.15115},
      archivePrefix={arXiv},
      primaryClass={cs.CL},
      url={https://arxiv.org/abs/2412.15115}, 
}

@inproceedings{frankle2020linear,
  author       = {Jonathan Frankle and
                  Gintare Karolina Dziugaite and
                  Daniel M. Roy and
                  Michael Carbin},
  title        = {Linear Mode Connectivity and the Lottery Ticket Hypothesis},
  booktitle    = {Proceedings of the 37th International Conference on Machine Learning,
                  {ICML} 2020, 13-18 July 2020, Virtual Event},
  series       = {Proceedings of Machine Learning Research},
  pages        = {3259--3269},
  publisher    = {{PMLR}},
  year         = {2020},
  url          = {http://proceedings.mlr.press/v119/frankle20a.html},
  bibsource    = {dblp computer science bibliography, https://dblp.org}
}

@inproceedings{wortsman2022model,
  author       = {Mitchell Wortsman and
                  Gabriel Ilharco and
                  Samir Yitzhak Gadre and
                  Rebecca Roelofs and
                  Raphael Gontijo Lopes and
                  Ari S. Morcos and
                  Hongseok Namkoong and
                  Ali Farhadi and
                  Yair Carmon and
                  Simon Kornblith and
                  Ludwig Schmidt},
  editor       = {Kamalika Chaudhuri and
                  Stefanie Jegelka and
                  Le Song and
                  Csaba Szepesv{\'{a}}ri and
                  Gang Niu and
                  Sivan Sabato},
  title        = {Model soups: averaging weights of multiple fine-tuned models improves
                  accuracy without increasing inference time},
  booktitle    = {International Conference on Machine Learning, {ICML} 2022, 17-23 July
                  2022, Baltimore, Maryland, {USA}},
  series       = {Proceedings of Machine Learning Research},
  pages        = {23965--23998},
  publisher    = {{PMLR}},
  year         = {2022},
  url          = {https://proceedings.mlr.press/v162/wortsman22a.html},
  bibsource    = {dblp computer science bibliography, https://dblp.org}
}

@inproceedings{jain2024livecodebench,
  author       = {Naman Jain and
                  King Han and
                  Alex Gu and
                  Wen{-}Ding Li and
                  Fanjia Yan and
                  Tianjun Zhang and
                  Sida Wang and
                  Armando Solar{-}Lezama and
                  Koushik Sen and
                  Ion Stoica},
  title        = {LiveCodeBench: Holistic and Contamination Free Evaluation of Large
                  Language Models for Code},
  booktitle    = {The Thirteenth International Conference on Learning Representations,
                  {ICLR} 2025, Singapore, April 24-28, 2025},
  publisher    = {OpenReview.net},
  year         = {2025},
  url          = {https://openreview.net/forum?id=chfJJYC3iL},
  bibsource    = {dblp computer science bibliography, https://dblp.org}
}

@inproceedings{rame2023rewarded,
  author       = {Alexandre Ram{\'{e}} and
                  Guillaume Couairon and
                  Corentin Dancette and
                  Jean{-}Baptiste Gaya and
                  Mustafa Shukor and
                  Laure Soulier and
                  Matthieu Cord},
  editor       = {Alice Oh and
                  Tristan Naumann and
                  Amir Globerson and
                  Kate Saenko and
                  Moritz Hardt and
                  Sergey Levine},
  title        = {Rewarded soups: towards Pareto-optimal alignment by interpolating
                  weights fine-tuned on diverse rewards},
  booktitle    = {Advances in Neural Information Processing Systems 36: Annual Conference
                  on Neural Information Processing Systems 2023, NeurIPS 2023, New Orleans,
                  LA, USA, December 10 - 16, 2023},
  year         = {2023},
  url          = {http://papers.nips.cc/paper\_files/paper/2023/hash/e12a3b98b67e8395f639fde4c2b03168-Abstract-Conference.html},
  bibsource    = {dblp computer science bibliography, https://dblp.org}
}

@inproceedings{ilharco2023editing,
  author       = {Gabriel Ilharco and
                  Marco T{\'{u}}lio Ribeiro and
                  Mitchell Wortsman and
                  Ludwig Schmidt and
                  Hannaneh Hajishirzi and
                  Ali Farhadi},
  title        = {Editing models with task arithmetic},
  booktitle    = {The Eleventh International Conference on Learning Representations,
                  {ICLR} 2023, Kigali, Rwanda, May 1-5, 2023},
  publisher    = {OpenReview.net},
  year         = {2023},
  url          = {https://openreview.net/forum?id=6t0Kwf8-jrj},
  bibsource    = {dblp computer science bibliography, https://dblp.org}
}

@misc{ma2025swefficiencylanguagemodelsoptimize,
      title={SWE-fficiency: Can Language Models Optimize Real-World Repositories on Real Workloads?}, 
      author={Jeffrey Jian Ma and Milad Hashemi and Amir Yazdanbakhsh and Kevin Swersky and Ofir Press and Enhui Li and Vijay Janapa Reddi and Parthasarathy Ranganathan},
      year={2025},
      eprint={2511.06090},
      archivePrefix={arXiv},
      primaryClass={cs.SE},
      url={https://arxiv.org/abs/2511.06090}, 
}

@inproceedings{ouyang2025kernelbench,
  author       = {Anne Ouyang and
                  Simon Guo and
                  Simran Arora and
                  Alex L. Zhang and
                  William Hu and
                  Christopher R{\'{e}} and
                  Azalia Mirhoseini},
  editor       = {Aarti Singh and
                  Maryam Fazel and
                  Daniel Hsu and
                  Simon Lacoste{-}Julien and
                  Felix Berkenkamp and
                  Tegan Maharaj and
                  Kiri Wagstaff and
                  Jerry Zhu},
  title        = {KernelBench: Can LLMs Write Efficient {GPU} Kernels?},
  booktitle    = {Forty-second International Conference on Machine Learning, {ICML}
                  2025, Vancouver, BC, Canada, July 13-19, 2025},
  series       = {Proceedings of Machine Learning Research},
  publisher    = {{PMLR} / OpenReview.net},
  year         = {2025},
  url          = {https://proceedings.mlr.press/v267/ouyang25a.html},
  bibsource    = {dblp computer science bibliography, https://dblp.org}
}

@misc{yu2025dapo,
      title={DAPO: An Open-Source LLM Reinforcement Learning System at Scale}, 
      author={Qiying Yu and Zheng Zhang and Ruofei Zhu and Yufeng Yuan and Xiaochen Zuo and Yu Yue and Weinan Dai and Tiantian Fan and Gaohong Liu and Lingjun Liu and Xin Liu and Haibin Lin and Zhiqi Lin and Bole Ma and Guangming Sheng and Yuxuan Tong and Chi Zhang and Mofan Zhang and Wang Zhang and Hang Zhu and Jinhua Zhu and Jiaze Chen and Jiangjie Chen and Chengyi Wang and Hongli Yu and Yuxuan Song and Xiangpeng Wei and Hao Zhou and Jingjing Liu and Wei-Ying Ma and Ya-Qin Zhang and Lin Yan and Mu Qiao and Yonghui Wu and Mingxuan Wang},
      year={2025},
      eprint={2503.14476},
      archivePrefix={arXiv},
      primaryClass={cs.LG},
      url={https://arxiv.org/abs/2503.14476}, 
}

@misc{liu2025drgrpo,
      title={Understanding R1-Zero-Like Training: A Critical Perspective}, 
      author={Zichen Liu and Changyu Chen and Wenjun Li and Penghui Qi and Tianyu Pang and Chao Du and Wee Sun Lee and Min Lin},
      year={2025},
      eprint={2503.20783},
      archivePrefix={arXiv},
      primaryClass={cs.LG},
      url={https://arxiv.org/abs/2503.20783}, 
}

@misc{ahmad2025opencodereasoning,
      title={OpenCodeReasoning-II: A Simple Test Time Scaling Approach via Self-Critique}, 
      author={Wasi Uddin Ahmad and Somshubra Majumdar and Aleksander Ficek and Sean Narenthiran and Mehrzad Samadi and Jocelyn Huang and Siddhartha Jain and Vahid Noroozi and Boris Ginsburg},
      year={2025},
      eprint={2507.09075},
      archivePrefix={arXiv},
      primaryClass={cs.CL},
      url={https://arxiv.org/abs/2507.09075}, 
}

@misc{moshkov2025openmathreasoning,
      title={AIMO-2 Winning Solution: Building State-of-the-Art Mathematical Reasoning Models with OpenMathReasoning dataset}, 
      author={Ivan Moshkov and Darragh Hanley and Ivan Sorokin and Shubham Toshniwal and Christof Henkel and Benedikt Schifferer and Wei Du and Igor Gitman},
      year={2025},
      eprint={2504.16891},
      archivePrefix={arXiv},
      primaryClass={cs.AI},
      url={https://arxiv.org/abs/2504.16891}, 
}

@misc{faircodegenteam2025cwmopenweightsllmresearch,
      title={CWM: An Open-Weights LLM for Research on Code Generation with World Models}, 
      author={{{FAIR CodeGen team}} and Jade Copet and Quentin Carbonneaux and Gal Cohen and Jonas Gehring and Jacob Kahn and Jannik Kossen and Felix Kreuk and Emily McMilin and Michel Meyer and Yuxiang Wei and David Zhang and Kunhao Zheng and Jordi Armengol-Estapé and Pedram Bashiri and Maximilian Beck and Pierre Chambon and Abhishek Charnalia and Chris Cummins and Juliette Decugis and Zacharias V. Fisches and François Fleuret and Fabian Gloeckle and Alex Gu and Michael Hassid and Daniel Haziza and Badr Youbi Idrissi and Christian Keller and Rahul Kindi and Hugh Leather and Gallil Maimon and Aram Markosyan and Francisco Massa and Pierre-Emmanuel Mazaré and Vegard Mella and Naila Murray and Keyur Muzumdar and Peter O'Hearn and Matteo Pagliardini and Dmitrii Pedchenko and Tal Remez and Volker Seeker and Marco Selvi and Oren Sultan and Sida Wang and Luca Wehrstedt and Ori Yoran and Lingming Zhang and Taco Cohen and Yossi Adi and Gabriel Synnaeve},
      year={2025},
      eprint={2510.02387},
      archivePrefix={arXiv},
      primaryClass={cs.SE},
      url={https://arxiv.org/abs/2510.02387}, 
}

@misc{xue2025simpletirendtoendreinforcementlearning,
      title={SimpleTIR: End-to-End Reinforcement Learning for Multi-Turn Tool-Integrated Reasoning}, 
      author={Zhenghai Xue and Longtao Zheng and Qian Liu and Yingru Li and Xiaosen Zheng and Zejun Ma and Bo An},
      year={2025},
      eprint={2509.02479},
      archivePrefix={arXiv},
      primaryClass={cs.LG},
      url={https://arxiv.org/abs/2509.02479}, 
}

@misc{minimax2025minimaxm1scalingtesttimecompute,
      title={MiniMax-M1: Scaling Test-Time Compute Efficiently with Lightning Attention}, 
      author={MiniMax and : and Aili Chen and Aonian Li and Bangwei Gong and Binyang Jiang and Bo Fei and Bo Yang and Boji Shan and Changqing Yu and Chao Wang and Cheng Zhu and Chengjun Xiao and Chengyu Du and Chi Zhang and Chu Qiao and Chunhao Zhang and Chunhui Du and Congchao Guo and Da Chen and Deming Ding and Dianjun Sun and Dong Li and Enwei Jiao and Haigang Zhou and Haimo Zhang and Han Ding and Haohai Sun and Haoyu Feng and Huaiguang Cai and Haichao Zhu and Jian Sun and Jiaqi Zhuang and Jiaren Cai and Jiayuan Song and Jin Zhu and Jingyang Li and Jinhao Tian and Jinli Liu and Junhao Xu and Junjie Yan and Junteng Liu and Junxian He and Kaiyi Feng and Ke Yang and Kecheng Xiao and Le Han and Leyang Wang and Lianfei Yu and Liheng Feng and Lin Li and Lin Zheng and Linge Du and Lingyu Yang and Lunbin Zeng and Minghui Yu and Mingliang Tao and Mingyuan Chi and Mozhi Zhang and Mujie Lin and Nan Hu and Nongyu Di and Peng Gao and Pengfei Li and Pengyu Zhao and Qibing Ren and Qidi Xu and Qile Li and Qin Wang and Rong Tian and Ruitao Leng and Shaoxiang Chen and Shaoyu Chen and Shengmin Shi and Shitong Weng and Shuchang Guan and Shuqi Yu and Sichen Li and Songquan Zhu and Tengfei Li and Tianchi Cai and Tianrun Liang and Weiyu Cheng and Weize Kong and Wenkai Li and Xiancai Chen and Xiangjun Song and Xiao Luo and Xiao Su and Xiaobo Li and Xiaodong Han and Xinzhu Hou and Xuan Lu and Xun Zou and Xuyang Shen and Yan Gong and Yan Ma and Yang Wang and Yiqi Shi and Yiran Zhong and Yonghong Duan and Yongxiang Fu and Yongyi Hu and Yu Gao and Yuanxiang Fan and Yufeng Yang and Yuhao Li and Yulin Hu and Yunan Huang and Yunji Li and Yunzhi Xu and Yuxin Mao and Yuxuan Shi and Yuze Wenren and Zehan Li and Zelin Li and Zhanxu Tian and Zhengmao Zhu and Zhenhua Fan and Zhenzhen Wu and Zhichao Xu and Zhihang Yu and Zhiheng Lyu and Zhuo Jiang and Zibo Gao and Zijia Wu and Zijian Song and Zijun Sun},
      year={2025},
      eprint={2506.13585},
      archivePrefix={arXiv},
      primaryClass={cs.CL},
      url={https://arxiv.org/abs/2506.13585}, 
}

@misc{kimiteam2026kimik2openagentic,
      title={Kimi K2: Open Agentic Intelligence}, 
      author={{{Kimi Team}} and Yifan Bai and Yiping Bao and Y. Charles and Cheng Chen and Guanduo Chen and Haiting Chen and Huarong Chen and Jiahao Chen and Ningxin Chen and Ruijue Chen and Yanru Chen and Yuankun Chen and Yutian Chen and Zhuofu Chen and Jialei Cui and Hao Ding and Mengnan Dong and Angang Du and Chenzhuang Du and Dikang Du and Yulun Du and Yu Fan and Yichen Feng and Kelin Fu and Bofei Gao and Chenxiao Gao and Hongcheng Gao and Peizhong Gao and Tong Gao and Yuyao Ge and Shangyi Geng and Qizheng Gu and Xinran Gu and Longyu Guan and Haiqing Guo and Jianhang Guo and Xiaoru Hao and Tianhong He and Weiran He and Wenyang He and Yunjia He and Chao Hong and Hao Hu and Yangyang Hu and Zhenxing Hu and Weixiao Huang and Zhiqi Huang and Zihao Huang and Tao Jiang and Zhejun Jiang and Xinyi Jin and Yongsheng Kang and Guokun Lai and Cheng Li and Fang Li and Haoyang Li and Ming Li and Wentao Li and Yang Li and Yanhao Li and Yiwei Li and Zhaowei Li and Zheming Li and Hongzhan Lin and Xiaohan Lin and Zongyu Lin and Chengyin Liu and Chenyu Liu and Hongzhang Liu and Jingyuan Liu and Junqi Liu and Liang Liu and Shaowei Liu and T. Y. Liu and Tianwei Liu and Weizhou Liu and Yangyang Liu and Yibo Liu and Yiping Liu and Yue Liu and Zhengying Liu and Enzhe Lu and Haoyu Lu and Lijun Lu and Yashuo Luo and Shengling Ma and Xinyu Ma and Yingwei Ma and Shaoguang Mao and Jie Mei and Xin Men and Yibo Miao and Siyuan Pan and Yebo Peng and Ruoyu Qin and Zeyu Qin and Bowen Qu and Zeyu Shang and Lidong Shi and Shengyuan Shi and Feifan Song and Jianlin Su and Zhengyuan Su and Lin Sui and Xinjie Sun and Flood Sung and Yunpeng Tai and Heyi Tang and Jiawen Tao and Qifeng Teng and Chaoran Tian and Chensi Wang and Dinglu Wang and Feng Wang and Hailong Wang and Haiming Wang and Jianzhou Wang and Jiaxing Wang and Jinhong Wang and Shengjie Wang and Shuyi Wang and Si Wang and Xinyuan Wang and Yao Wang and Yejie Wang and Yiqin Wang and Yuxin Wang and Yuzhi Wang and Zhaoji Wang and Zhengtao Wang and Zhengtao Wang and Zhexu Wang and Chu Wei and Qianqian Wei and Haoning Wu and Wenhao Wu and Xingzhe Wu and Yuxin Wu and Chenjun Xiao and Jin Xie and Xiaotong Xie and Weimin Xiong and Boyu Xu and Jinjing Xu and L. H. Xu and Lin Xu and Suting Xu and Weixin Xu and Xinran Xu and Yangchuan Xu and Ziyao Xu and Jing Xu and Jing Xu and Junjie Yan and Yuzi Yan and Hao Yang and Xiaofei Yang and Yi Yang and Ying Yang and Zhen Yang and Zhilin Yang and Zonghan Yang and Haotian Yao and Xingcheng Yao and Wenjie Ye and Zhuorui Ye and Bohong Yin and Longhui Yu and Enming Yuan and Hongbang Yuan and Mengjie Yuan and Siyu Yuan and Haobing Zhan and Dehao Zhang and Hao Zhang and Wanlu Zhang and Xiaobin Zhang and Yadong Zhang and Yangkun Zhang and Yichi Zhang and Yizhi Zhang and Yongting Zhang and Yu Zhang and Yutao Zhang and Yutong Zhang and Zheng Zhang and Haotian Zhao and Yikai Zhao and Zijia Zhao and Huabin Zheng and Shaojie Zheng and Longguang Zhong and Jianren Zhou and Xinyu Zhou and Zaida Zhou and Jinguo Zhu and Zhen Zhu and Weiyu Zhuang and Xinxing Zu},
      year={2026},
      eprint={2507.20534},
      archivePrefix={arXiv},
      primaryClass={cs.LG},
      url={https://arxiv.org/abs/2507.20534}, 
}

@inproceedings{hendrycks2021measuringcodingchallengecompetence,
  author       = {Dan Hendrycks and
                  Steven Basart and
                  Saurav Kadavath and
                  Mantas Mazeika and
                  Akul Arora and
                  Ethan Guo and
                  Collin Burns and
                  Samir Puranik and
                  Horace He and
                  Dawn Song and
                  Jacob Steinhardt},
  editor       = {Joaquin Vanschoren and
                  Sai{-}Kit Yeung},
  title        = {Measuring Coding Challenge Competence With {APPS}},
  booktitle    = {Proceedings of the Neural Information Processing Systems Track on
                  Datasets and Benchmarks 1, NeurIPS Datasets and Benchmarks 2021, December
                  2021, virtual},
  year         = {2021},
  url          = {https://datasets-benchmarks-proceedings.neurips.cc/paper/2021/hash/c24cd76e1ce41366a4bbe8a49b02a028-Abstract-round2.html},
  bibsource    = {dblp computer science bibliography, https://dblp.org}
}

@misc{li2023tacotopicsalgorithmiccode,
      title={TACO: Topics in Algorithmic COde generation dataset}, 
      author={Rongao Li and Jie Fu and Bo-Wen Zhang and Tao Huang and Zhihong Sun and Chen Lyu and Guang Liu and Zhi Jin and Ge Li},
      year={2023},
      eprint={2312.14852},
      archivePrefix={arXiv},
      primaryClass={cs.AI},
      url={https://arxiv.org/abs/2312.14852}, 
}

@inproceedings{wang2024localizingtaskinformationimproved,
  author       = {Ke Wang and
                  Nikolaos Dimitriadis and
                  Guillermo Ortiz{-}Jim{\'{e}}nez and
                  Fran{\c{c}}ois Fleuret and
                  Pascal Frossard},
  editor       = {Ruslan Salakhutdinov and
                  Zico Kolter and
                  Katherine A. Heller and
                  Adrian Weller and
                  Nuria Oliver and
                  Jonathan Scarlett and
                  Felix Berkenkamp},
  title        = {Localizing Task Information for Improved Model Merging and Compression},
  booktitle    = {Forty-first International Conference on Machine Learning, {ICML} 2024,
                  Vienna, Austria, July 21-27, 2024},
  series       = {Proceedings of Machine Learning Research},
  pages        = {50268--50287},
  publisher    = {{PMLR} / OpenReview.net},
  year         = {2024},
  url          = {https://proceedings.mlr.press/v235/wang24k.html},
  bibsource    = {dblp computer science bibliography, https://dblp.org}
}

@misc{rame2024warpbenefitsweightaveraged,
      title={WARP: On the Benefits of Weight Averaged Rewarded Policies}, 
      author={Alexandre Ramé and Johan Ferret and Nino Vieillard and Robert Dadashi and Léonard Hussenot and Pierre-Louis Cedoz and Pier Giuseppe Sessa and Sertan Girgin and Arthur Douillard and Olivier Bachem},
      year={2024},
      eprint={2406.16768},
      archivePrefix={arXiv},
      primaryClass={cs.LG},
      url={https://arxiv.org/abs/2406.16768}, 
}

@inproceedings{rame2024warmbenefitsweightaveraged,
  author       = {Alexandre Ram{\'{e}} and
                  Nino Vieillard and
                  L{\'{e}}onard Hussenot and
                  Robert Dadashi and
                  Geoffrey Cideron and
                  Olivier Bachem and
                  Johan Ferret},
  editor       = {Ruslan Salakhutdinov and
                  Zico Kolter and
                  Katherine A. Heller and
                  Adrian Weller and
                  Nuria Oliver and
                  Jonathan Scarlett and
                  Felix Berkenkamp},
  title        = {{WARM:} On the Benefits of Weight Averaged Reward Models},
  booktitle    = {Forty-first International Conference on Machine Learning, {ICML} 2024,
                  Vienna, Austria, July 21-27, 2024},
  series       = {Proceedings of Machine Learning Research},
  pages        = {42048--42073},
  publisher    = {{PMLR} / OpenReview.net},
  year         = {2024},
  url          = {https://proceedings.mlr.press/v235/rame24a.html},
  bibsource    = {dblp computer science bibliography, https://dblp.org}
}

@inproceedings{yang2024adamergingadaptivemodelmerging,
  author       = {Enneng Yang and
                  Zhenyi Wang and
                  Li Shen and
                  Shiwei Liu and
                  Guibing Guo and
                  Xingwei Wang and
                  Dacheng Tao},
  title        = {AdaMerging: Adaptive Model Merging for Multi-Task Learning},
  booktitle    = {The Twelfth International Conference on Learning Representations,
                  {ICLR} 2024, Vienna, Austria, May 7-11, 2024},
  publisher    = {OpenReview.net},
  year         = {2024},
  url          = {https://openreview.net/forum?id=nZP6NgD3QY},
  bibsource    = {dblp computer science bibliography, https://dblp.org}
}

@inproceedings{lin2024mitigatingalignmenttaxrlhf,
  author       = {Yong Lin and
                  Hangyu Lin and
                  Wei Xiong and
                  Shizhe Diao and
                  Jianmeng Liu and
                  Jipeng Zhang and
                  Rui Pan and
                  Haoxiang Wang and
                  Wenbin Hu and
                  Hanning Zhang and
                  Hanze Dong and
                  Renjie Pi and
                  Han Zhao and
                  Nan Jiang and
                  Heng Ji and
                  Yuan Yao and
                  Tong Zhang},
  editor       = {Yaser Al{-}Onaizan and
                  Mohit Bansal and
                  Yun{-}Nung Chen},
  title        = {Mitigating the Alignment Tax of {RLHF}},
  booktitle    = {Proceedings of the 2024 Conference on Empirical Methods in Natural
                  Language Processing, {EMNLP} 2024, Miami, FL, USA, November 12-16,
                  2024},
  pages        = {580--606},
  publisher    = {Association for Computational Linguistics},
  year         = {2024},
  url          = {https://doi.org/10.18653/v1/2024.emnlp-main.35},
  doi          = {10.18653/V1/2024.EMNLP-MAIN.35},
  bibsource    = {dblp computer science bibliography, https://dblp.org}
}

@inproceedings{zheng2025modelextrapolationexpeditesalignment,
  author       = {Chujie Zheng and
                  Ziqi Wang and
                  Heng Ji and
                  Minlie Huang and
                  Nanyun Peng},
  editor       = {Wanxiang Che and
                  Joyce Nabende and
                  Ekaterina Shutova and
                  Mohammad Taher Pilehvar},
  title        = {Model Extrapolation Expedites Alignment},
  booktitle    = {Proceedings of the 63rd Annual Meeting of the Association for Computational
                  Linguistics (Volume 1: Long Papers), {ACL} 2025, Vienna, Austria,
                  July 27 - August 1, 2025},
  pages        = {1025--1041},
  publisher    = {Association for Computational Linguistics},
  year         = {2025},
  url          = {https://aclanthology.org/2025.acl-long.51/},
  bibsource    = {dblp computer science bibliography, https://dblp.org}
}

@inproceedings{du2024mercurycodeefficiencybenchmark,
  author       = {Mingzhe Du and
                  Anh Tuan Luu and
                  Bin Ji and
                  Qian Liu and
                  See{-}Kiong Ng},
  editor       = {Amir Globersons and
                  Lester Mackey and
                  Danielle Belgrave and
                  Angela Fan and
                  Ulrich Paquet and
                  Jakub M. Tomczak and
                  Cheng Zhang},
  title        = {Mercury: {A} Code Efficiency Benchmark for Code Large Language Models},
  booktitle    = {Advances in Neural Information Processing Systems 38: Annual Conference
                  on Neural Information Processing Systems 2024, NeurIPS 2024, Vancouver,
                  BC, Canada, December 10 - 15, 2024},
  year         = {2024},
  url          = {http://papers.nips.cc/paper\_files/paper/2024/hash/1df1df43b58845650b8dada00fca9772-Abstract-Datasets\_and\_Benchmarks\_Track.html},
  bibsource    = {dblp computer science bibliography, https://dblp.org}
}

@inproceedings{waghjale2024eccoimprovemodelgeneratedcode,
  author       = {Siddhant Waghjale and
                  Vishruth Veerendranath and
                  Zhiruo Wang and
                  Daniel Fried},
  editor       = {Yaser Al{-}Onaizan and
                  Mohit Bansal and
                  Yun{-}Nung Chen},
  title        = {{ECCO:} Can We Improve Model-Generated Code Efficiency Without Sacrificing
                  Functional Correctness?},
  booktitle    = {Proceedings of the 2024 Conference on Empirical Methods in Natural
                  Language Processing, {EMNLP} 2024, Miami, FL, USA, November 12-16,
                  2024},
  pages        = {15362--15376},
  publisher    = {Association for Computational Linguistics},
  year         = {2024},
  url          = {https://doi.org/10.18653/v1/2024.emnlp-main.859},
  doi          = {10.18653/V1/2024.EMNLP-MAIN.859},
  bibsource    = {dblp computer science bibliography, https://dblp.org}
}

@inproceedings{huang2025effibenchbenchmarkingefficiencyautomatically,
  author       = {Dong Huang and
                  Yuhao Qing and
                  Weiyi Shang and
                  Heming Cui and
                  Jie Zhang},
  editor       = {Amir Globersons and
                  Lester Mackey and
                  Danielle Belgrave and
                  Angela Fan and
                  Ulrich Paquet and
                  Jakub M. Tomczak and
                  Cheng Zhang},
  title        = {EffiBench: Benchmarking the Efficiency of Automatically Generated
                  Code},
  booktitle    = {Advances in Neural Information Processing Systems 38: Annual Conference
                  on Neural Information Processing Systems 2024, NeurIPS 2024, Vancouver,
                  BC, Canada, December 10 - 15, 2024},
  year         = {2024},
  url          = {http://papers.nips.cc/paper\_files/paper/2024/hash/15807b6e09d691fe5e96cdecde6d7b80-Abstract-Datasets\_and\_Benchmarks\_Track.html},
  bibsource    = {dblp computer science bibliography, https://dblp.org}
}

@misc{du2025afterburnerreinforcementlearningfacilitates,
      title={Afterburner: Reinforcement Learning Facilitates Self-Improving Code Efficiency Optimization}, 
      author={Mingzhe Du and Luu Anh Tuan and Yue Liu and Yuhao Qing and Dong Huang and Xinyi He and Qian Liu and Zejun Ma and See-kiong Ng},
      year={2025},
      eprint={2505.23387},
      archivePrefix={arXiv},
      primaryClass={cs.SE},
      url={https://arxiv.org/abs/2505.23387}, 
}

@misc{chambon2025bigobenchllmsgenerate,
      title={BigO(Bench) -- Can LLMs Generate Code with Controlled Time and Space Complexity?}, 
      author={Pierre Chambon and Baptiste Roziere and Benoit Sagot and Gabriel Synnaeve},
      year={2025},
      eprint={2503.15242},
      archivePrefix={arXiv},
      primaryClass={cs.CL},
      url={https://arxiv.org/abs/2503.15242}, 
}

@inproceedings{gehring2025rlefgroundingcodellms,
  author       = {Jonas Gehring and
                  Kunhao Zheng and
                  Jade Copet and
                  Vegard Mella and
                  Taco Cohen and
                  Gabriel Synnaeve},
  editor       = {Aarti Singh and
                  Maryam Fazel and
                  Daniel Hsu and
                  Simon Lacoste{-}Julien and
                  Felix Berkenkamp and
                  Tegan Maharaj and
                  Kiri Wagstaff and
                  Jerry Zhu},
  title        = {{RLEF:} Grounding Code LLMs in Execution Feedback with Reinforcement
                  Learning},
  booktitle    = {Forty-second International Conference on Machine Learning, {ICML}
                  2025, Vancouver, BC, Canada, July 13-19, 2025},
  series       = {Proceedings of Machine Learning Research},
  publisher    = {{PMLR} / OpenReview.net},
  year         = {2025},
  url          = {https://proceedings.mlr.press/v267/gehring25a.html},
  bibsource    = {dblp computer science bibliography, https://dblp.org}
}

@inproceedings{zheng2025makeslargelanguagemodels,
  author       = {Kunhao Zheng and
                  Juliette Decugis and
                  Jonas Gehring and
                  Taco Cohen and
                  Benjamin N{\'{e}}Xuanjinggrevergne and
                  Gabriel Synnaeve},
  title        = {What Makes Large Language Models Reason in (Multi-Turn) Code Generation?},
  booktitle    = {The Thirteenth International Conference on Learning Representations,
                  {ICLR} 2025, Singapore, April 24-28, 2025},
  publisher    = {OpenReview.net},
  year         = {2025},
  url          = {https://openreview.net/forum?id=Zk9guOl9NS},
  bibsource    = {dblp computer science bibliography, https://dblp.org}
}

@inproceedings{liu2023codegeneratedchatgptreally,
  author       = {Jiawei Liu and
                  Chunqiu Steven Xia and
                  Yuyao Wang and
                  Lingming Zhang},
  editor       = {Alice Oh and
                  Tristan Naumann and
                  Amir Globerson and
                  Kate Saenko and
                  Moritz Hardt and
                  Sergey Levine},
  title        = {Is Your Code Generated by ChatGPT Really Correct? Rigorous Evaluation
                  of Large Language Models for Code Generation},
  booktitle    = {Advances in Neural Information Processing Systems 36: Annual Conference
                  on Neural Information Processing Systems 2023, NeurIPS 2023, New Orleans,
                  LA, USA, December 10 - 16, 2023},
  year         = {2023},
  url          = {http://papers.nips.cc/paper\_files/paper/2023/hash/43e9d647ccd3e4b7b5baab53f0368686-Abstract-Conference.html},
  bibsource    = {dblp computer science bibliography, https://dblp.org}
}

@misc{kim2026scalingtesttimecomputeagentic,
      title={Scaling Test-Time Compute for Agentic Coding}, 
      author={Joongwon Kim and Wannan Yang and Kelvin Niu and Hongming Zhang and Yun Zhu and Eryk Helenowski and Ruan Silva and Zhengxing Chen and Srinivasan Iyer and Manzil Zaheer and Daniel Fried and Hannaneh Hajishirzi and Sanjeev Arora and Gabriel Synnaeve and Ruslan Salakhutdinov and Anirudh Goyal},
      year={2026},
      eprint={2604.16529},
      archivePrefix={arXiv},
      primaryClass={cs.SE},
      url={https://arxiv.org/abs/2604.16529}, 
}

@inproceedings{li-etal-2025-codeprm,
    title = "{C}ode{PRM}: Execution Feedback-enhanced Process Reward Model for Code Generation",
    author = "Li, Qingyao  and
      Dai, Xinyi  and
      Li, Xiangyang  and
      Zhang, Weinan  and
      Wang, Yasheng  and
      Tang, Ruiming  and
      Yu, Yong",
    editor = "Che, Wanxiang  and
      Nabende, Joyce  and
      Shutova, Ekaterina  and
      Pilehvar, Mohammad Taher",
    booktitle = "Findings of the Association for Computational Linguistics: ACL 2025",
    month = jul,
    year = "2025",
    address = "Vienna, Austria",
    publisher = "Association for Computational Linguistics",
    url = "https://aclanthology.org/2025.findings-acl.428/",
    doi = "10.18653/v1/2025.findings-acl.428",
    pages = "8169--8182",
    ISBN = "979-8-89176-256-5"
}

@misc{dai2025processsupervisionguidedpolicyoptimization,
      title={Process Supervision-Guided Policy Optimization for Code Generation}, 
      author={Ning Dai and Zheng Wu and Renjie Zheng and Ziyun Wei and Wenlei Shi and Xing Jin and Guanlin Liu and Chen Dun and Liang Huang and Lin Yan},
      year={2025},
      eprint={2410.17621},
      archivePrefix={arXiv},
      primaryClass={cs.AI},
      url={https://arxiv.org/abs/2410.17621}, 
}

@misc{jin2025revealselfevolvingcodeagents,
      title={ReVeal: Self-Evolving Code Agents via Reliable Self-Verification}, 
      author={Yiyang Jin and Kunzhao Xu and Hang Li and Xueting Han and Yanmin Zhou and Cheng Li and Jing Bai},
      year={2025},
      eprint={2506.11442},
      archivePrefix={arXiv},
      primaryClass={cs.SE},
      url={https://arxiv.org/abs/2506.11442}, 
}

@inproceedings{li2025stesttimescaling,
  author       = {Dacheng Li and
                  Shiyi Cao and
                  Chengkun Cao and
                  Xiuyu Li and
                  Shangyin Tan and
                  Kurt Keutzer and
                  Jiarong Xing and
                  Joseph E. Gonzalez and
                  Ion Stoica},
  editor       = {Christos Christodoulopoulos and
                  Tanmoy Chakraborty and
                  Carolyn Rose and
                  Violet Peng},
  title        = {S*: Test Time Scaling for Code Generation},
  booktitle    = {Findings of the Association for Computational Linguistics: {EMNLP}
                  2025, Suzhou, China, November 4-9, 2025},
  pages        = {15964--15978},
  publisher    = {Association for Computational Linguistics},
  year         = {2025},
  url          = {https://aclanthology.org/2025.findings-emnlp.865/},
  bibsource    = {dblp computer science bibliography, https://dblp.org}
}

@misc{maimon2026selfexecutionsimulationimprovescoding,
      title={Self-Execution Simulation Improves Coding Models}, 
      author={Gallil Maimon and Ori Yoran and Felix Kreuk and Michael Hassid and Gal Cohen and Pierre Chambon and Yossi Adi},
      year={2026},
      eprint={2604.03253},
      archivePrefix={arXiv},
      primaryClass={cs.CL},
      url={https://arxiv.org/abs/2604.03253}, 
}

@misc{gan2026neuralthicketsdiversetask,
      title={Neural Thickets: Diverse Task Experts Are Dense Around Pretrained Weights}, 
      author={Yulu Gan and Phillip Isola},
      year={2026},
      eprint={2603.12228},
      archivePrefix={arXiv},
      primaryClass={cs.LG},
      url={https://arxiv.org/abs/2603.12228}, 
}

@inproceedings{tang2025optimizinglanguagemodelsinference,
  author       = {Yunhao Tang and
                  Kunhao Zheng and
                  Gabriel Synnaeve and
                  R{\'{e}}mi Munos},
  editor       = {Aarti Singh and
                  Maryam Fazel and
                  Daniel Hsu and
                  Simon Lacoste{-}Julien and
                  Felix Berkenkamp and
                  Tegan Maharaj and
                  Kiri Wagstaff and
                  Jerry Zhu},
  title        = {Optimizing Language Models for Inference Time Objectives using Reinforcement
                  Learning},
  booktitle    = {Forty-second International Conference on Machine Learning, {ICML}
                  2025, Vancouver, BC, Canada, July 13-19, 2025},
  series       = {Proceedings of Machine Learning Research},
  publisher    = {{PMLR} / OpenReview.net},
  year         = {2025},
  url          = {https://proceedings.mlr.press/v267/tang25o.html},
  bibsource    = {dblp computer science bibliography, https://dblp.org}
}

@inproceedings{verine2025improvingdiversitylanguagemodels,
  author       = {Alexandre Verine and
                  Florian Le Bronnec and
                  Kunhao Zheng and
                  Alexandre Allauzen and
                  Yann Chevaleyre and
                  Benjamin N{\'{e}}grevergne},
  editor       = {Aarti Singh and
                  Maryam Fazel and
                  Daniel Hsu and
                  Simon Lacoste{-}Julien and
                  Felix Berkenkamp and
                  Tegan Maharaj and
                  Kiri Wagstaff and
                  Jerry Zhu},
  title        = {Improving Diversity in Language Models: When Temperature Fails, Change
                  the Loss},
  booktitle    = {Forty-second International Conference on Machine Learning, {ICML}
                  2025, Vancouver, BC, Canada, July 13-19, 2025},
  series       = {Proceedings of Machine Learning Research},
  publisher    = {{PMLR} / OpenReview.net},
  year         = {2025},
  url          = {https://proceedings.mlr.press/v267/verine25a.html},
  bibsource    = {dblp computer science bibliography, https://dblp.org}
}

@misc{cai2026predictabilityreinforcementlearningdynamics,
      title={On Predictability of Reinforcement Learning Dynamics for Large Language Models}, 
      author={Yuchen Cai and Ding Cao and Xin Xu and Zijun Yao and Yuqing Huang and Zhenyu Tan and Benyi Zhang and Guangzhong Sun and Guiquan Liu and Junfeng Fang},
      year={2026},
      eprint={2510.00553},
      archivePrefix={arXiv},
      primaryClass={cs.LG},
      url={https://arxiv.org/abs/2510.00553}, 
}

@inproceedings{pan2022effects,
  author    = {Alexander Pan and
               Kush Bhatia and
               Jacob Steinhardt},
  title     = {The Effects of Reward Misspecification: Mapping and Mitigating
               Misaligned Models},
  booktitle = {International Conference on Learning Representations},
  year      = {2022},
  url       = {https://openreview.net/forum?id=JYtwGwIL7ye},
  bibsource = {dblp computer science bibliography, https://dblp.org}
}

@inproceedings{skalse2022defining,
  author    = {Joar Skalse and
               Nikolaus H. R. Howe and
               Dmitrii Krasheninnikov and
               David Krueger},
  title     = {Defining and Characterizing Reward Gaming},
  booktitle = {Advances in Neural Information Processing Systems},
  volume    = {35},
  pages     = {9460--9471},
  year      = {2022},
  url       = {https://papers.nips.cc/paper_files/paper/2022/hash/3d719fee332caa23d5038b8a90e81796-Abstract-Conference.html},
  bibsource = {dblp computer science bibliography, https://dblp.org}
}

@article{baker2025monitoring,
  author     = {Bowen Baker and
                Joost Huizinga and
                Leo Gao and
                Zehao Dou and
                Melody Y. Guan and
                Aleksander Madry and
                Wojciech Zaremba and
                Jakub Pachocki and
                David Farhi},
  title      = {Monitoring Reasoning Models for Misbehavior and the Risks of
                Promoting Obfuscation},
  journal    = {CoRR},
  volume     = {abs/2503.11926},
  year       = {2025},
  url        = {https://doi.org/10.48550/arXiv.2503.11926},
  doi        = {10.48550/ARXIV.2503.11926},
  eprinttype = {arXiv},
  eprint     = {2503.11926},
  bibsource  = {dblp computer science bibliography, https://dblp.org}
}

@article{helff2026gaming,
  author     = {Lukas Helff and
                Quentin Delfosse and
                David Steinmann and
                Ruben H{\"a}rle and
                Hikaru Shindo and
                Patrick Schramowski and
                Wolfgang Stammer and
                Kristian Kersting and
                Felix Friedrich},
  title      = {{LLMs} Gaming Verifiers: {RLVR} Can Lead to Reward Hacking},
  journal    = {CoRR},
  volume     = {abs/2604.15149},
  year       = {2026},
  url        = {https://doi.org/10.48550/arXiv.2604.15149},
  doi        = {10.48550/ARXIV.2604.15149},
  eprinttype = {arXiv},
  eprint     = {2604.15149},
  bibsource  = {dblp computer science bibliography, https://dblp.org}
}

@inproceedings{gao2023scaling,
  author       = {Leo Gao and
                  John Schulman and
                  Jacob Hilton},
  editor       = {Andreas Krause and
                  Emma Brunskill and
                  Kyunghyun Cho and
                  Barbara Engelhardt and
                  Sivan Sabato and
                  Jonathan Scarlett},
  title        = {Scaling Laws for Reward Model Overoptimization},
  booktitle    = {International Conference on Machine Learning, {ICML} 2023, 23-29 July
                  2023, Honolulu, Hawaii, {USA}},
  series       = {Proceedings of Machine Learning Research},
  volume       = {202},
  pages        = {10835--10866},
  publisher    = {{PMLR}},
  year         = {2023},
  url          = {https://proceedings.mlr.press/v202/gao23h.html},
  bibsource    = {dblp computer science bibliography, https://dblp.org}
}

@inproceedings{izmailov2018averaging,
  author    = {Pavel Izmailov and
               Dmitrii Podoprikhin and
               Timur Garipov and
               Dmitry P. Vetrov and
               Andrew Gordon Wilson},
  title     = {Averaging Weights Leads to Wider Optima and Better Generalization},
  booktitle = {Proceedings of the Thirty-Fourth Conference on
               Uncertainty in Artificial Intelligence},
  pages     = {876--885},
  publisher = {{AUAI} Press},
  year      = {2018},
  url       = {http://auai.org/uai2018/proceedings/papers/313.pdf}
}

@inproceedings{garipov2018loss,
  author    = {Timur Garipov and
               Pavel Izmailov and
               Dmitrii Podoprikhin and
               Dmitry P. Vetrov and
               Andrew Gordon Wilson},
  title     = {Loss Surfaces, Mode Connectivity, and Fast Ensembling of DNNs},
  booktitle = {Advances in Neural Information Processing Systems 31},
  pages     = {8803--8812},
  year      = {2018},
  url       = {https://proceedings.neurips.cc/paper/2018/hash/be3087e74e9100d4bc4c6268cdbe8456-Abstract.html}
}

@inproceedings{nikishin2018improving,
  author    = {Evgenii Nikishin and
               Pavel Izmailov and
               Ben Athiwaratkun and
               Dmitrii Podoprikhin and
               Timur Garipov and
               Pavel Shvechikov and
               Dmitry Vetrov and
               Andrew Gordon Wilson},
  title     = {Improving Stability in Deep Reinforcement Learning
               with Weight Averaging},
  booktitle = {UAI 2018 Workshop on Uncertainty in Deep Learning},
  year      = {2018},
  url       = {https://www.gatsby.ucl.ac.uk/~balaji/udl-camera-ready/UDL-24.pdf}
}

@inproceedings{rimsky2024steering,
  author    = {Nina Rimsky and
               Nick Gabrieli and
               Julian Schulz and
               Meg Tong and
               Evan Hubinger and
               Alexander Matt Turner},
  title     = {Steering Llama 2 via Contrastive Activation Addition},
  booktitle = {Proceedings of the 62nd Annual Meeting of the
               Association for Computational Linguistics
               (Volume 1: Long Papers)},
  pages     = {15504--15522},
  publisher = {Association for Computational Linguistics},
  year      = {2024},
  doi       = {10.18653/V1/2024.ACL-LONG.828},
  url       = {https://doi.org/10.18653/v1/2024.acl-long.828}
}

\newpage
\appendix

\clearpage
\section{Experimental Details}
\label{app:experimental-details}

\label{app:main-figures}
\label{sec:appendix-frontier-by-setting}

\subsection{Test Input Length Distribution}
\label{sec:appendix-input-len}

\begin{figure}[H]
    \centering
    \includegraphics[width=0.38\linewidth]{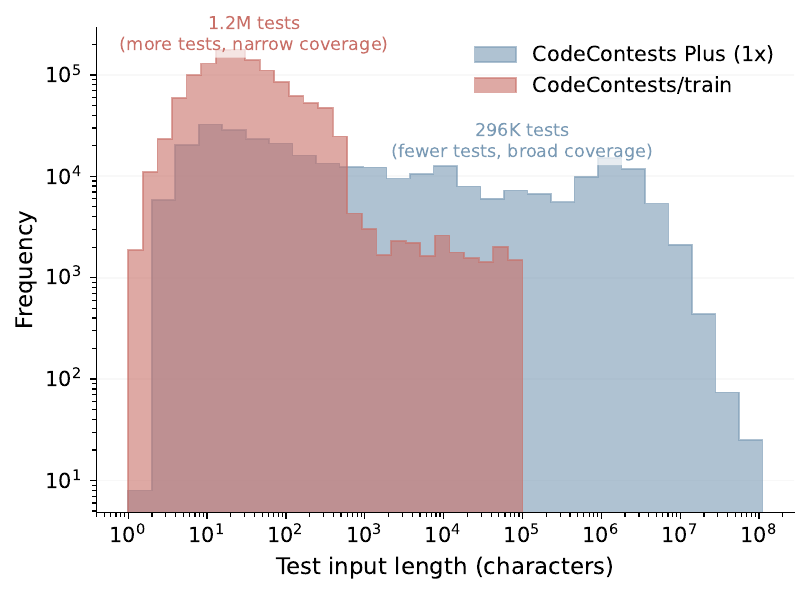}
    \hfill
    \includegraphics[width=0.58\linewidth]{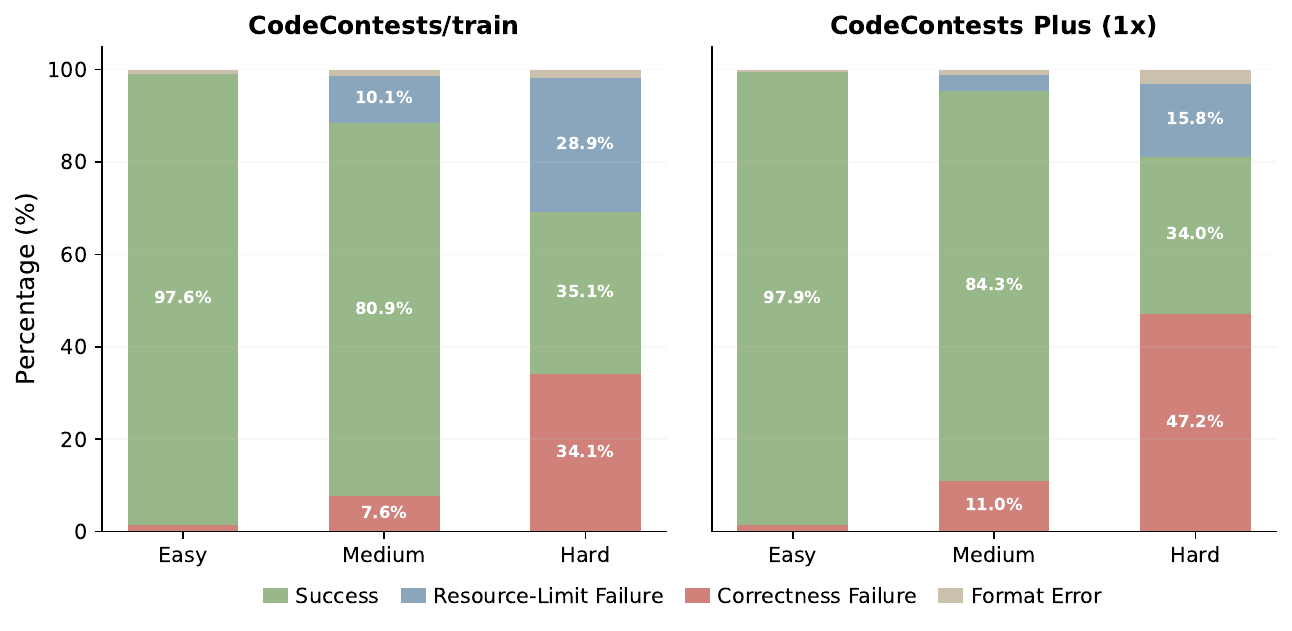}
    \caption{\textbf{Left}: Distribution of test
    case input lengths.
    \codecontestsplus (blue) has fewer tests than
    CodeContests/train (red) but covers a much wider
    range of input sizes.
    \textbf{Right}: Error composition after RL
    training on CodeContests/train vs.\
    \codecontestsplus, by difficulty split on LCB.
    Despite having more tests,
    training on CodeContests/train produces more
    resource-limit failures on LCB/hard problems
    (28.9\% vs.\ 15.8\%), because its tests rarely
    reach the input sizes that stress algorithmic
    efficiency.}
    \label{fig:input-len}
\end{figure}

Figure~\ref{fig:input-len} compares the input character length distributions of CodeContests/train and \codecontestsplus, alongside error decompositions of RL checkpoints trained on each and evaluated on LCB/hard. Despite having more test
cases, CodeContests/train produces more resource-limit failures on LCB/hard (28.9\% vs.\ 15.8\%), because its tests rarely reach the input sizes that stress algorithmic efficiency. The number of tests alone does not determine coverage; it is the
range of input sizes, and thus the optimization pressure they induce, that drives the trade-off frontier.

\subsection{7B Model SFT Training Details}
\label{sec:appendix-7b-sft-training}

We start from
Qwen~2.5~7B~\citep{qwen2025qwen25technicalreport} base model and fine-tune on a mixture of OpenCodeReasoning-2~\citep{ahmad2025opencodereasoning} (Python and C++) and OpenMathReasoning~\citep{moshkov2025openmathreasoning} for exactly 1 epoch. Training runs for 24,829 steps with 16 H100 nodes. We use the following hyperparameters: local batch size 2, sequence length 32,768, learning rate $8.6 \times 10^{-6}$ with 1,000 warmup steps followed by a constant schedule, and gradient norm clipped at 1.0. We use FSDP in bf16 with tensor parallelism size 4. Linear layers use rowwise FP8 quantization (one scaling factor per matrix row for both activations and weights); all other computations are in bf16.

\subsection{7B Model RL Training Details}
\label{sec:appendix-7b-rl-training}

We train with a simplified asynchronous online RL recipe as shown in~\citet{faircodegenteam2025cwmopenweightsllmresearch}, namely Dr.GRPO~\citep{liu2025drgrpo} using DAPO-style~\citep{yu2025dapo} asymmetric clipping ($\epsilon_{\mathrm{low}}{=}0.2$, $\epsilon_{\mathrm{high}}{=}0.25$) and zero-advantage skipping, with no KL penalty ($\beta{=}0$) and no reference model. Each prompt produces 16 rollouts at temperature 1.0. We use AdamW ($\beta_1{=}0.9$, $\beta_2{=}0.95$, weight decay $0.1$) with learning rate $1.4 \times 10^{-7}$, 100 warmup steps followed by a constant schedule, and gradient norm clipped at 0.1. Batching is token-based with a maximum of 32,768 tokens; sequence length is 32,768. The worker model is updated every 4 gradient steps in the trainer side. Training runs for 6,000 steps across 32 H100 nodes (no tensor parallelism) in bf16 with selective activation checkpointing, with 1:1 trainer:worker ratio. No FP8 is used for RL training.

We train on \codecontestsplus (1$\times$) with binary reward, sweeping the input-length threshold $k \in \{2,\ldots,6,\infty\}$ across independent runs. Evaluation on LCB/v5 is with temperature 0.6, top-$p$ 0.95, and 10 samples per problem. A single full RL run takes $\sim$19 hours wall-clock.

\subsection{32B Pure-Reasoning and Agentic Coding RL Training Details}
\label{sec:appendix-32b-pr-agentic}

Both the pure-reasoning and agentic-coding settings
start from the same CWM-SFT 32B checkpoint and share
most hyperparameters with the tool-use setting described in Appendix~\ref{sec:appendix-tool-training}. We list the shared configuration
first, then note setting-specific differences.

\paragraph{Shared configuration.}
Both settings use Dr.GRPO with DAPO-style asymmetric
clipping ($\epsilon_{\mathrm{low}}{=}0.2$,
$\epsilon_{\mathrm{high}}{=}0.25$) and zero-advantage
skipping, with no KL penalty and no reference model.
Reward function is binary.
AdamW with learning rate $1.4 \times 10^{-7}$,
constant schedule with 400-step warmup, and gradient
norm clipped at 0.1.
Each prompt produces 16 rollouts at temperature 1.0.
The trainer-to-worker ratio is 1:1, with the model
updated every 4 worker steps.
Training uses 256 H100 GPUs
(32 DP $\times$ 8 TP) with FSDP in bf16,
torch.compile, and selective activation checkpointing.
Pure reasoning trains for 8\,000 steps
($\sim$44 hours per run); agentic coding trains for
4\,000 steps ($\sim$16 hours per run).
We sweep the input-length threshold
$k \in \{2,\ldots,6,\infty\}$ across independent runs,
identical to the tool-use setting.

\paragraph{Pure reasoning.}
The pure-reasoning setting use sequence length
32\,768 and no tool calls.
The model generates a single reasoning trace in a
\texttt{<think>} block followed by the code
submission. Evaluation uses temperature 0.6 and
top-$p$ 0.95 with 10 samples per problem.

\paragraph{Agentic coding.}
The agentic setting uses a sandboxed terminal
environment backed by Modal, a cloud sandbox
service, with sequence length 61\,440 and up to
72 interaction turns per trajectory. The model
can execute arbitrary shell commands, write and run
code, and observe terminal output within a
sandbox that enforces a 120-second session timeout
and 1-8\,GB memory limits with network access
blocked.
Evaluation uses temperature 1.0 and top-$p$ 0.95
with 5 samples per problem.

\subsection{32B Tool-Use Model Training Details}
\label{sec:appendix-tool-training}

We describe two parts of the 32B tool-use setup here.

First, two RL-to-SFT bootstrapping iterations build one shared SFT initialization, because CWM-SFT does not reliably use our stdin-oriented \texttt{python-io} interface for competitive-programming tasks.

Second, starting from that same initialization, i.e., SFT checkpoint, the final RL phase launches the verifier-strictness sweep analyzed in the main text. In this final sweep, the compared checkpoints are trained under nested binary coverage rewards, with the task distribution, tool interface, and RL hyperparameters fixed across thresholds. Thus the comparison varies unit-test coverage, not the earlier bootstrapping curriculum.

The bootstrapping iteration proceeds as follows:
\begin{enumerate}
    \item \textbf{Iteration~0 (RL).}
      Starting from CWM-SFT, run RL with reward
      shaping on competitive-programming problems
      to teach the model to invoke a Python
      interpreter that accepts standard input
      (\texttt{python-io}).
      Correct trajectories are collected, filtered and downsampled for SFT.
    \item \textbf{Iteration~1 (SFT $\to$ RL).}
      Distill the collected tool-use trajectories from Iteration~0
      via SFT back into CWM-SFT;
      then run RL on \codecontestsplus with reward
      shaping that encourages a stress-testing tool
      (\texttt{python-stress}), lets the model
      generate large inputs and check whether its
      solution timeouts. This tool functions the same as normal python tool but uses a different name for better behavior tracking. Correct trajectories are collected, filtered and downsampled for SFT.
    \item \textbf{Iteration~2 (SFT).}
        Distill the collected tool-use trajectories from Iteration~1
      via SFT back into CWM-SFT.
\end{enumerate}

Finally, we launch the RL sweep under different input thresholds, starting with the SFT checkpoint produced from Iteration~2.

\subsubsection{Iteration 0: Tool-Use RL}
\label{sec:appendix-iter0}

We start from CWM-SFT 32B, which can reason in a
\texttt{<think>} block but has never used a Python tool
for code that reads from stdin.
We run RL with Dr.GRPO on CodeContests/train (12\,275 problems) with two tools available:
\texttt{python} (execute a code snippet) and
\texttt{python-io} (execute code that requires stdin).
The prompt includes a small number of few-shot examples
demonstrating \texttt{python-io} usage.
The reward function gives hierarchical partial credit
for tool invocation: full reward ($+1$) for passing
with \texttt{python-io} in the first turn,
reduced credit ($+0.8$, $+0.5$, $0$) for passing with
progressively less desirable tool usage, and
asymmetric penalties on failure that are smaller when
the model at least attempted tool use.
This encourages the model to explore tool use early in
training.

\subsubsection{Iteration 1: Distillation and
  Stress-Testing RL (SFT $\to$ RL)}
\label{sec:appendix-iter1}

\paragraph{SFT phase.}
From the Iteration 0 RL run, we collect trajectories
that (i)~contain at least one \texttt{python-io} call
and (ii)~lead to a correct final answer.
We apply diversity filtering to balance across dialog
lengths, tool usage patterns, and task coverage,
reducing the dataset from 804k to 276k examples.
We also filter to a maximum of 32\,768 tokens per
trajectory and reformat reasoning tokens from the
\texttt{<think>} tag into the native
\texttt{<|reasoning\_thinking\_start|>} format reserved for CWM-SFT.
SFT trains for 1\,600 steps at learning rate
$10^{-6}$, starting again from CWM-SFT but not the RL
checkpoint, with batch size 2 and sequence length
32\,768 across 32 H100 nodes (32 DP $\times$ 8 TP).
Linear layers use rowwise FP8 quantization; all other
computations are in bf16.

\paragraph{RL phase.}
Starting from the SFT checkpoint, we run RL on
\codecontestsplus (1$\times$) with the same tool-use
environment as iteration~0 plus the
\texttt{python-stress} tool, which lets the model
write an input generator alongside its solution and
reports wall-clock execution time.
The reward function shifts the shaping target from
\texttt{python-io} to \texttt{python-stress}:
full reward ($+1$) for passing when
\texttt{python-stress} was used anywhere in the
trajectory, reduced credit ($+0.5$) for passing with
only generic tool use, and a small penalty ($-0.2$)
for failing despite using \texttt{python-stress}
versus full penalty ($-1$) for failing without any
tool use.
This encourages the model to adopt stress testing
for efficiency verification before submission.

\subsubsection{Iteration 2: Distillation and RL Sweep}

\paragraph{SFT phase.} We adopt the same SFT strategy as Iteration~1. Instead of merge the SFT dataset
gathered from Iteration~0 and Iteration~1, we use exclusively the dataset from Iteration~1 which supersets
the behavior presented in Iteration~0. We use the same hyperparameters and training steps as in Iteration~1.

\paragraph{RL phase.} Starting from the SFT checkpoint, we run RL on
\codecontestsplus (1$\times$) with the same tool-use
environment as iteration~1. The reward function becomes binary which grades only the final correctness. This is the RL run where we sweep different test coverage threshold.

\subsubsection{Shared RL Hyperparameters}

All three RL phases use Dr.GRPO~\citep{liu2025drgrpo} with DAPO-style~\citep{yu2025dapo}
asymmetric clipping
($\epsilon_{\mathrm{low}}{=}0.2$,
$\epsilon_{\mathrm{high}}{=}0.25$) and
zero-advantage skipping, with no KL penalty and no
reference model.
We use AdamW with learning rate
$1.4 \times 10^{-7}$, constant schedule with
400-step warmup, and gradient norm clipped at 0.1.
Each prompt produces 16 rollouts at temperature 1.0,
with a maximum of 6 tool calls per trajectory.
The trainer-to-worker ratio is 1:1, with the model
updated every 4 worker steps.
Training uses 256 H100 GPUs
(32 DP $\times$ 8 TP) with FSDP in bf16,
torch.compile, and selective activation
checkpointing.
The RL phases differ in training data, reward
target, sequence length, and step count:
iteration~0 trains for 4\,000 steps on
CodeContests/train with
\texttt{python-io} reward shaping and sequence
length 131\,072;
iteration~1~and~2 trains for 8\,000 steps on
\codecontestsplus (1$\times$) with
\texttt{python-stress} reward shaping and binary pass-fail reward respectively and sequence
length 65\,536 (approximately 18 hours per run).

\paragraph{Observation.}
During iteration~0, we ran RL on both DMC/train and
\codecontestsplus.
The two training sets yield similar overall pass rates,
but different error profiles:
\codecontestsplus produces fewer timeouts and more
wrong answers.
This observation motivated the core investigation
of this paper (Section~\ref{sec:tradeoff}).

Appendix~\ref{sec:appendix-tool-examples} shows one representative rollout from each iteration.

\subsection{Compute Resources}
\label{sec:appendix-compute}

All experiments use NVIDIA H100 80\,GB GPUs.
Table~\ref{tab:compute} summarizes the hardware layout
and wall-clock time for each training configuration.
For the 7B, 32B pure-reasoning, and 32B agentic-coding
settings, we sweep the input-length threshold
$k \in \{2,\ldots,6,\infty\}$ across six independent
runs; total GPU-hours for those rows reflect this
$\times 6$ factor.
The 32B tool-call pipeline requires a bootstrapping
phase (2 RL + 2 SFT runs) before the final
threshold sweep. The bootstrapping rows are single
runs. The Iter~0~RL uses 131k sequence length while the Iter~1~RL uses 64k sequence length, which compensates
the training steps and results in roughly the same wall-clock training time.

\begin{table}[h]
\centering
\caption{Compute resources per training configuration.
DP = data parallelism, TP = tensor parallelism.
Bootstrapping rows (marked $\dagger$) are single runs
that build the starting policy for the tool-call
threshold sweep.}
\vspace{0.2em}
\label{tab:compute}
\small
\begin{tabular}{lcccccl}
\toprule
Setting & Model & GPUs & DP$\times$TP & Steps & Hours/run & Total GPU-h \\
\midrule
\multicolumn{7}{l}{\textit{Threshold-swept settings ($\times 6$ runs)}} \\
7B pure reasoning RL & 7B & 32 & 32$\times$1 & 6\,000 & $\sim$19 & $\sim$3\,648 \\
32B pure reasoning RL & 32B & 256 & 32$\times$8 & 8\,000 & $\sim$44 & $\sim$67\,584 \\
32B tool-call RL & 32B & 256 & 32$\times$8 & 8\,000 & $\sim$18 & $\sim$27\,648 \\
32B agentic coding RL & 32B & 256 & 32$\times$8 & 4\,000 & $\sim$16 & $\sim$24\,576 \\
\midrule
\multicolumn{7}{l}{\textit{Tool-call bootstrapping$^\dagger$ (2 RL + 2 SFT, single runs)}} \\
Iter 0 RL & 32B & 256 & 32$\times$8 & 4\,000 & $\sim$18 & $\sim$4\,608 \\
Iter 1 SFT & 32B & 256 & 32$\times$8 & 1\,600 & $<$1 & $\sim$256 \\
Iter 1 RL & 32B & 256 & 32$\times$8 & 8\,000 & $\sim$18 & $\sim$4\,608 \\
Iter 2 SFT & 32B & 256 & 32$\times$8 & 1\,600 & $<$1 & $\sim$256 \\
\bottomrule
\end{tabular}
\end{table}

\clearpage
\section{Frontier Validation and Robustness}
\label{app:frontier-validation}

This appendix checks that the frontier does not come from mislabeled failures (Appendix~\ref{sec:appendix-reclassification}) or from training noise (Appendix~\ref{sec:appendix-seed-variance}), and collects further evidence for the frontier across settings and over training (Appendix~\ref{app:frontier-evidence}).

\subsection{Error Taxonomy Validation with Extended Reclassification}
\label{sec:appendix-reclassification}

A submission that exceeds resource limits does not
necessarily implement a correct algorithm; it may
contain a bug masked by the resource limit.
To assess the sensitivity of our results to this
ambiguity, we apply a two-stage reclassification
protocol to every submission originally labeled as
a resource-limit failure.
We detail the protocol in
Section~\ref{sec:appendix-reclassification}, with
Figure~\ref{fig:reclassification-comparison} showing
the comparison of the extended reclassification and
the raw sandbox signal used in the main text.
Section~\ref{sec:appendix-hidden-oom} reveals one
interesting finding: a substantial fraction of
timed-out submissions are in fact memory-exhaustion
failures masked by the time limit.
Whether a slow submission will eventually exhaust
memory or time is fundamentally difficult to
determine without running it to completion, which
we believe explains why most existing work
aggregates both failure modes into a single
timeout category.

\subsubsection{Reclassification Protocol}

\paragraph{Stage 1: extended re-evaluation.}
We re-execute each submission labeled as resource-limit failure with a
$100\times$ time budget.
Under this extended budget, submissions fall into
one of four categories:
(i)~\emph{all-pass}, confirming a correct but
algorithmically slow implementation;
(ii)~\emph{wrong answer}, revealing a correctness
bug that was masked by the original time limit;
(iii)~\emph{out of memory}, indicating that the
submission is memory-intensive and the original
timeout pre-empted the true failure mode;
or (iv)~\emph{still timeout}, remaining ambiguous.

Table~\ref{tab:reeval-breakdown} reports the
breakdown for the high-coverage training condition (all test
cases included) across all three inference settings.
A substantial fraction of resource-limit failures
are confirmed correct under the extended budget
(27-40\% across settings and thresholds), while
out-of-memory errors account for 5-14\% of
originally timed-out submissions.
Table~\ref{tab:reeval-breakdown} reports the
full breakdown.

\begin{table}[H]
\centering
\small
\caption{Stage~1 re-evaluation outcomes
  (\% of timed-out submissions) under a
  $100\times$ time budget.
  Each column corresponds to one test-coverage
  threshold; each row group to one inference
  setting.
  ``Total'' is the number of submissions
  originally labeled as resource-limit failures.}
\label{tab:reeval-breakdown}
\setlength{\tabcolsep}{4pt}
\begin{tabular}{l l rrrrrr}
\toprule
& & $\le\!10^2$
& $\le\!10^3$
& $\le\!10^4$
& $\le\!10^5$
& $\le\!10^6$
& all \\
\midrule
\multicolumn{8}{l}{%
  \textit{Pure Reasoning}} \\
& Total
  & 153 & 167 & 157 & 158 & 124 & 104 \\
& All-pass
  & 37.3\% & 31.1\% & 36.9\% & 39.9\%
  & 40.3\% & 34.6\% \\
& Wrong answer
  & 3.3\% & 5.4\% & 4.5\% & 0.6\%
  & 2.4\% & 2.9\% \\
& Out of memory
  & 9.2\% & 4.8\% & 7.6\% & 7.6\%
  & 5.6\% & 8.7\% \\
& Still timeout
  & 47.1\% & 53.9\% & 43.3\% & 50.0\%
  & 46.8\% & 49.0\% \\
& Other
  & 3.3\% & 4.8\% & 7.6\% & 1.9\%
  & 4.8\% & 4.8\% \\
\midrule
\multicolumn{8}{l}{%
  \textit{Tool Use}} \\
& Total
  & 140 & 139 & 126 & 109 & 99 & 89 \\
& All-pass
  & 22.1\% & 17.3\% & 15.9\% & 22.0\%
  & 33.3\% & 36.0\% \\
& Wrong answer
  & 4.3\% & 3.6\% & 5.6\% & 3.7\%
  & 3.0\% & 11.2\% \\
& Out of memory
  & 10.0\% & 7.2\% & 14.3\% & 11.9\%
  & 9.1\% & 10.1\% \\
& Still timeout
  & 57.9\% & 68.3\% & 58.7\% & 58.7\%
  & 51.5\% & 41.6\% \\
& Other
  & 5.7\% & 3.6\% & 5.6\% & 3.7\%
  & 3.0\% & 1.1\% \\
\midrule
\multicolumn{8}{l}{%
  \textit{Agentic Coding}} \\
& Total
  & 139 & 131 & 114 & 89 & 80 & 69 \\
& All-pass
  & 26.6\% & 26.7\% & 31.6\% & 25.8\%
  & 27.5\% & 27.5\% \\
& Wrong answer
  & 2.9\% & 4.6\% & 2.6\% & 3.4\%
  & 3.8\% & 5.8\% \\
& Out of memory
  & 7.2\% & 6.1\% & 9.6\% & 6.7\%
  & 10.0\% & 13.0\% \\
& Still timeout
  & 56.8\% & 55.7\% & 49.1\% & 60.7\%
  & 53.8\% & 52.2\% \\
& Other
  & 6.5\% & 6.9\% & 7.0\% & 3.4\%
  & 5.0\% & 1.4\% \\
\bottomrule
\end{tabular}
\end{table}

\paragraph{Stage 2: LLM judgment.}
For submissions that still exceed resource limits
or sandbox error after Stage~1 (the ``still timeout'',
``out of memory'' and ``Other'' rows), we query a
strong model, Claude-4.5-Sonnet
with multiple generations and majority
voting to binary classify whether the implementation is
logically correct but algorithmically naive, or
whether it contains a correctness bug.
Table~\ref{tab:reclass-final} reports the
fraction of resource-limit failures reclassified as
correctness failures after combining both stages,
across all settings and thresholds.
Crucially, the number of successful submissions
is unchanged in every condition: reclassification
only moves submissions from the resource-limit
failure category to correctness failure, never
to success.

\begin{table}[H]
\centering
\small
\caption{Final reclassification outcome after
  both stages (\% of original resource-limit
  failures).
  The two rows sum to 100\% for each condition.}
\label{tab:reclass-final}
\setlength{\tabcolsep}{4pt}
\begin{tabular}{l l rrrrrr}
\toprule
& & $\le\!10^2$
& $\le\!10^3$
& $\le\!10^4$
& $\le\!10^5$
& $\le\!10^6$
& all \\
\midrule
\multicolumn{8}{l}{%
  \textit{Pure Reasoning}} \\
& Reclass.\ correctness failure
  & 17.6\% & 16.2\% & 15.9\% & 17.1\%
  & 15.3\% & 20.2\% \\
& Remain resource-limit failure
  & 82.4\% & 83.8\% & 84.1\% & 82.9\%
  & 84.7\% & 79.8\% \\
\midrule
\multicolumn{8}{l}{%
  \textit{Tool Use}} \\
& Reclass.\ correctness failure
  & 5.0\% & 10.1\% & 13.5\% & 15.6\%
  & 6.1\% & 22.5\% \\
& Remain resource-limit failure
  & 95.0\% & 89.9\% & 86.5\% & 84.4\%
  & 93.9\% & 77.5\% \\
\midrule
\multicolumn{8}{l}{%
  \textit{Agentic Coding}} \\
& Reclass.\ correctness failure
  & 20.1\% & 27.5\% & 27.2\% & 19.1\%
  & 22.5\% & 31.9\% \\
& Remain resource-limit failure
  & 79.9\% & 72.5\% & 72.8\% & 80.9\%
  & 77.5\% & 68.1\% \\
\bottomrule
\end{tabular}
\end{table}

\paragraph{Robustness of the trade-off.}
Figure~\ref{fig:reclassification-comparison}
compares the trade-off curve before and after
reclassification across all three inference
settings and all six test-coverage thresholds.
In each panel, raw sandbox labels (solid) and
reclassified labels (dashed) trace nearly
identical frontiers.
The reclassified curve shifts slightly toward
higher correctness failure rates, as expected
from unmasking hidden bugs, but the shape and
slope of the frontier are preserved.
Since the relative orders are preserved,
the two labeling schemes agree closely and can be seen as a global shift
of the trade-off curve,
we use raw sandbox labels throughout the main
paper to avoid the computational cost of
running the full reclassification pipeline at
every evaluation checkpoint.

\subsubsection{Hidden Memory Exhaustion in Timeout Labels}
\label{sec:appendix-hidden-oom}

Standard competitive-programming evaluation
pipelines enforce both a time limit and a memory
limit, but when a submission exceeds the time
limit first, the sandbox reports a timeout
regardless of whether the code would also have
exhausted memory.
Memory exhaustion is therefore aggregated into
the timeout category, making it invisible in
downstream analysis.

Our extended re-evaluation (Stage~1 of the
reclassification protocol,
Appendix~\ref{sec:appendix-reclassification})
surfaces this hidden failure mode.
Under the $100\times$ time budget, 5\% to 14\%
of originally timed-out submissions across all
settings and thresholds terminate with an
out-of-memory error rather than a wrong answer
or a continued timeout
(Table~\ref{tab:reeval-breakdown}).

These submissions implement memory-intensive
algorithms (e.g., building large lookup tables
or storing full search trees) whose true failure
mode is masked by the shorter time limit.
Since competitive-programming benchmarks are
the primary training and evaluation environment
for code RL, this conflation means that a
non-trivial fraction of what existing work treats
as ``slow but possibly correct, i.e., timeout error'' code is in fact
code that would crash from memory exhaustion
even given enough time.
Disentangling time and memory failures may be
important for designing reward signals that
distinguish algorithmic efficiency from
implementation correctness.

\subsection{Seed Variance Analysis}
\label{sec:appendix-seed-variance}

To quantify the statistical reliability of our main findings,
we repeat the pure-reasoning RL runs for each of the three
coverage thresholds
(Keep~$\leq 10^{3}$, Keep~$\leq 10^{5}$, and All tests)
with three independent random seeds
at both the 7B and 32B model scales,
and report mean~$\pm$~one standard deviation.

\begin{figure}[h]
\centering
\begin{subfigure}[t]{0.48\linewidth}
    \centering
    \includegraphics[width=\linewidth]{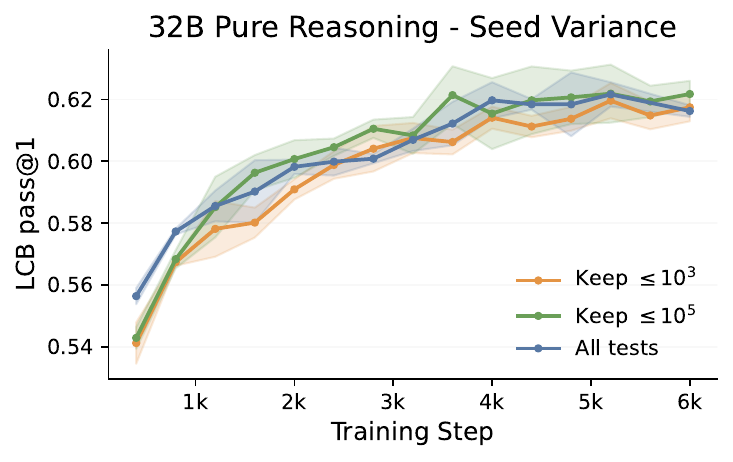}
\end{subfigure}
\hfill
\begin{subfigure}[t]{0.48\linewidth}
    \centering
    \includegraphics[width=\linewidth]{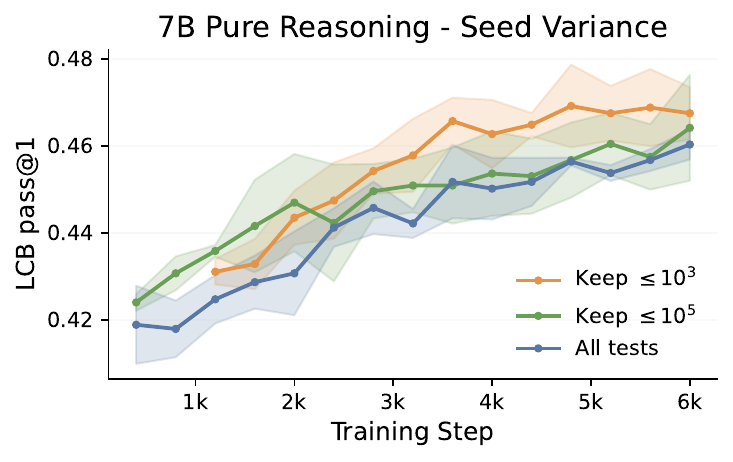}
\end{subfigure}
\caption{%
    Pass@1 on LCB-hard across training steps for
    32B (left) and 7B (right) pure-reasoning runs.
    Shaded regions denote $\pm 1$ standard deviation
    over three seeds.
}
\label{fig:seed-variance-pass1}
\end{figure}

\paragraph{Pass@1 stability.}
Figure~\ref{fig:seed-variance-pass1} shows
the LCB pass@1 training curves across seeds.
At the 32B scale, all three thresholds exhibit
tight variance bands throughout training,
with standard deviations below 0.5\%
at step~6k
(e.g.\ Keep~$\leq 10^{3}$:
$61.7 \pm 0.5\%$;
All tests: $61.6 \pm 0.2\%$).
The 7B runs show moderately larger variance
($\sim$0.7-1.0\%), consistent with smaller models
being more sensitive to initialization,
but the relative ordering among thresholds is preserved
across all seeds.

\begin{figure}[h]
\centering
    \includegraphics[width=0.75\linewidth]{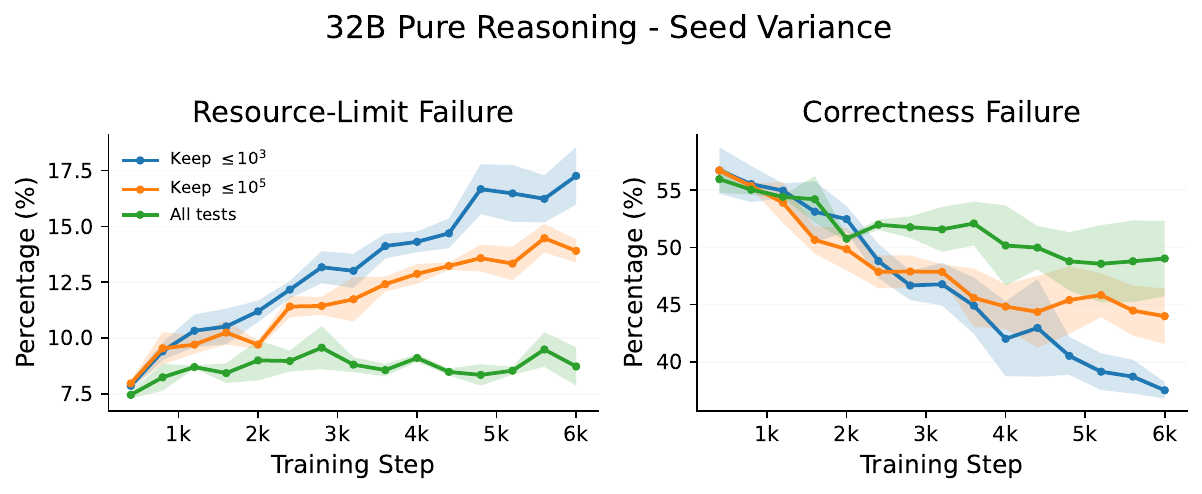}
    \includegraphics[width=0.75\linewidth]{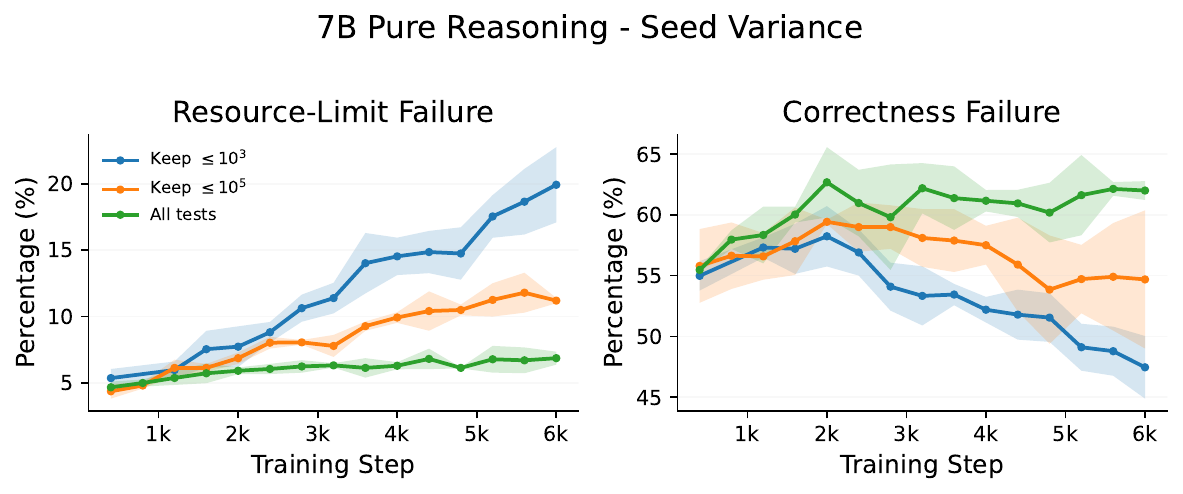}
    \centering
\caption{%
    Resource-limit failure and correctness failure
    rates on LCB-hard for 32B (up) and
    7B (bottom).
    Shaded regions denote $\pm 1$ standard deviation
    over three seeds.
    The trade-off between failure modes is consistent
    across seeds at both scales.
}
\label{fig:seed-variance-failures}
\end{figure}

\paragraph{Failure-mode stability.}
Figure~\ref{fig:seed-variance-failures} decomposes
the failure rate on LCB-hard into
\emph{resource-limit failures} and
\emph{correctness failures}.
The optimization-correctness trade-off reported
in the main text is robust across seeds:
stricter filtering (Keep~$\leq 10^{3}$) consistently
yields higher resource-limit failure rates
(32B: $17.3 \pm 1.3\%$ vs.\ $8.7 \pm 0.9\%$
for All tests at step~6k)
while reducing correctness failure rates
($37.5 \pm 0.7\%$ vs.\ $49.0 \pm 3.3\%$).
The same pattern holds at the 7B scale.
In all cases the inter-seed standard deviation
is substantially smaller than the
inter-threshold differences,
confirming that the reported trade-offs
reflect genuine training dynamics rather
than seed-dependent fluctuations.

\subsection{Extended Frontier Evidence}
\label{app:frontier-evidence}

\subsubsection{Per-Setting Extrapolation Plots}
\label{app:per-setting-extrapolation}
\label{app:7b-hard-extrapolation}

Figures~\ref{fig:teaser} and~\ref{fig:extrapolation-strip} in
the main text show the extrapolative continuation as compact
panels. This subsection presents standalone plots for
each setting and scale with full interpolation, extrapolation,
and format-error detail.

\begin{figure}[H]
    \centering
    \includegraphics[width=0.9\linewidth]{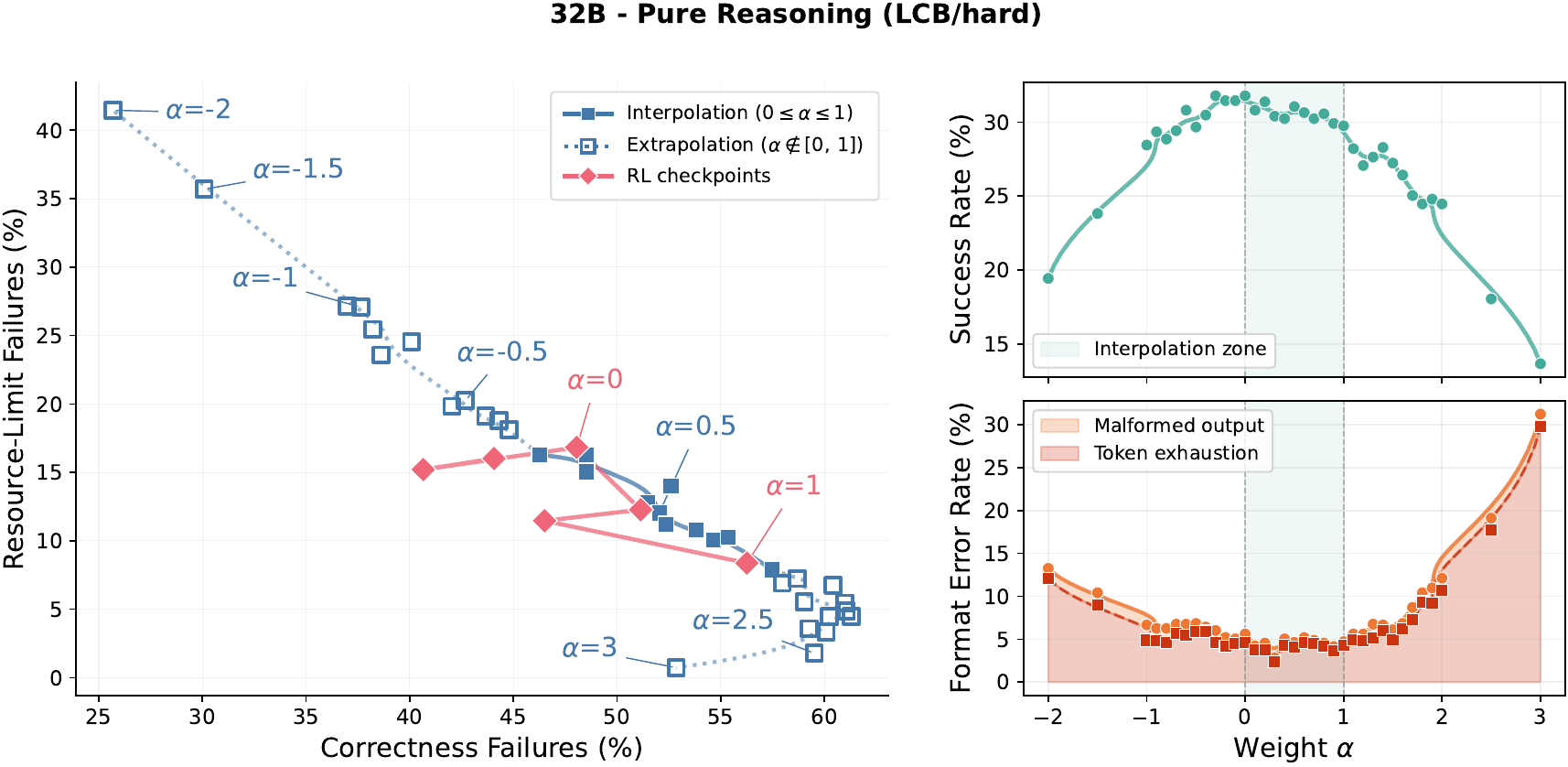}
    \caption{32B pure reasoning on LCB/hard.}
    \label{fig:32b-pr-extrapolation}
\end{figure}

\begin{figure}[H]
    \centering
    \includegraphics[width=0.9\linewidth]{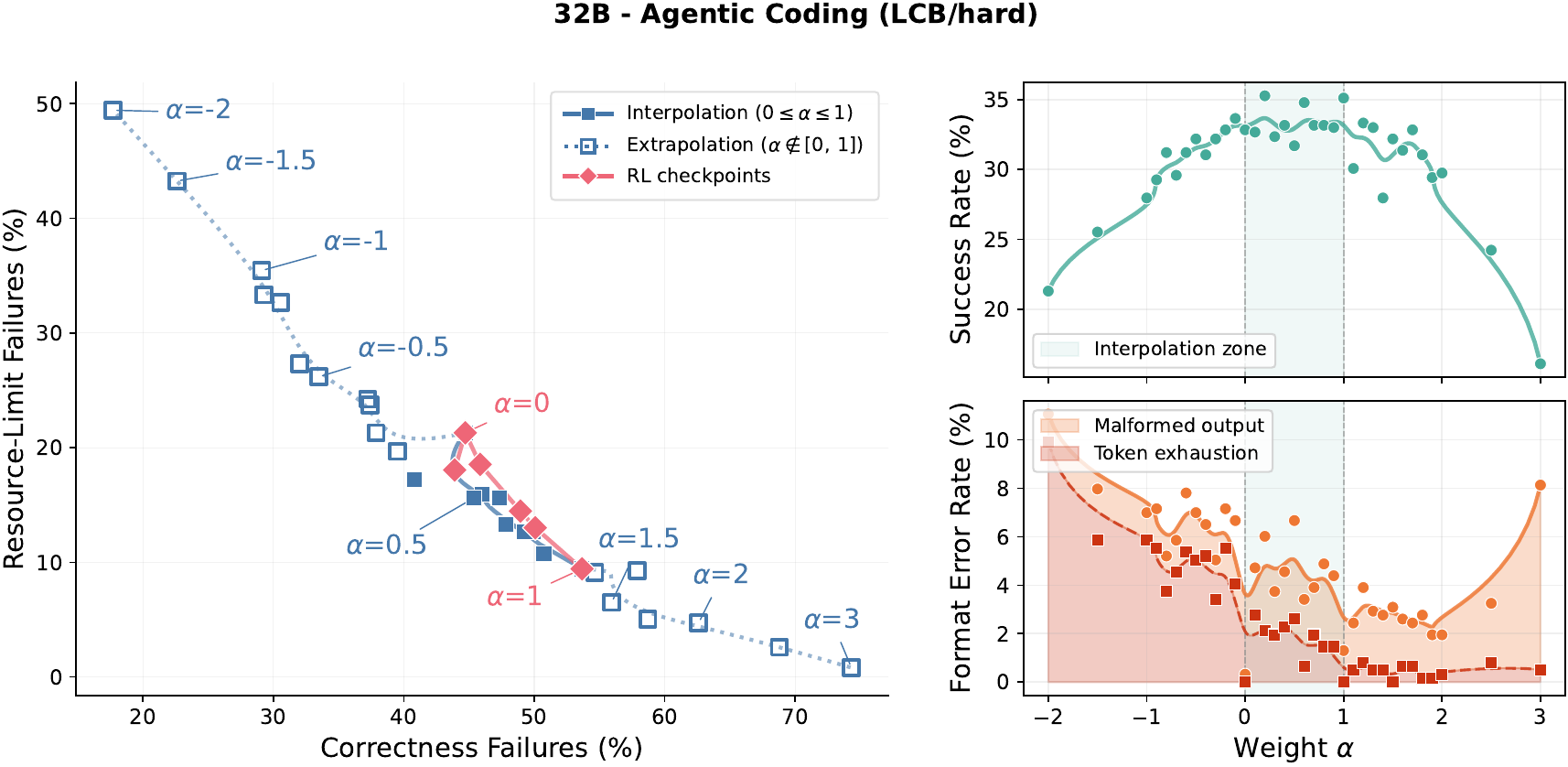}
    \caption{32B agentic coding on LCB/hard.}
    \label{fig:32b-agentic-extrapolation}
\end{figure}

\begin{figure}[H]
    \centering
    \includegraphics[width=0.9\linewidth]{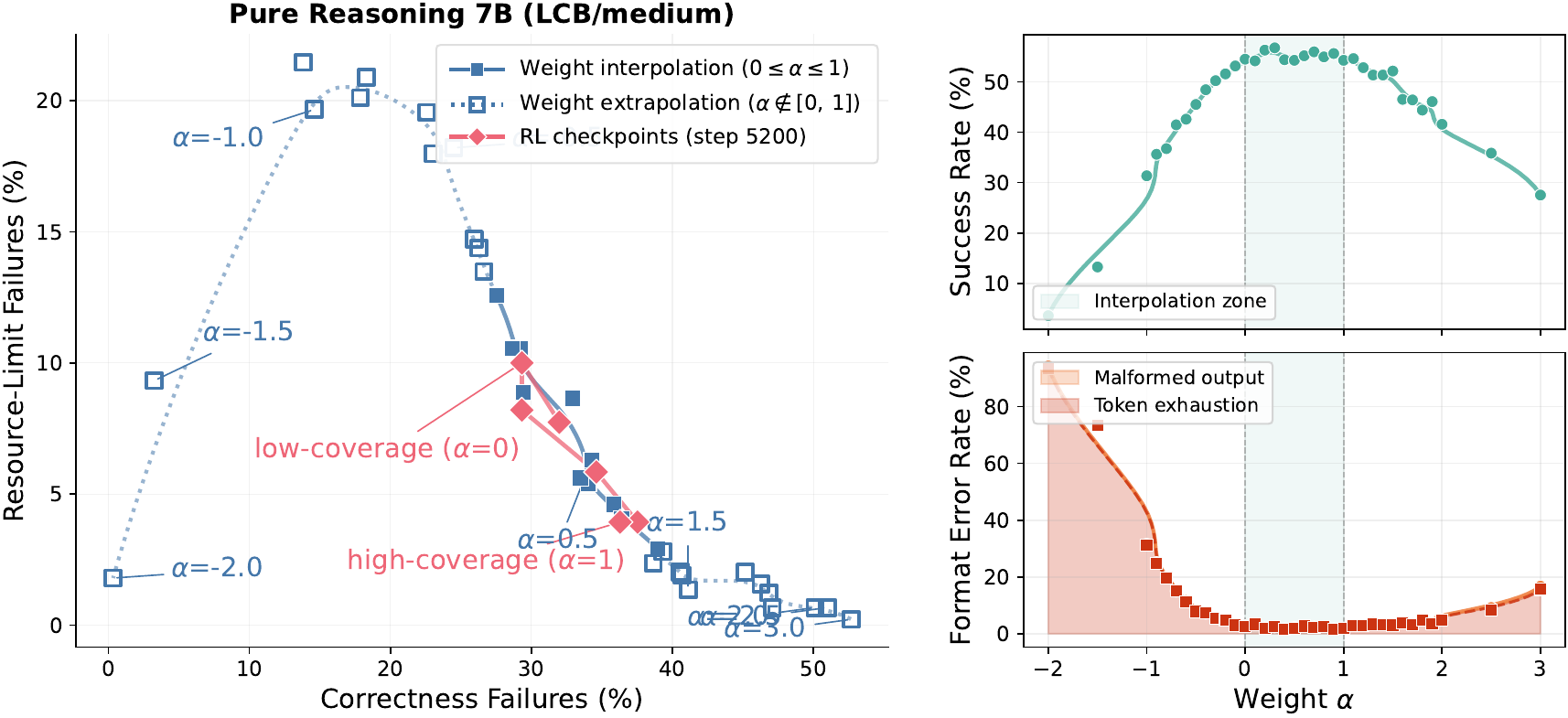}
    \caption{7B pure reasoning on LCB/medium (capability-matched split).}
    \label{fig:7b-medium-extrapolation}
\end{figure}

\begin{figure}[H]
    \centering
    \includegraphics[width=0.9\linewidth]{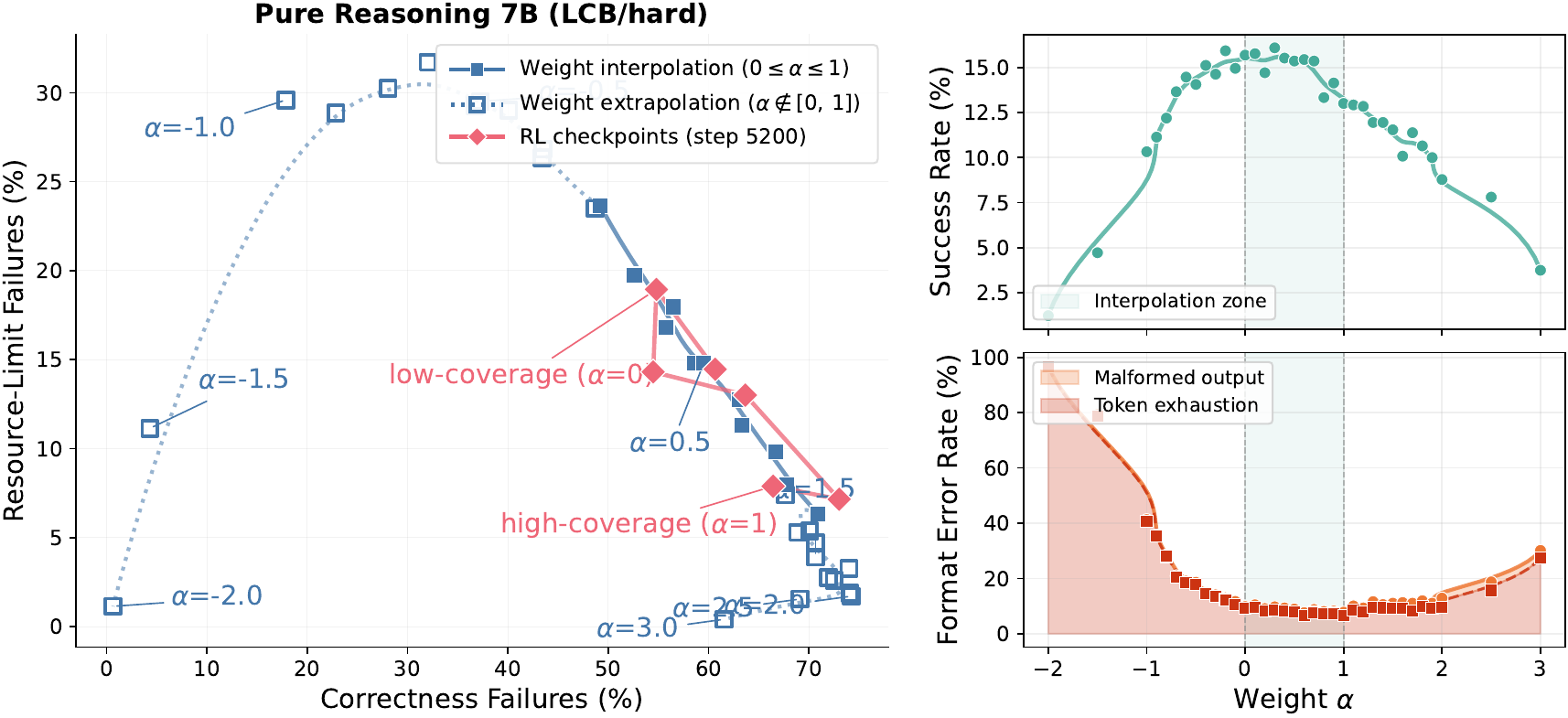}
    \caption{7B pure reasoning on LCB/hard. The stable extrapolation range is narrower than LCB/medium, with format errors dominating earlier on the low-$\alpha$ side.}
    \label{fig:7b-hard-extrapolation}
\end{figure}

\subsubsection{Full Pairwise Transition Matrices}
\label{sec:appendix-transition-matrices}

\begin{figure}[h]
    \centering
    \includegraphics[width=0.4\linewidth]{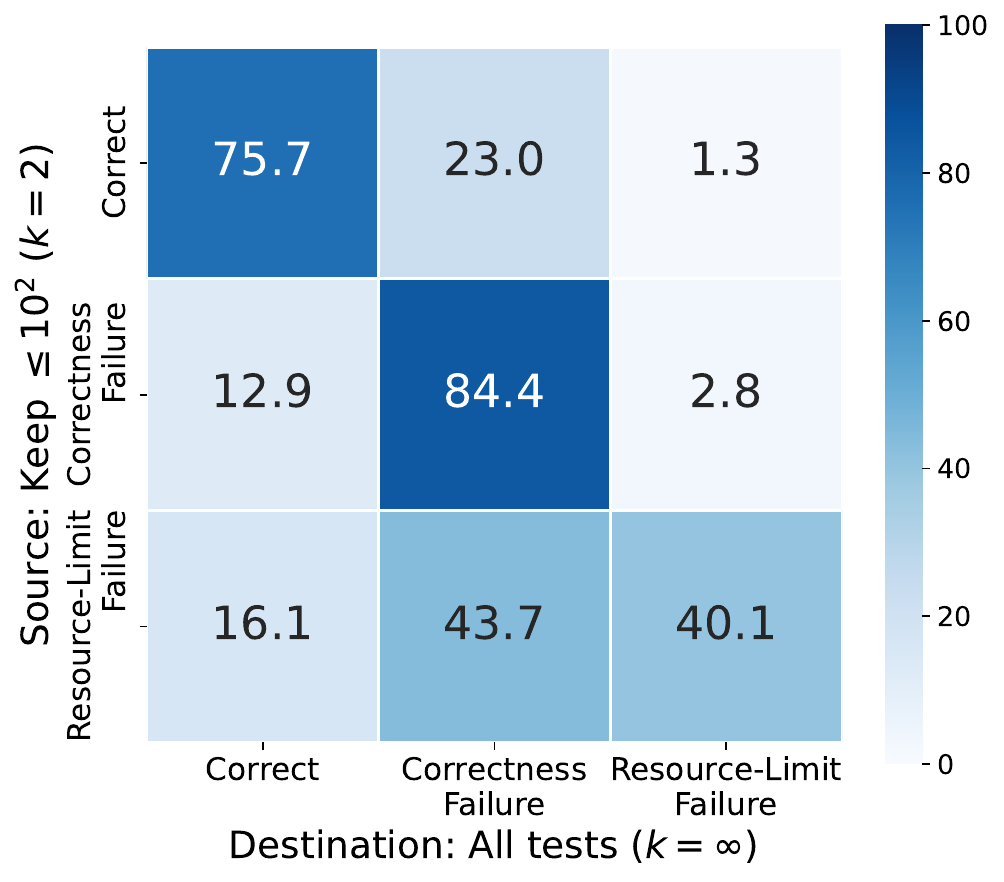}
    \caption{Task-level transition matrix from
      $k{=}2$ (keep $\leq 10^2$) to
      $k{=}\infty$ (all tests) on 123 hard problems. Taken from pure-reasoning setting in late stage of the training (7.2k steps).
      Rows are normalized to 100\%.
      Of resource-limit failures under low coverage,
      43.7\% become correctness failures under high
      coverage, ruling out an aggregation artifact.}
    \label{fig:transition-matrix}
\end{figure}

\begin{figure}[h]
    \centering
    \includegraphics[width=\linewidth]{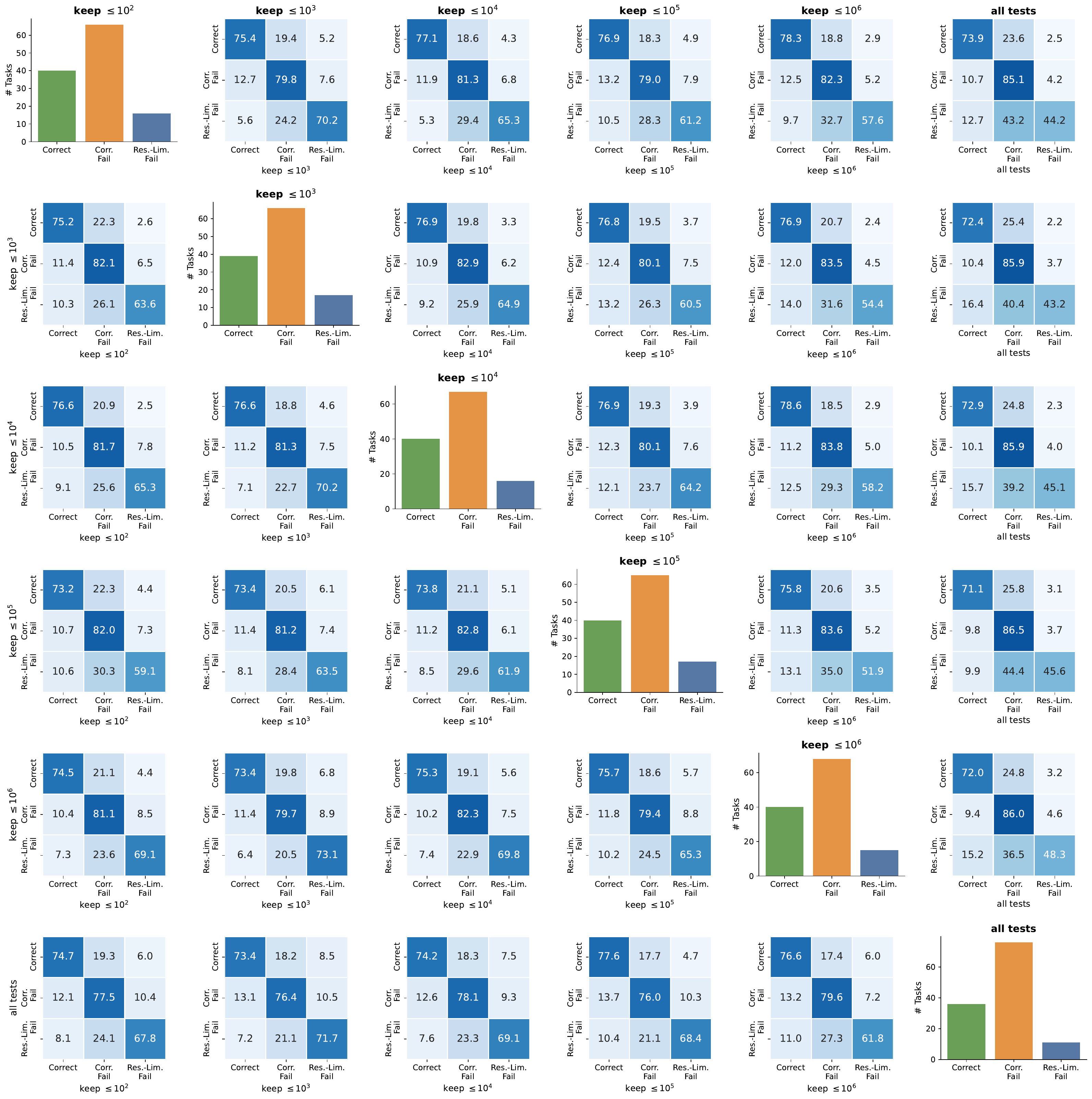}
    \caption{Full grid of pairwise task-level transition matrices on hard problems.
    Diagonal panels show per-condition status distributions. Taken from the pure-reasoning setting in early stage of the training (3.6k steps).
    Off-diagonal cells are row-normalized to 100\%.}
    \label{fig:transition-grid}
\end{figure}

Figure~\ref{fig:transition-grid} shows the complete set of
task-level transition matrices between all pairs of
test-coverage conditions on hard problems.
The pattern is consistent throughout: the dominant
off-diagonal transitions are between resource-limit failures
and correctness failures, confirming that the trade-off
operates at the individual task level across all condition
pairs.

\subsubsection{Emergence of the Trade-off Curve During Training}

\begin{figure}[H]
    \centering
    \begin{subfigure}[ht]{0.32\linewidth}
        \includegraphics[width=\linewidth]{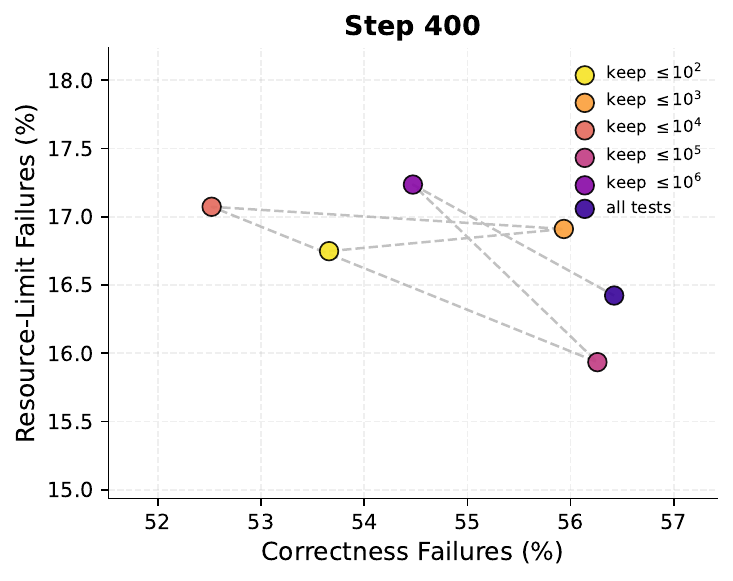}
    \end{subfigure}
    \hfill
    \begin{subfigure}[ht]{0.32\linewidth}
        \includegraphics[width=\linewidth]{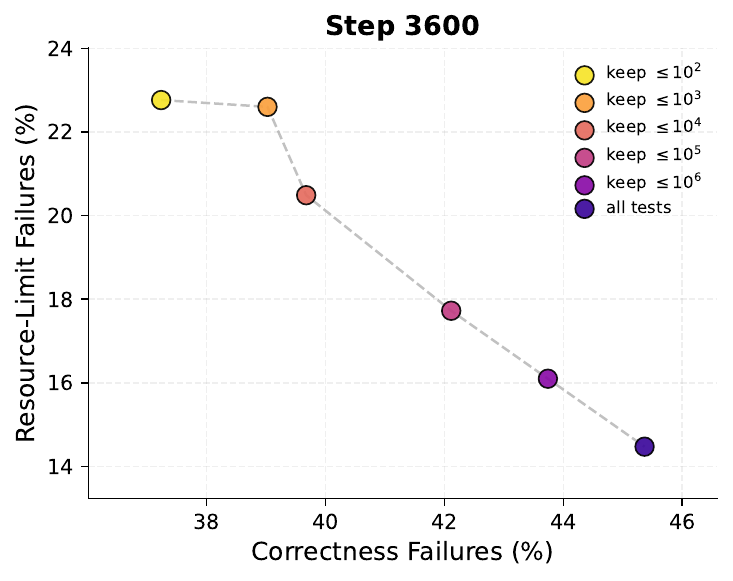}
    \end{subfigure}
    \hfill
    \begin{subfigure}[ht]{0.32\linewidth}
        \includegraphics[width=\linewidth]{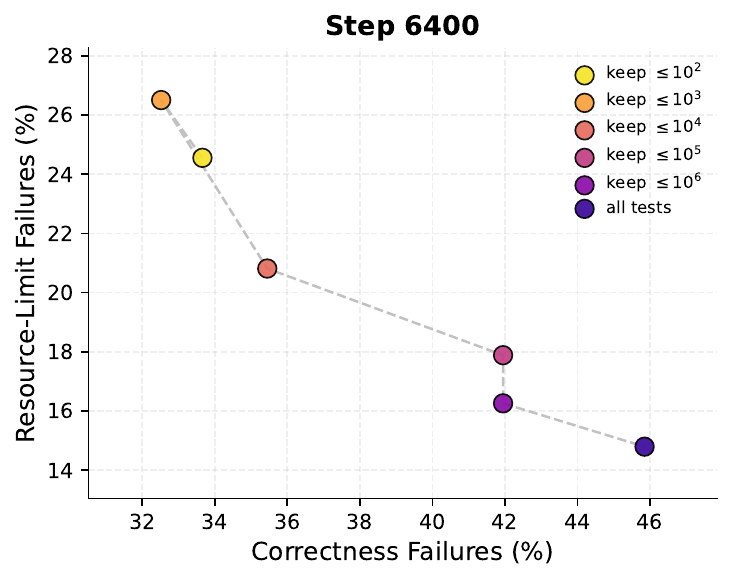}
    \end{subfigure}
    \caption{Emergence of the trade-off curve during RL training. At step 400, all six test-coverage
    conditions are clustered together. By step 3600, they have spread into a clear trade-off curve.
    By step 6400, the separation widens further.}
    \label{fig:pareto-emergence}
\end{figure}

The trade-off curve emerges progressively during RL training.
At step 400 (Figure~\ref{fig:pareto-emergence}, left), all six conditions cluster in a
tight region of the (Correctness Failure, Resource-Limit Failure) plane.
By step 3600 (center), the runs have diverged along a clear trade-off curve,
with the ordering matching the test-coverage threshold monotonically.
By step 6400 (right), the separation widens further.
This divergence occurs relatively early in training and then stabilizes,
suggesting that RL quickly discovers the point on the frontier dictated by its
test-coverage level.

\subsubsection{Training Dynamics}

\begin{figure}[h]
    \centering
    \includegraphics[width=\linewidth]{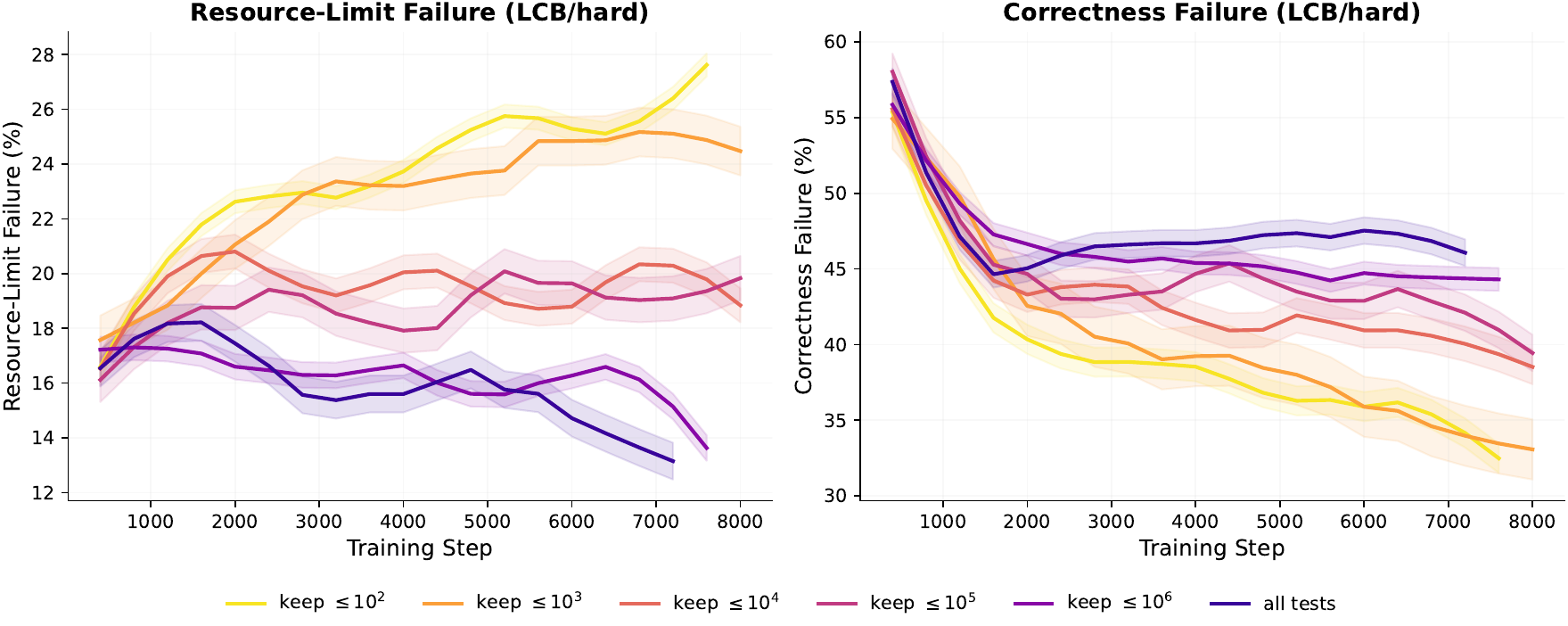}
    \caption{Extended training dynamics showing resource-limit failure and correctness failure
    counts on hard problems vs.\ training step on LCB/hard.}
    \label{fig:step-metrics-extended}
\end{figure}

Figure~\ref{fig:step-metrics-extended} shows the
resource-limit failure and correctness failure counts over an
extended training horizon on 32B pure-reasoning setting.
The separation between conditions is sustained at later
steps, indicating that the trade-off is not a transient
phenomenon of early training.

\newpage

\section{Inference-Time Ensemble Details}
\label{app:ensemble-details}

\subsection{Full Solved-Set Overlap}
\label{app:anatomy-full}

\begin{figure}[h]
    \centering
    \includegraphics[width=0.7\linewidth]{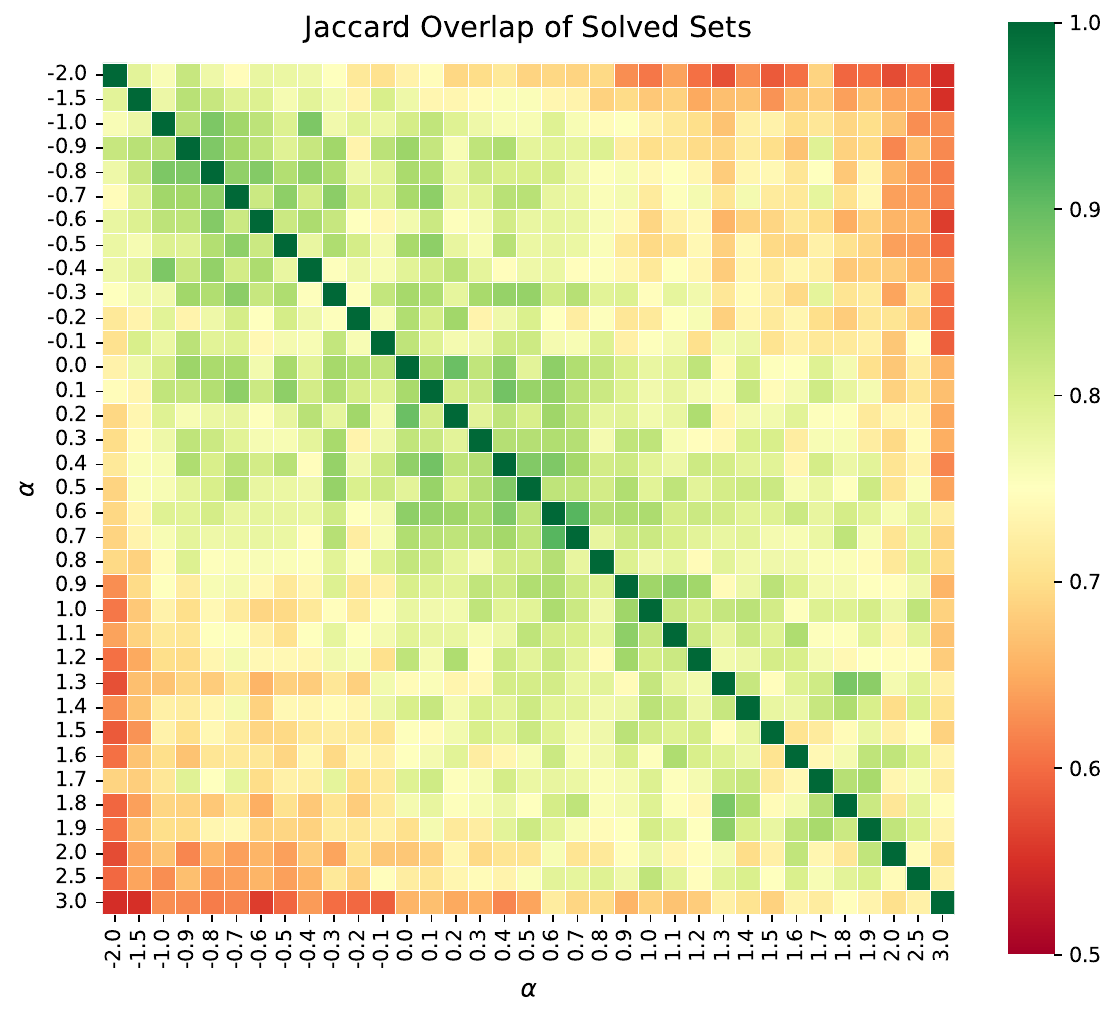}
    \caption{Full pairwise Jaccard similarity between
    solved sets across the extrapolation range
    $\alpha \in [-2, 3]$.
    Overlap decreases steadily with distance in
    $\alpha$ but remains above 0.54 even at the
    extremes, confirming gradual drift of the
    success set across the entire interpolation and
    extrapolation range.}
    \label{fig:jaccard-full}
\end{figure}

Section~\ref{sec:anatomy} summarizes how solved-set overlap varies
with distance in $\alpha$.
Figure~\ref{fig:jaccard-full} shows the full pairwise matrix
over $\alpha \in [-2, 3]$.
Overlap generally decreases with distance in $\alpha$.
Even at the extremes ($\alpha = -2$ vs.\
$\alpha = 3$), the Jaccard similarity remains above
0.54, indicating that over half of solved problems
are shared.

\subsection{Coverage Headroom Decomposition}
\label{app:coverage-headroom}

\begin{figure}[H]
\centering
\begin{minipage}{0.5\linewidth}
\centering
\includegraphics[width=\linewidth]{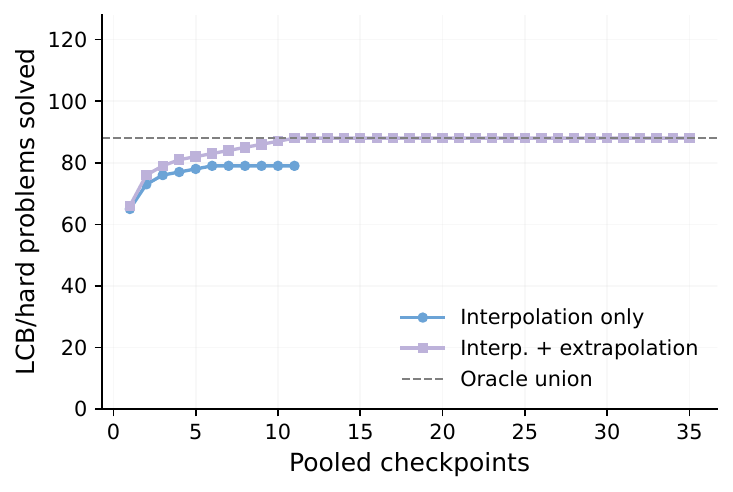}
\end{minipage}
\hfill
\begin{minipage}{0.42\linewidth}
\centering
\includegraphics[width=\linewidth]{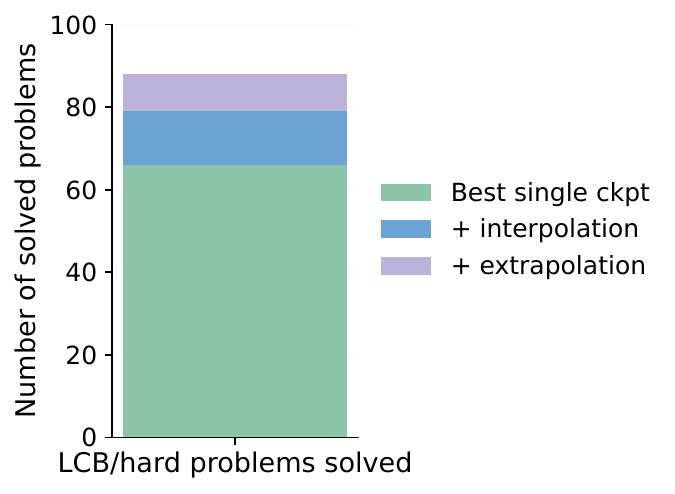}
\end{minipage}
\caption{\textbf{Left}: Cumulative coverage of LCB/hard (123 problems) as a function of the
number of checkpoints in the pool, under a greedy ordering. The
interpolation-only pool (blue) plateaus at $79$ problems; admitting
extrapolated checkpoints (red) continues to grow until it reaches $88$.
\textbf{Right}: Decomposition of the $22$-problem coverage gain over the best
single checkpoint into the contributions from interpolation and extrapolation.
Extrapolation accounts for $9$ of the $22$ extra problems and is the only path
to the final coverage band.}
\label{fig:coverage-headroom}
\end{figure}

Different checkpoints along the interpolation axis solve different subsets of hard problems, so pooling multiple checkpoints and selecting the best submission per problem can exceed the solve rate of every individual checkpoint.
We quantify this \emph{coverage headroom} with a greedy set-cover procedure: at each step, the checkpoint whose solved set adds the most new problems is appended to the pool, and we track the cumulative number of problems solved by at least one member.
Figure~\ref{fig:coverage-headroom} (left) shows the resulting curve for interpolation-only and interpolation-plus-extrapolation pools; the right panel decomposes the total gain over the best single checkpoint into contributions from each regime.

\subsection{Filtered Submission Budgets on LCB/v5}
\label{app:v5-filtered-pass-n}

\begin{figure}[ht]
    \centering
    \includegraphics[width=0.55\linewidth]{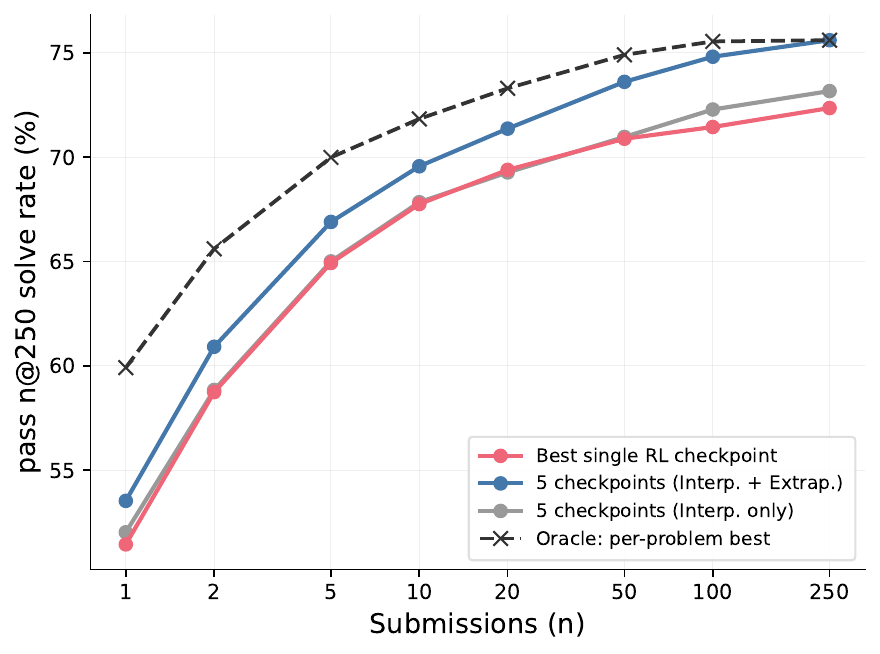}
    \caption{Pass $n$@$250$ with public-test filtering on LCB/v5 hard. The
    extrapolation-inclusive five-checkpoint pool is higher than both the best
    single checkpoint and the interpolation-only pool at each displayed
    submission budget $n$. This is an in-domain diagnostic of how filtering
    converts candidate diversity into gains under smaller submission budgets.}
    \label{fig:v5-pass-n-at-250}
\end{figure}

In practice, competitive programming judges allow
only a handful of submissions per problem, so
raw pass@$k$ with large $k$ overstates
the value of a diverse candidate pool.
Pass $n$@$k$ measures the solve rate when
$k$ candidates are sampled, filtered by
public tests, and the top $n$ are submitted.
Figure~\ref{fig:v5-pass-n-at-250} fixes
$k{=}250$ and sweeps $n$ from $1$ to $250$
on the LCB/v5 hard split.
The extrapolation-inclusive five-checkpoint pool
dominates the best single RL checkpoint
and the interpolation-only pool
at every submission budget.
At $n{=}1$, also called best@$k$, the gap is already roughly two
percentage points, showing that even a
single filtered submission benefits from
the broader coverage of the extrapolated
checkpoints.
The per-problem oracle selects
the best checkpoint for each problem
independently; the extrapolation pool
nearly closes this gap by $n{=}250$.

\section{Weight-Space Geometry}
\label{app:weight-geometry}

This appendix characterizes the weight-space
direction that separates the two RL endpoints
along three axes: the singular-value spectrum of
each layer's weight difference
(Section~\ref{app:svd-spectrum}), the distribution
of squared weight change across layers
(Section~\ref{app:layer-distribution}), and layer
stitching experiments that isolate causal
contributions to behavior
(Section~\ref{app:layer-stitching}).

The central finding is a dissociation.
RL modifies every layer by similar magnitudes and
in high-dimensional directions: the top 10
singular values explain only 10 to 19\% of the
variance within each layer
(Section~\ref{app:svd-spectrum}), and the total
squared weight change is nearly uniform across
layers 2 through 63
(Section~\ref{app:layer-distribution}).
Yet behavioral sensitivity is concentrated in the
middle of the network.
Layer stitching shows that the first and last
quarter of layers (0 to 15 and 48 to 63) are
nearly inert, while the middle half (16 to 47)
carries the full trade-off shift, with layers
24 to 31 accounting for roughly half of it
(Section~\ref{app:layer-stitching}).

\subsection{SVD Spectrum}
\label{app:svd-spectrum}

\begin{figure}[h]
    \centering
    \includegraphics[width=0.6\linewidth]{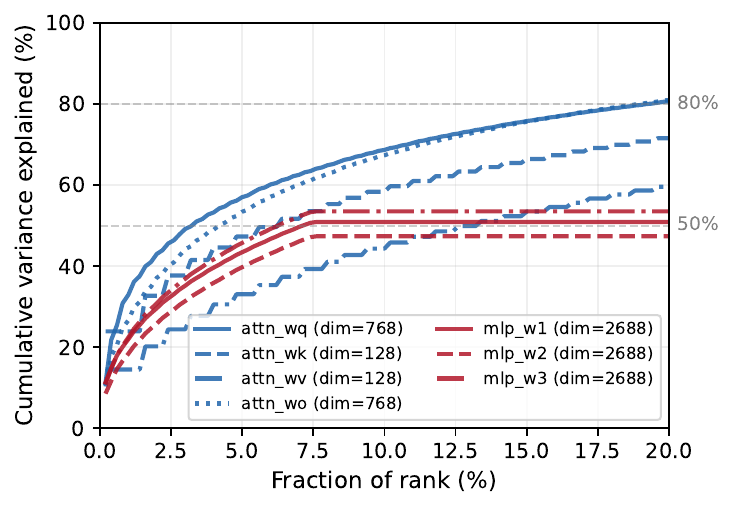}
    \caption{Cumulative variance explained by the top
    singular values of the weight difference
    $\delta = \theta_{\mathrm{high}} -
    \theta_{\mathrm{low}}$, averaged across layers and
    normalized by rank.
    Attention matrices concentrate variance faster
    than MLP matrices, but neither is low-rank: at
    20\% of rank, no component type exceeds 80\%
    explained variance.}
    \label{fig:svd-spectrum}
\end{figure}

For each weight matrix in the 32B 64-layer model, we
compute the SVD of
$\delta = \theta_{\mathrm{high}} -
\theta_{\mathrm{low}}$ and measure the cumulative
variance explained by the top $k$ singular values.
Figure~\ref{fig:svd-spectrum} shows the result,
normalized by the rank of each matrix.
No component type is low-rank: the top 10 singular
values of each layer's difference explain only
10 to 19\% of variance, and even at 20\% of rank
no component exceeds 80\% explained variance.

\subsection{Per-Layer Variance Distribution}
\label{app:layer-distribution}

\begin{figure}[h]
    \centering
    \includegraphics[width=0.7\linewidth]{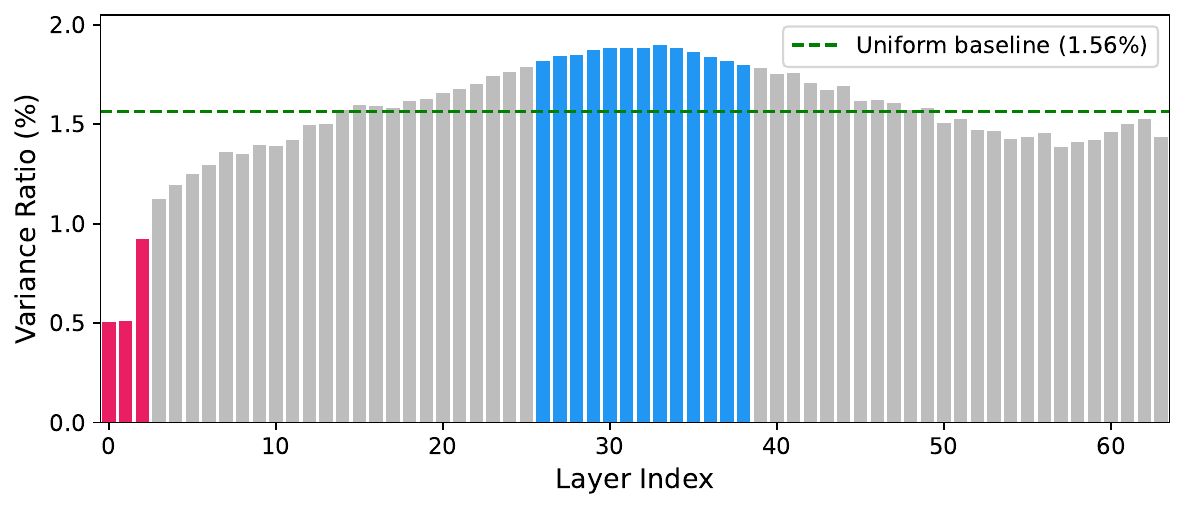}
    \caption{Per-layer variance ratio (fraction of total
    squared weight change).
    The dashed line marks the uniform baseline
    (1.56\% per layer).
    Layers 0-1 (red) are markedly below baseline;
    layers 24-36 (blue) are mildly above.
    The distribution is otherwise close to uniform.}
    \label{fig:varratio}
\end{figure}

We define the \emph{variance ratio} of layer $\ell$
as the fraction of total squared change
($\|\delta\|^2$ summed across all parameters)
attributable to that layer; under a uniform
distribution each of 64 layers would contribute
1.56\%.
Figure~\ref{fig:varratio} shows that the variance
ratio is nearly uniform across layers 2 through 63,
with only layers 0 and 1 markedly below baseline.

\subsection{Layer Stitching}
\label{app:layer-stitching}

The previous two analyses show that weight changes
are distributed and high-dimensional, but this does
not imply uniform causal importance.
We construct Frankenstein models by taking layers
$0$ through $k{-}1$ from $\theta_{\mathrm{high}}$
and layers $k$ through $63$ from
$\theta_{\mathrm{low}}$ (and vice versa), sweeping
cutpoints $k \in \{0, 16, 24, 32, 48, 64\}$.
We additionally test block swaps of layers 24 to 31
(8 layers) and 16 to 47 (32 layers).

\begin{figure}[h]
    \centering
    \includegraphics[width=0.7\linewidth]{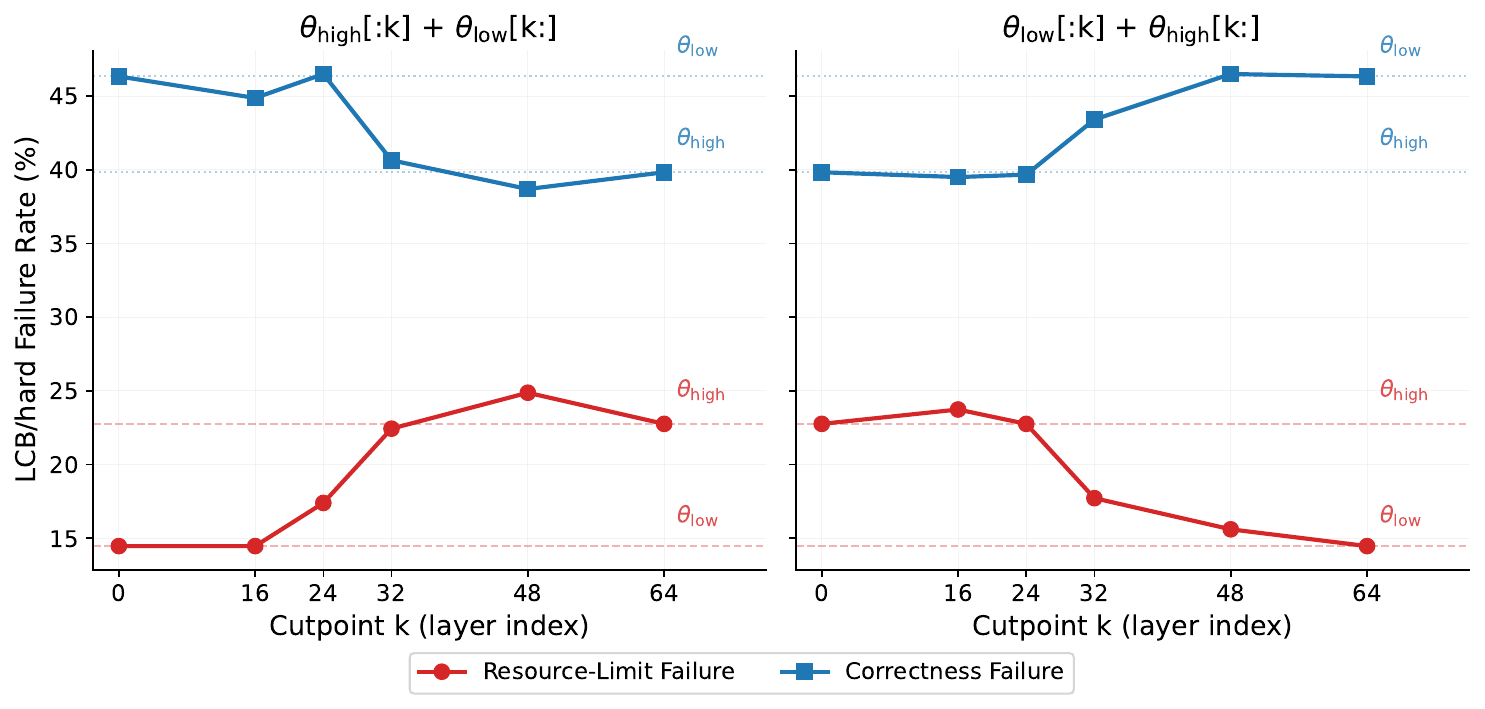}
    \caption{Layer stitching (Frankenstein models):
    resource-limit failure rate and correctness failure
    rate on LCB/hard as a function of cutpoint $k$.
    \textbf{Left}: layers $0$ through $k{-}1$ from
    $\theta_{\mathrm{high}}$, layers $k$ through $63$
    from $\theta_{\mathrm{low}}$.
    \textbf{Right}: the reverse direction.
    Dashed lines mark the endpoint levels.
    The behavioral transition is concentrated in
    the middle layers (roughly 16-48); edge layers
    are nearly inert.}
    \label{fig:layer-stitching}
\end{figure}

\begin{figure}[h]
    \centering
    \includegraphics[width=0.7\linewidth]{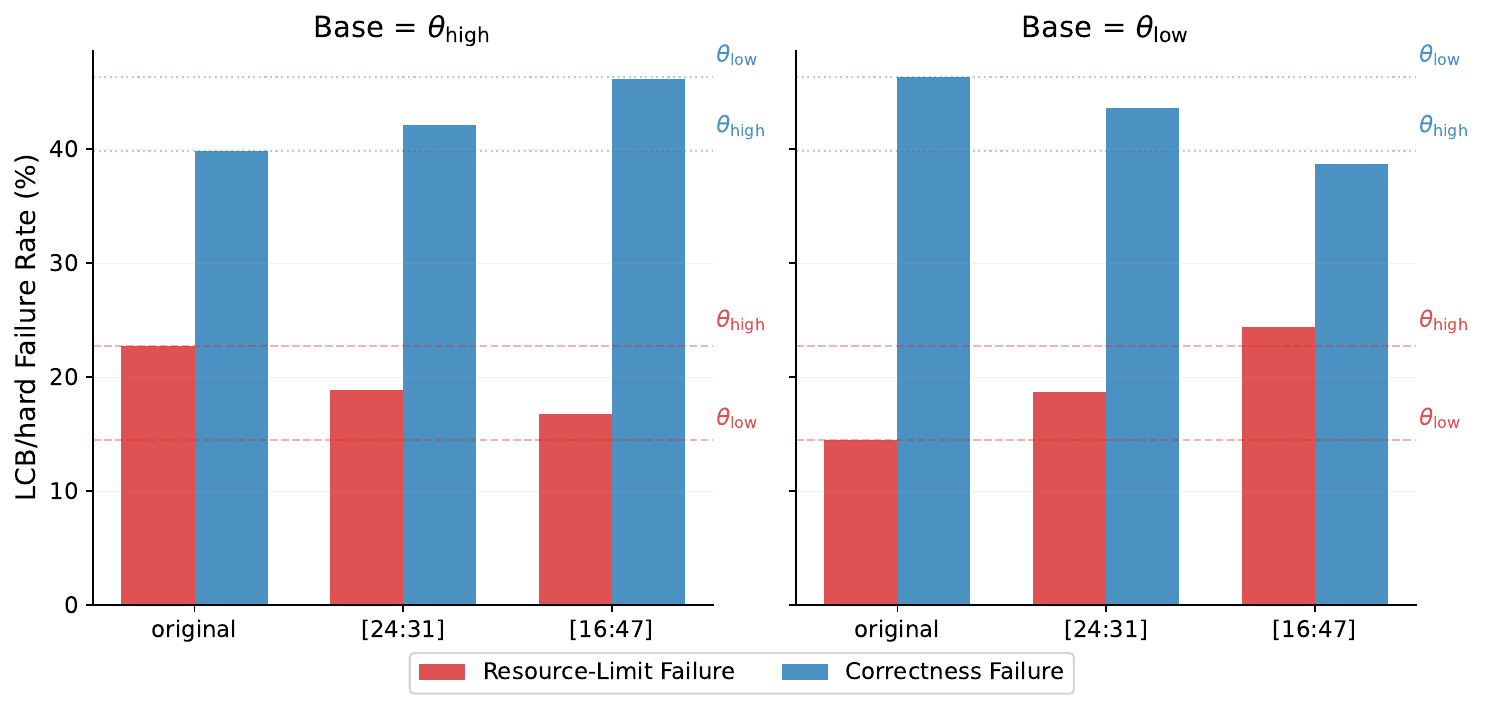}
    \caption{Block swap: replacing a contiguous range
    of middle layers in one endpoint with the
    corresponding layers from the other.
    \textbf{Left}: $\theta_{\mathrm{high}}$ base with
    $\theta_{\mathrm{low}}$ layers swapped in.
    \textbf{Right}: $\theta_{\mathrm{low}}$ base with
    $\theta_{\mathrm{high}}$ layers swapped in.
    Swapping layers 16-47 (half the model) nearly
    reproduces the full behavioral shift;
    swapping layers 24-31 (8 layers) produces a
    partial shift, confirming that a dense core of
    middle layers controls the trade-off.}
    \label{fig:block-swap}
\end{figure}

All stitched models remain on the trade-off curve,
with success rates in the 32.8 to 37.4\% band
(Figure~\ref{fig:layer-stitching}).
The first and last quarter of layers are nearly
inert: cutpoints at $k{=}16$ and $k{=}48$ shift
timeout counts by fewer than 6 relative to the
endpoints.
The middle layers carry the full behavioral shift.
Swapping layers 24 to 31 alone (12.5\% of the
network) accounts for roughly half the timeout
shift; swapping layers 16 to 47 reproduces or
slightly overshoots the full shift
(Figure~\ref{fig:block-swap}).

\clearpage
\section{Additional Experiments}
\label{app:additional-experiments}

This appendix reports experiments beyond the main line: a second weight-space frontier, between pass@1 and pass@$k$ (Appendix~\ref{app:auxiliary-frontier}), and two attempts to move the correctness-efficiency frontier, reward shaping (Appendix~\ref{app:reward-shaping}) and failure-conditioned training (Appendix~\ref{app:failure-conditioned}).

\subsection{A Second Weight-Space Frontier: Pass@1 vs Pass@$k$ Performance}
\label{app:auxiliary-frontier}

The main experiments use nested verifier coverage to create a controlled correctness-efficiency axis.
To test whether the weight-averaging geometry is specific to that axis, we ran an auxiliary experiment
between two RL checkpoints, trained from the same initialization (the SFT checkpoint from Qwen 2.5 7B as described in main text) but optimized for different sampling behavior.
The firstll is trained using standard GRPO which exhibits stronger pass@1, and the second is
trained under pass@8 objective as shown in~\citet{tang2025optimizinglanguagemodelsinference}. We form interpolation and extrapolation checkpoints:
\begin{equation*}
\theta_\alpha \;=\; \alpha\,\theta_{\mathrm{p@8}} \;+\; (1-\alpha)\,\theta_{\mathrm{p@1}},
\end{equation*}
and evaluate checkpoints for $\alpha \in [-0.5, 1.5]$ on LCB v5 + v6 (20240801-20250501), 454 problems in total, with temperature 1.0 and 200 samples per prompt.

\begin{figure}[h]
    \centering
    \includegraphics[width=\textwidth]{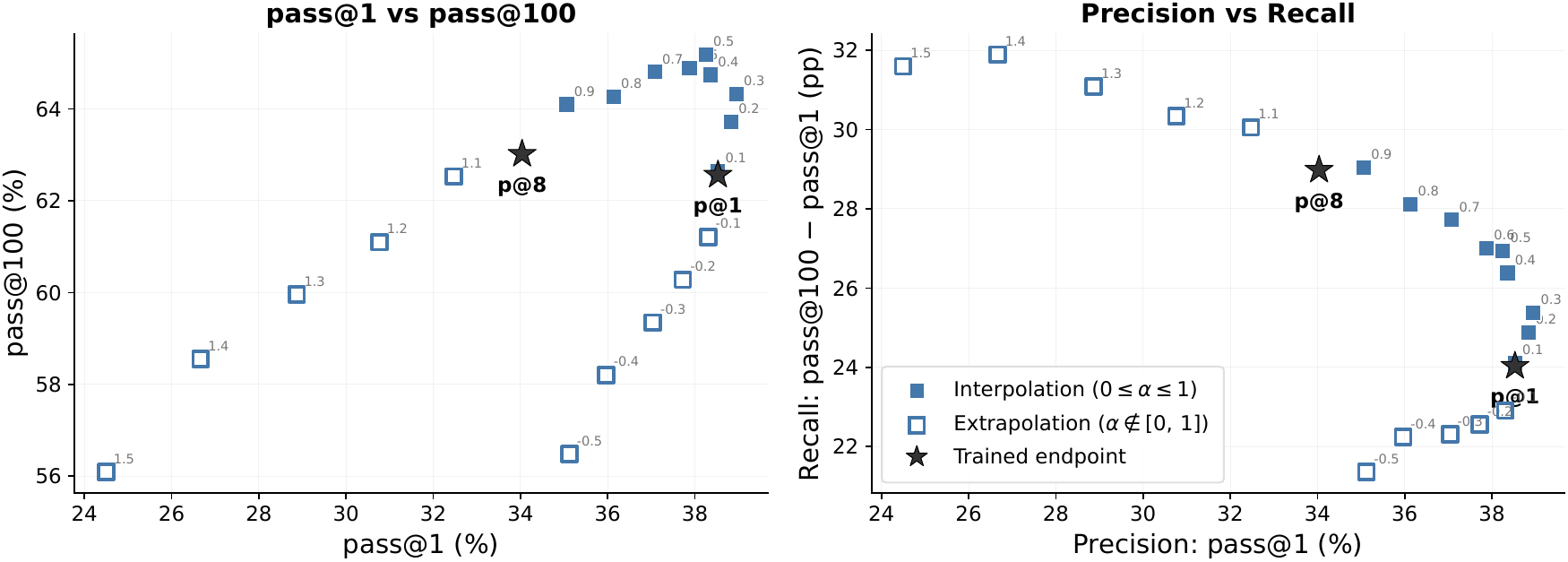}
    \caption{%
        \textbf{A second weight-space frontier: pass@1 vs pass@100.}
        We average a pass@1-oriented endpoint (p@1, $\alpha{=}0$) and a pass@8-oriented endpoint (p@8, $\alpha{=}1$).
        \textbf{Left:}~Interpolation traces a smooth curve in pass@1/pass@100 space;
        extrapolation remains locally meaningful but degrades at extreme values.
        \textbf{Right:}~In the Precision/Recall view, sampling headroom peaks
        in the interpolation region, indicating that interpolated checkpoints have the most
        untapped potential under larger sampling budgets.
    }
    \label{fig:auxiliary-scatter}
\end{figure}

\paragraph{Interpolation produces a smooth behavioral frontier.}
Figure~\ref{fig:auxiliary-scatter} shows that interpolation traces a smooth curve in pass@1/pass@100 space.
We follow the Precision/Recall framework~\citep{verine2025improvingdiversitylanguagemodels} which uses pass@1
as a proxy for Precision, and pass@100 - pass@1 as a measure for Recall.
Moving along the curve changes both the Precision (pass@1) and the Recall (pass@100$\,-\,$pass@1).
This confirms that the interpolation direction captures a meaningful behavioral axis even when the endpoints
differ by training objective rather than by verifier coverage.
Extrapolation is stable near the endpoints but gives diminishing returns: the most extreme extrapolated
checkpoints ($\alpha = -0.5$, $\alpha = 1.5$) degrade on both pass@1 and pass@100.

\begin{figure}[h]
    \centering
    \includegraphics[width=0.6\textwidth]{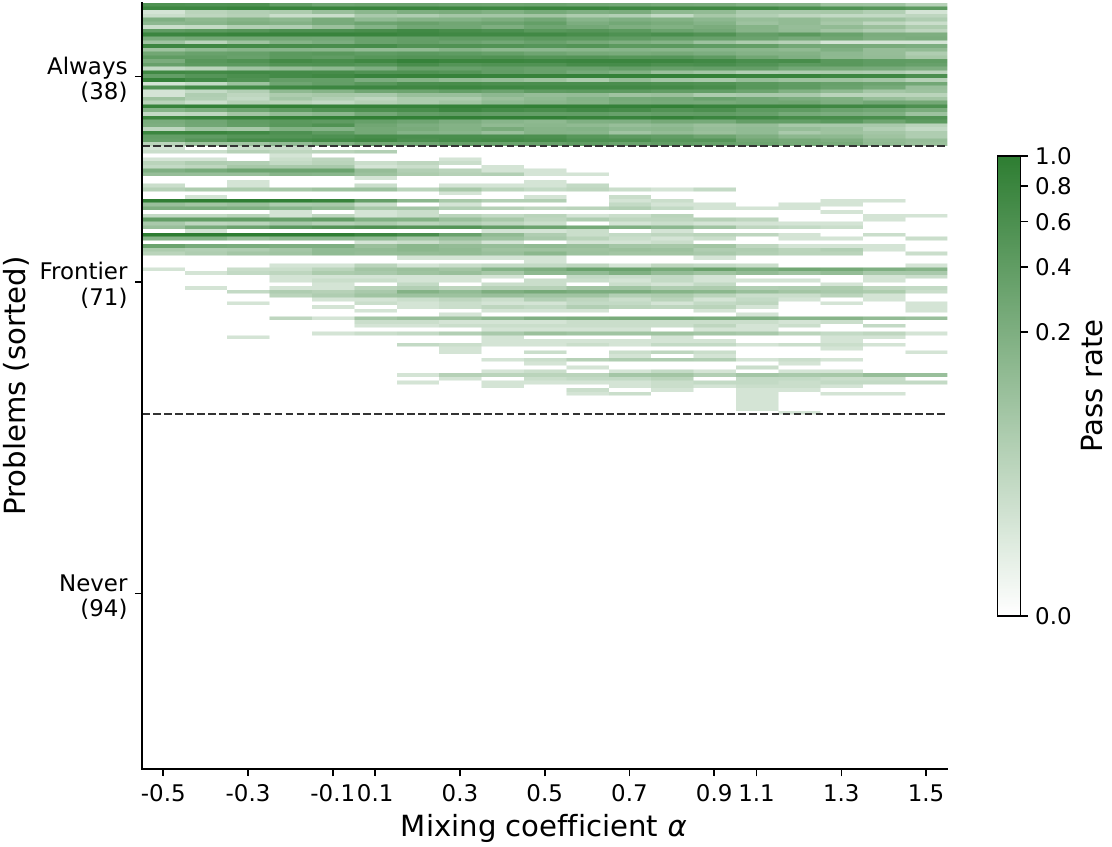}
    \caption{%
        Solved-set turnover along the interpolation axis of pass@1-trained checkpoint and pass@8-trained checkpoint (LCB/hard).
        Each row is one problem; columns are $\alpha$ values.
        Unlike the main-text heatmap (Figure~\ref{fig:anatomy}), which uses a linear color scale,
        we apply a power-law normalization ($\gamma{=}0.3$) to highlight problems solved at low pass rates.
        Of 203 hard problems, 38 are always solved, 94 are never solved, and 71 frontier problems change solvability with $\alpha$, providing the diversity that checkpoint banks exploit.
    }
    \label{fig:auxiliary-heatmap}
\end{figure}

\begin{figure}[h]
    \centering
    \includegraphics[width=\textwidth]{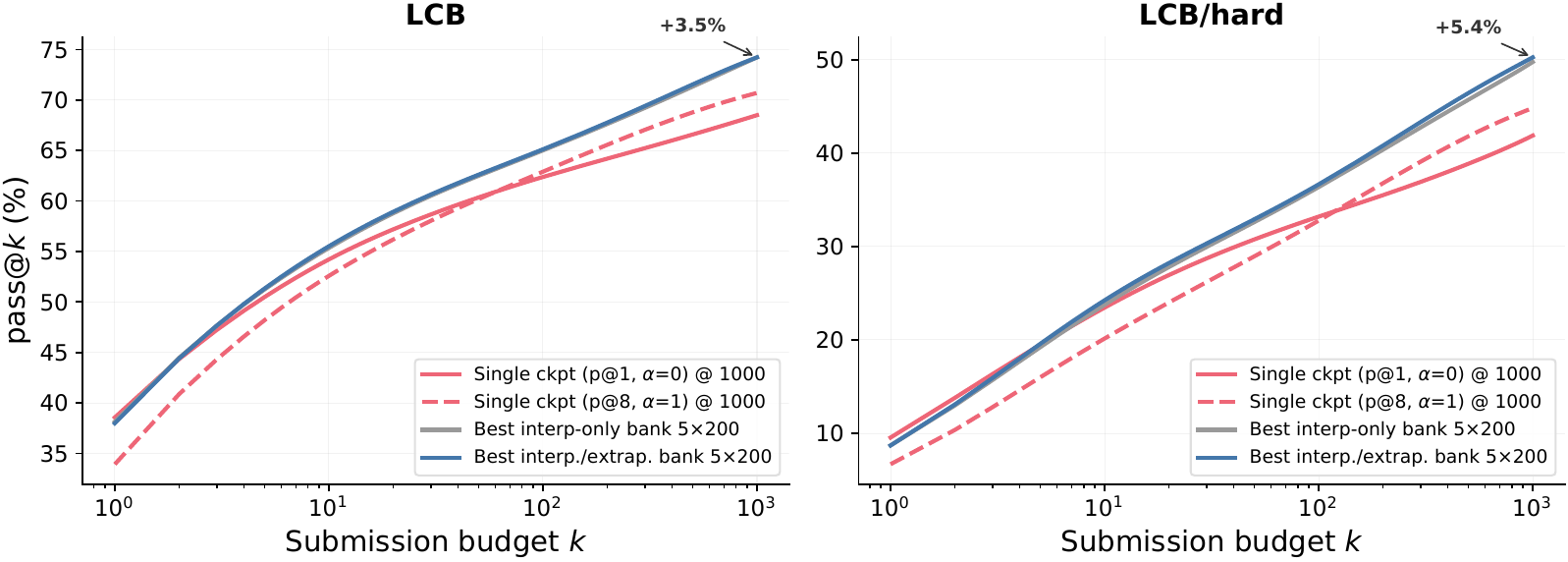}
    \caption{%
        \textbf{Matched-budget checkpoint banks on LCB.}
        \textbf{Left:}~On all problems, interpolation-only and interpolation/extrapolation banks overlap,
        both outperforming single-checkpoint sampling by +3.5\% at pass@1000.
        \textbf{Right:}~On hard problems, including an extrapolated checkpoint edges out
        interpolation-only by ${\sim}$0.5\%, trading medium-difficulty coverage for hard-problem diversity.
    }
    \label{fig:auxiliary-bank}
\end{figure}

\paragraph{Checkpoint banks exploit solved-set diversity.}
As in the verifier-coverage frontier, nearby values of $\alpha$ solve overlapping but distinct problem sets (Figure~\ref{fig:auxiliary-heatmap}).
This turnover can be exploited by checkpoint banks.
Figure~\ref{fig:auxiliary-bank} shows that five-checkpoint banks with 200 samples each outperform either single endpoint with 1000 samples at pass@$k$ for $k \geq 10$, reaching +3.5\% on all problems and +5.4\% on hard at pass@1000.
Including one extrapolated checkpoint ($\alpha{=}{-0.2}$) adds marginal diversity on hard problems: it uniquely solves 7 hard problems that no interpolation checkpoint covers.
Conversely, 12 problems are uniquely solved by interpolation (6~hard, 6~medium), so extrapolation trades medium-difficulty coverage for hard-problem diversity: precisely the frontier exploration that weight averaging exposes.

\paragraph{Scope and caveats.}
We treat this experiment as supportive but not primary evidence.
Unlike the nested verifier sweep, this setting is not a controlled sweep over a single monotone reward family,
and the endpoints differ in both objective and training stage.
Its role is to suggest that the frontier-navigation view can extend beyond correctness-efficiency,
not to replace the controlled evidence in the main paper.

\subsection{Reward Shaping Experiments}
\label{app:reward-shaping}
\label{sec:appendix-reward-shaping}

Reward shaping is the simplest static attempt to
soften the binary pass-fail reward. It assigns
partial credit for progress on the same test
suite, but it never reveals why a rollout
failed. It therefore changes the scalarization
of the same nested test outcomes without
changing the information available to learning.
The definitions below make this limitation
concrete.

In the reward-shaping experiment, the model
receives partial credit according to the largest
test-coverage threshold its submission passes.
We use the threshold set
$\mathcal{K} = \{2, 3, 4, 5, 6, \infty\}$.
The shaped reward is
\begin{align}
R_{\mathrm{shape}}(y) =
\begin{cases}
0   & \text{fails some test with input length} \le 10^2, \\
1/6 & \text{passes all tests with input length} \le 10^2, \\
2/6 & \text{passes all tests with input length} \le 10^3, \\
3/6 & \text{passes all tests with input length} \le 10^4, \\
4/6 & \text{passes all tests with input length} \le 10^5, \\
5/6 & \text{passes all tests with input length} \le 10^6, \\
1   & \text{passes all tests}.
\end{cases}
\end{align}

Because the thresholded rewards are nested, this
partial-credit objective is exactly the average of
the binary rewards from
Section~\ref{sec:weight-averaging}:
\begin{align}
R_{\mathrm{shape}}(y)
= \frac{1}{|\mathcal{K}|}
\sum_{k \in \mathcal{K}} R_k(y).
\end{align}
This identity is the main reason reward shaping
behaves like a reweighting of the same static
outcome family rather than a new source of
information.

\paragraph{Why shaping stays on the frontier: a KL-regularized lens.}
Our actual training uses Dr.GRPO without an explicit KL penalty.
To build intuition for why reward shaping does not escape the frontier, we analyze a simplified KL-regularized surrogate.  The argument is not a claim about GRPO dynamics; it is a closed-form calculation that illustrates why static shaping recombines existing outcome information rather than introducing new signal.

For a reward $R$ and reference policy $\pi_0$,
it is known that the KL-regularized optimal policy with KL penalty $\beta$ is
\begin{align}
\pi^*(y \mid x)
\propto
\pi_0(y \mid x)\,\exp\bigl(R(y)/\beta\bigr).
\end{align}
Write $\pi_k^*$ for the optimum under the
single-threshold binary reward $R_k$.
Substituting $R_{\mathrm{shape}}
= \frac{1}{|\mathcal{K}|}\sum_k R_k$ gives
\begin{align}
\pi_{\mathrm{shape}}^*(y \mid x)
&\propto
\pi_0(y \mid x)\,
\exp\Bigl(
\frac{1}{|\mathcal{K}| \beta}
\sum_{k \in \mathcal{K}} R_k(y)
\Bigr)
\notag\\
&\propto
\Bigl(
\prod_{k \in \mathcal{K}}
\pi_k^*(y \mid x)
\Bigr)^{1/|\mathcal{K}|}.
\end{align}
The optimal policy for the shaped reward is therefore the normalized geometric mean of the single-threshold optimal policy, or equivalently
the softmax of their arithmetic-mean logits.
Because $\pi_{\mathrm{shape}}^*$ is fully determined by the single-threshold optima $\{\pi_k^*\}$, reward shaping recombines the same nested outcome information rather than introducing a genuinely new training signal.
This provides intuition for why shaped rewards do not escape the frontier in our experiments, although we note that the geometric mean is not a convex combination in the probability simplex, so a formal convex-hull guarantee does not follow from this analysis alone.

This echoes the geometry of Section~\ref{sec:geometry}: weight averaging combines the same family of policies in parameter space, and reward shaping combines them in reward space.
These two can be connected if the model weights are under small perturbation that can be linearly approximated.

\begin{figure}[H]
    \centering
    \includegraphics[width=0.6\linewidth]{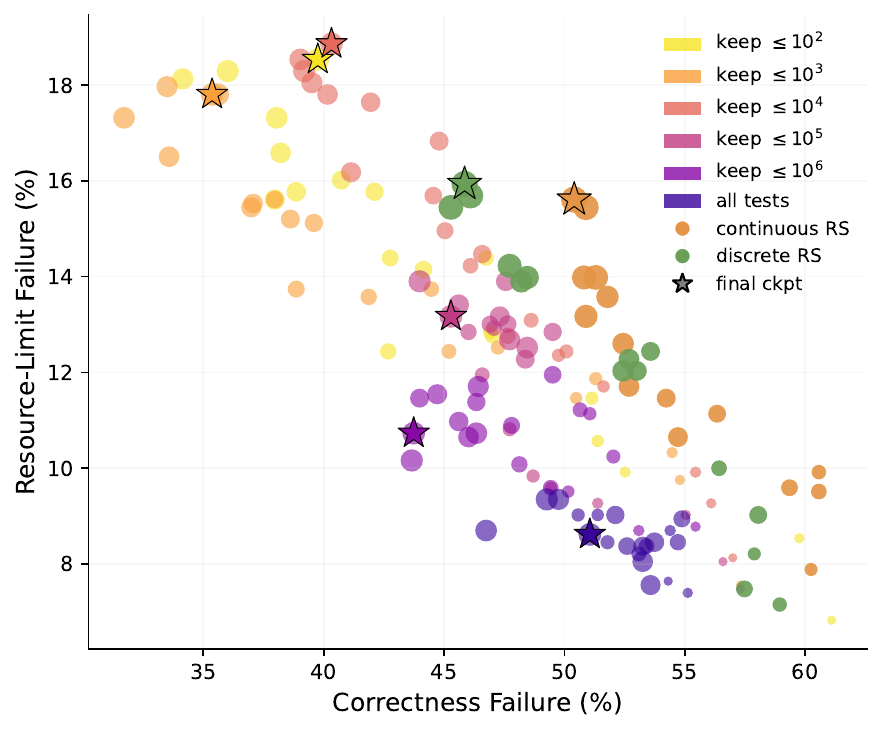}
    \caption{Reward shaping lands on the existing trade-off curve (32B pure reasoning, LCB/hard).
    Larger dots stand for later stage of training.
    Baseline single-threshold runs are shown as a color gradient from low coverage to high coverage. Both discrete (green) and continuous (orange) reward shaping checkpoints fall on the same frontier.
    Stars mark final checkpoints. Reward shaping, whether discrete or continuous, cannot escape the trade-off.}
    \label{fig:reward-shaping-trade-off}
\end{figure}

\paragraph{Continuous reward shaping.}
In addition to the discrete variant above, we
test a continuous variant that interpolates
linearly in log-space based on the smallest
failing test input.
Let $s(y)$ denote the character length of the
smallest test input on which submission $y$ fails.
The continuous shaped reward is
\begin{align}
R_{\mathrm{cont}}(y) =
\begin{cases}
1   & \text{passes all tests}, \\[4pt]
\displaystyle
\frac{\log_{10} s(y) - 2}{4}
     & 10^2 \le s(y) < 10^6, \\[4pt]
0   & s(y) < 10^2.
\end{cases}
\end{align}
This interpolates linearly in log-space from
$0$ (fails on the smallest inputs) to $1$
(passes everything).
Both variants remain static functions of the same test outcomes.
As Figure~\ref{fig:reward-shaping-trade-off}
shows, both land on the existing trade-off
curve, with continuous reward shaping slightly worse.
We posit that the scalarization does not matter and what matters is that neither variant reveals why a submission failed.

\begin{figure}[H]
    \centering
    \includegraphics[width=\linewidth]{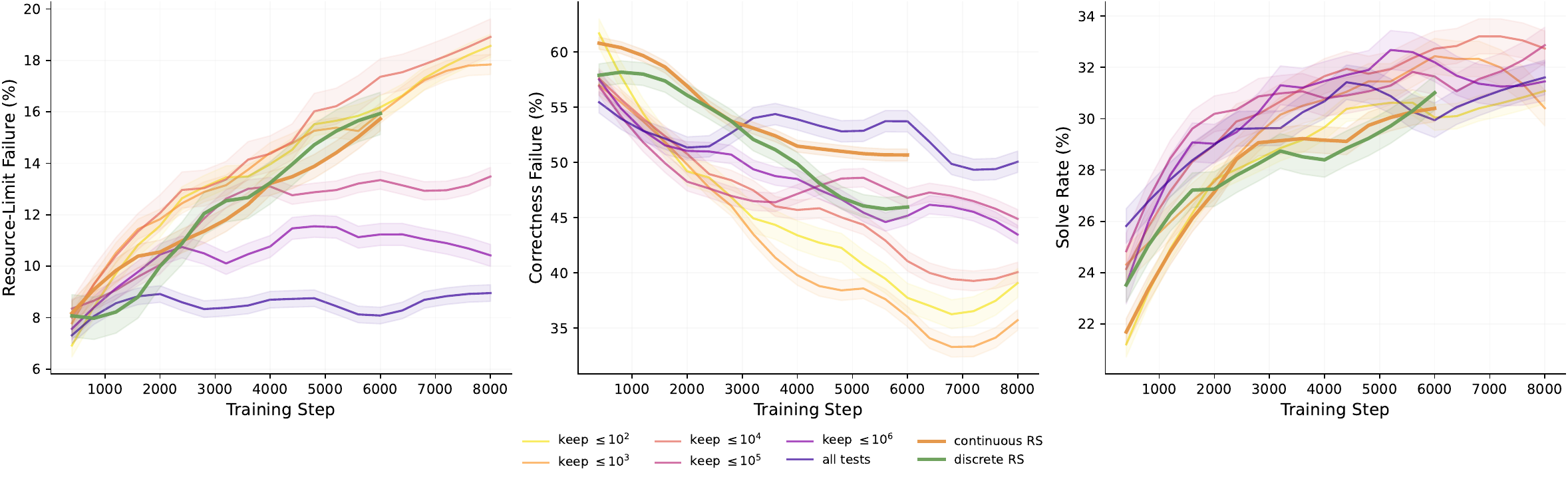}
    \caption{Training dynamics of 32B pure-reasoning setting under reward shaping.
    Partial credit produces a smoother reward signal but does not improve the final solve rate.
    The shaped-reward policy converges to the same region of the trade-off curve as the corresponding uniform-weight binary-reward baseline.}
    \label{fig:reward-shaping-optionc}
\end{figure}

Figure~\ref{fig:reward-shaping-optionc} shows the training dynamics for the two shaped-reward runs.

\subsection{Failure-Conditioned Training Experiments}
\label{app:failure-conditioned}

The sections in the main text establish that the
trade-off is robust across settings and scales.
Longer training can still push the frontier
outward, but a sharper question is whether a
different use of the same training budget can
improve it at matched trainer compute. The
\emph{online-fixing} experiments in this appendix
address that question by changing how worker-side
sampling is spent after failures.

Online-fixing changes two aspects of training at
once: it spends more worker-side sampling on
problems the model currently fails on, and it
augments the online prompt distribution with
failure-conditioned prompts. The
\emph{resample-only} control matches the extra
retry budget without adding failure context, so
it isolates the value of the failure-conditioned
prompt itself.

A key property of the design is that it keeps
trainer compute fixed while increasing worker
compute adaptively after failures. The
computation graph is cut at the boundary between
the failed rollout and the fixing prompt, so the
fixing rollout is treated as a new sample rather
than a second turn in the same trajectory. The
trainer therefore processes the same number of
gradient steps regardless of how many
failure-conditioned rollouts are spawned.

\subsubsection{Repair Prompt and Repairability Metric}
\label{sec:appendix-repair-prompt}

The repair prompt $z = \phi(x, y)$ concatenates
the original problem statement $x$, the failed
code $y$, and aggregated execution statistics, along with brief advice
on interpreting each failure type.
It does not expose private test inputs, expected
outputs, or detailed failing traces.

We define fix@$k$ as the pass@$k$ of repair attempts conditioned on a failed original sample.
Similar to pass@$k$, we report the aggregate number across all the fixing prompts created from the failing attempts.

\subsubsection{Online-Fixing Objective}
\label{sec:appendix-online-fix-objective}

For a prompt $x \sim \mathcal{D}$, let
$y \sim \pi_{\theta}(\cdot \mid x)$
be a rollout and $z = \phi(x,y)$ the fixing prompt
constructed from $x$ and its failed code $y$.
The online-fix objective adds a
failure-conditioned correction term to the
standard GRPO loss $\mathcal{L}_{\mathrm{GRPO}}$:
\begin{align}
\mathcal{L}_{\mathrm{online\mbox{-}fixing}}(\theta)
=
\mathbb{E}_{x \sim \mathcal{D}}
\bigl[
\mathcal{L}_{\mathrm{GRPO}}(x; \theta)
\bigr] + \mathbb{E}_{
  x \sim \mathcal{D},\,
  y \sim \pi_{\theta}(\cdot \mid x)
}
\bigl[
\mathbf{1}[r(y)=0]\,
\mathcal{L}_{\mathrm{GRPO}}(z; \theta)
\bigr].
\end{align}
The \emph{resample-only} control keeps the same second
term, but replaces $z = \phi(x,y)$ with a fresh
retry prompt that omits the failed code and
execution statistics.

The prompt $z$ is constructed using the following template.
\begin{trajbox}[System Prompt]
You are a helpful AI assistant. You always reason
before responding, using the following format:

<think>
your internal reasoning
</think>
your external response
\end{trajbox}

\begin{trajbox}[User Prompt Template]
Solve the following competitive programming
question.

Question:
{problem statement}

For your reference, here is a previous solution
attempt that failed some unit tests:

```python
{failed code}
```

Unit tests execution statistics:
{exec_stats}

You may use this code as reference or inspiration,
but you are free to write a completely new
solution. Make sure to avoid repeating the same
errors. Consider:

- FAILURE or EXCEPTION cases: edge cases,
  boundary values, off-by-one errors, or logic
  flaws.
- TIMEOUT cases: algorithm complexity,
  better data structures, or redundant
  computations.

Your code should be enclosed in triple backticks
like so: ```python YOUR CODE HERE```. Use
the backticks for your code only.
\end{trajbox}

\subsubsection{Effect on the Correctness-Efficiency Frontier}
\label{sec:appendix-online-fix}

\begin{figure}[H]
\centering
\includegraphics[width=\linewidth]{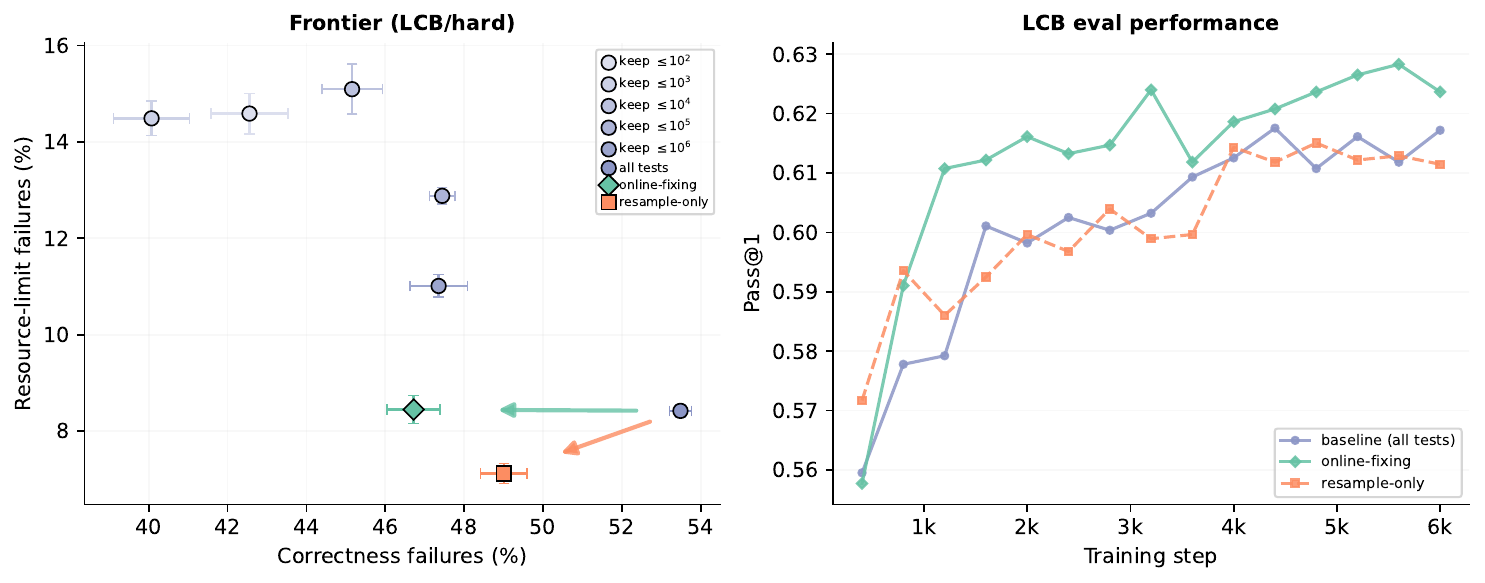}
\caption{Failure-conditioned correction compared with a compute-matched
  resample-only control on 32B pure reasoning setting. \textbf{Left}: on the original code-generation task,
  both online-fixing (green diamond) and the resample-only control (orange
  square) reduce resource-limit failures relative to standard RL checkpoints
  (circles), but only online-fixing simultaneously reduces correctness
  failures. Arrows indicate the shift from the no-filter baseline.
  \textbf{Right}: pass@1 on LCB over training for the original code-generation
  task; the online-fix curve is ahead earlier in training, though the gap
  narrows later. See Appendix~\ref{sec:appendix-online-fix-results} for the
  full per-setting tables.}
\label{fig:online-fix-combined}
\end{figure}

\subsubsection{Results Across Settings}
\label{sec:appendix-online-fix-results}

The clearest case is 32B pure reasoning: relative to the resample-only control, online fixing raises fix@1 on LCB/hard from $14.2\%$ to $17.9\%$ and pass@1 on full LCB from $61.2\%$ to $62.6\%$ (Table~\ref{tab:online-fix-summary}).
Outside this setting, gains on the original code-generation task are smaller and less consistent. We therefore treat these results as exploratory evidence for failure-conditioned training.

\begin{table}[H]
\centering
\small
\caption{Pass rate and self-fixing solve rate for
  additional worker compute comparison
  (32B pure-reasoning setting), including
  online-fixing and its resample-only control,
  reported on LCB/hard and full LCB.
  Numbers are averaged from the last three
  evaluations.}
\vspace{0.5em}
\label{tab:online-fix-summary}
\setlength{\tabcolsep}{4pt}
\begin{tabular}{l cc cc cc cc}
\toprule
& \multicolumn{2}{c}{LCB/hard}
& \multicolumn{2}{c}{LCB}
& \multicolumn{2}{c}{LCB/hard (fix)}
& \multicolumn{2}{c}{LCB (fix)} \\
\cmidrule(lr){2-3} \cmidrule(lr){4-5}
\cmidrule(lr){6-7} \cmidrule(lr){8-9}
& pass@1 & pass@5
& pass@1 & pass@5
& fix@1 & fix@5
& fix@1 & fix@5 \\
\midrule
Baseline
  & 29.2{\tiny$\pm$0.7} & 46.5{\tiny$\pm$1.4}
  & 61.5{\tiny$\pm$0.2} & 72.7{\tiny$\pm$0.8}
  & 14.1{\tiny$\pm$1.4} & 33.5{\tiny$\pm$0.5}
  & 22.2{\tiny$\pm$3.4} & 43.7{\tiny$\pm$2.9} \\
Resample only
  & 30.8{\tiny$\pm$0.3} & \textbf{48.4}{\tiny$\pm$0.7}
  & 61.2{\tiny$\pm$0.1} & \textbf{73.3}{\tiny$\pm$0.4}
  & 14.2{\tiny$\pm$0.5} & 34.9{\tiny$\pm$1.3}
  & 20.8{\tiny$\pm$0.5} & 44.1{\tiny$\pm$0.7} \\
Online-fixing
  & \textbf{30.9}{\tiny$\pm$1.1} & 45.8{\tiny$\pm$0.6}
  & \textbf{62.6}{\tiny$\pm$0.2} & 72.8{\tiny$\pm$0.3}
  & \textbf{17.9}{\tiny$\pm$1.3} & \textbf{37.5}{\tiny$\pm$1.0}
  & \textbf{28.1}{\tiny$\pm$1.0} & \textbf{49.1}{\tiny$\pm$0.9} \\
\bottomrule
\end{tabular}
\end{table}

\begin{table}[H]
\centering
\small
\caption{Pass rate and self-fixing solve rate for
  additional worker compute comparison
  (7B pure-reasoning setting), including
  online-fixing and its resample-only control,
  reported on LCB/hard and full LCB.
  Numbers are averaged from the last three
  evaluations.}
\vspace{0.5em}
\label{tab:online-fix-summary-7b}
\setlength{\tabcolsep}{4pt}
\begin{tabular}{l cc cc cc cc}
\toprule
& \multicolumn{2}{c}{LCB/hard}
& \multicolumn{2}{c}{LCB}
& \multicolumn{2}{c}{LCB/hard (fix)}
& \multicolumn{2}{c}{LCB (fix)} \\
\cmidrule(lr){2-3} \cmidrule(lr){4-5}
\cmidrule(lr){6-7} \cmidrule(lr){8-9}
& pass@1 & pass@5
& pass@1 & pass@5
& fix@1 & fix@5
& fix@1 & fix@5 \\
\midrule
Baseline
  & 13.7{\tiny$\pm$0.6} & 27.4{\tiny$\pm$0.9}
  & 45.7{\tiny$\pm$0.1} & 59.6{\tiny$\pm$0.3}
  & 7.4{\tiny$\pm$0.5} & 19.5{\tiny$\pm$0.9}
  & 15.4{\tiny$\pm$1.9} & 34.2{\tiny$\pm$2.2} \\
Resample only
  & \textbf{14.3}{\tiny$\pm$0.3} & \textbf{29.2}{\tiny$\pm$1.0}
  & 45.6{\tiny$\pm$0.3} & \textbf{60.6}{\tiny$\pm$0.7}
  & 7.9{\tiny$\pm$0.5} & 21.9{\tiny$\pm$1.5}
  & 17.2{\tiny$\pm$0.9} & 37.2{\tiny$\pm$0.1} \\
Online-fixing
  & 14.2{\tiny$\pm$1.0} & 28.0{\tiny$\pm$0.3}
  & \textbf{45.9}{\tiny$\pm$0.6} & 60.1{\tiny$\pm$0.1}
  & \textbf{9.7}{\tiny$\pm$0.7} & \textbf{22.8}{\tiny$\pm$0.7}
  & \textbf{18.5}{\tiny$\pm$0.6} & \textbf{38.2}{\tiny$\pm$0.5} \\
\bottomrule
\end{tabular}
\end{table}

\begin{table}[H]
\centering
\small
\caption{Pass rate and self-fixing solve rate for
  additional worker compute comparison
  (32B tool-interleaved setting), including
  online-fixing and its resample-only control,
  reported on LCB/hard and full LCB.
  Numbers are averaged from the last three
  evaluations.}
\vspace{0.5em}
\label{tab:online-fix-summary-32b-tool}
\setlength{\tabcolsep}{4pt}
\begin{tabular}{l cc cc cc cc}
\toprule
& \multicolumn{2}{c}{LCB/hard}
& \multicolumn{2}{c}{LCB}
& \multicolumn{2}{c}{LCB/hard (fix)}
& \multicolumn{2}{c}{LCB (fix)} \\
\cmidrule(lr){2-3} \cmidrule(lr){4-5}
\cmidrule(lr){6-7} \cmidrule(lr){8-9}
& pass@1 & pass@5
& pass@1 & pass@5
& fix@1 & fix@5
& fix@1 & fix@5 \\
\midrule
Baseline
  & 34.6{\tiny$\pm$1.3} & 52.0{\tiny$\pm$1.3}
  & 64.5{\tiny$\pm$0.7} & 75.6{\tiny$\pm$0.6}
  & 15.5{\tiny$\pm$0.5} & 36.2{\tiny$\pm$1.9}
  & 20.3{\tiny$\pm$1.3} & 42.7{\tiny$\pm$2.1} \\
Resample only
  & 34.7{\tiny$\pm$0.9} & \textbf{54.8}{\tiny$\pm$2.5}
  & 64.3{\tiny$\pm$0.7} & \textbf{77.6}{\tiny$\pm$1.4}
  & 13.4{\tiny$\pm$2.4} & 32.7{\tiny$\pm$2.7}
  & 19.8{\tiny$\pm$3.3} & 41.7{\tiny$\pm$4.0} \\
Online-fixing
  & \textbf{36.2}{\tiny$\pm$1.0} & 52.9{\tiny$\pm$1.8}
  & \textbf{65.2}{\tiny$\pm$0.4} & 75.9{\tiny$\pm$0.9}
  & \textbf{16.3}{\tiny$\pm$1.5} & \textbf{36.3}{\tiny$\pm$1.7}
  & \textbf{21.2}{\tiny$\pm$0.7} & \textbf{43.9}{\tiny$\pm$0.6} \\
\bottomrule
\end{tabular}
\end{table}

\begin{figure}[h]
  \centering
  \includegraphics[width=\linewidth]{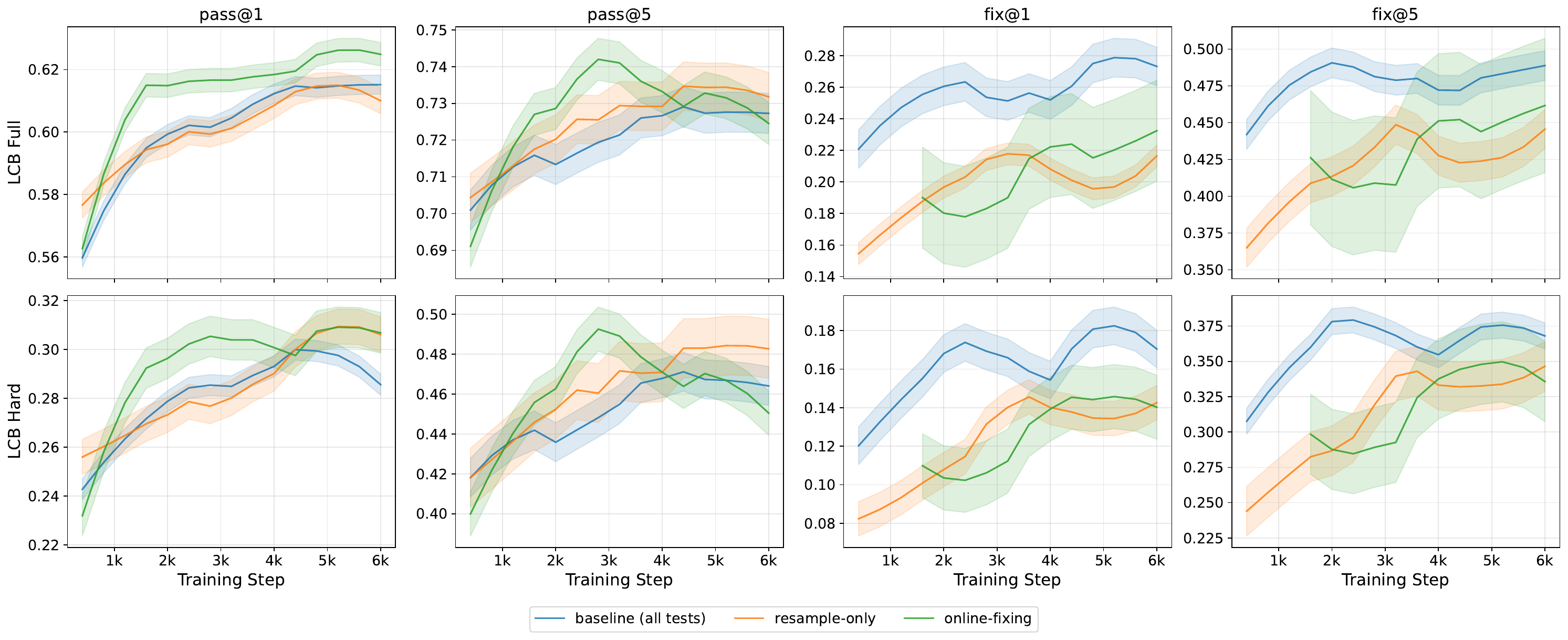}
\caption{Evaluation curves for the 32B pure-reasoning runs over the first 6k training steps.
  \textbf{Top row}: full LCB evaluation set.
  \textbf{Bottom row}: hard subset only.
  Columns show pass@1, pass@5, fix@1, and fix@5 respectively.
  Online-fixing matches or slightly exceeds the baseline and resample-only control on the standard code-generation metrics (pass@1, pass@5), while showing a clear advantage on the self-repair metrics (fix@1, fix@5).}
\label{fig:eval-combined}
\end{figure}

Beyond the original code-generation metrics,
online-fix training also strengthens the
self-fixing task throughout training.
Figure~\ref{fig:eval-combined} shows that
the online-fix model consistently outperforms
both the resample-only control and the baseline
on fix@1 and fix@5.

\clearpage
\section{Case Studies: Full Trajectories Across the Frontier}
\label{app:case-studies}

We present two case studies from the tool-coding setting that
concretely illustrate the correctness-efficiency trade-off.
Both come from LCB/hard problems that are
\emph{differentially solved} across the interpolation axis.
For each, we show the \emph{complete multi-turn trajectory}
of a representative rollout from the two RL endpoints:
\begin{itemize}
\item \textbf{Low-coverage} ($\alpha{=}0$;
  \codecontestsplus{} with test inputs $\leq 10^2$):
  trained against small-input tests only.
\item \textbf{High-coverage} ($\alpha{=}1$;
  \codecontestsplus{} unfiltered):
  trained against the full test suite including
  large-input stress tests.
\end{itemize}
Both checkpoints are evaluated at step~3600 with
5~samples per problem.
Thinking blocks are truncated for space; all code and tool
output are shown verbatim.

\subsection{Case 1: Falling Bars (hard/770)}
\label{app:case-study-770}

\paragraph{Problem.}
Given an $H \times W$ grid with $N$ horizontal bars, simulate
gravity: at each time step, bars are processed in
index order $i{=}1,\ldots,N$; bar~$i$ moves down one row if
all cells directly below it are unoccupied.
Output each bar's final row at $t{=}10^{100}$.
Constraints: $H, W, N \leq 2 \times 10^5$.

\paragraph{Results.}
Low-coverage: \textbf{5/5 success.}\quad
High-coverage: \textbf{0/5 success} (all wrong answer,
e.g.\ \texttt{2,3,4,4} instead of \texttt{2,4,3,4}).

\subsubsection{Low-Coverage Trajectory (Success)}

\begin{trajbox}[Assistant -- Turn 1 (python-io): Reverse-order single pass]
<think>
... The key insight here is that once a bar stops moving, it
might block other bars from moving. So, the order in which bars
are processed matters. ... One approach is to ... process bars
in reverse index order, tracking the highest occupied row per
column. ...
</think>

<tool: python-io>
<code>
bars.sort(key=lambda x: -x[3])  # reverse index order
occupied = {}  # column -> highest occupied row
result = [0] * N
for R, C, L, idx in bars:
    columns = list(range(C, C + L))
    max_row = H
    for c in columns:
        h = occupied.get(c, 0)
        current_max = h - 1 if h else H
        max_row = min(max_row, current_max)
    result[idx] = max_row
    for c in columns:
        occupied[c] = max(occupied.get(c, 0), max_row)
for row in result:
    print(row)
</code>
<stdin>4 4 4
1 2 3
3 2 2
2 1 2
2 4 1</stdin>
</tool>
\end{trajbox}

\begin{trajbox}[Tool Response]
completed. [stdout]2
3
4
4[/stdout]

<budget> Remaining turns: 5, Remaining tokens: 45462 </budget>
\end{trajbox}

\paragraph{Turns 2 and 3.} Same reverse-order approach with
minor refactors; output remains \texttt{2,3,4,4}.
Stress test confirms the bug.

\begin{trajbox}[Assistant -- Turn 4 (python-io): Forward order with static lookahead]
<think>
... The reverse processing approach is fundamentally flawed for
this problem. Let me think differently. Perhaps the correct
approach is to process bars in original order and also check
initial positions of unprocessed bars ...
</think>

<tool: python-io>
<code>
for R, C, L, idx in bars:  # forward order now
    columns = list(range(C, C+L))
    min_max_row = H
    for c in columns:
        current_h = occupied.get(c, 0)
        candidate = current_h - 1 if current_h else H
        min_max_row = min(min_max_row, candidate)
    # Also check initial positions of unprocessed bars
    for j in range(N):
        if j > idx:  # unprocessed (higher index)
            r_j, c_j, l_j, _ = bars[j]
            if r_j > R:  # bar j is initially below
                overlap_start = max(C, c_j)
                overlap_end = min(C + L - 1, c_j + l_j - 1)
                if overlap_start <= overlap_end:
                    min_max_row = min(min_max_row, r_j)
    result[idx] = min_max_row
    ...
</code>
<stdin>4 4 4
1 2 3
3 2 2
2 1 2
2 4 1</stdin>
</tool>
\end{trajbox}

\begin{trajbox}[Tool Response]
completed. [stdout]2
1
1
1[/stdout]

<budget> Remaining turns: 2, Remaining tokens: 39584 </budget>
\end{trajbox}

\begin{trajbox}[Assistant -- Turn 5 (python-io): Brute-force simulation |$\leftarrow$| \textbf{pivot}]
<think>
... The code now produces output 2, 1, 1, 1, which is even
worse. Clearly, this approach is incorrect. The issue is that
checking the initial positions of unprocessed bars is not
sufficient, as those bars may move down in subsequent time
steps. ... Given the need to provide a correct solution, I'll
try a direct simulation approach.
</think>

<tool: python-io>
<code>
current_rows = [R for R, _, _, _ in bars]
occupied = [set() for _ in range(H+2)]
for i in range(N):
    R, C, L, _ = bars[i]
    for c in range(C, C+L):
        occupied[R].add(c)
changed = True
while changed:
    changed = False
    for i in range(N):
        R, C, L, _ = bars[i]
        current_R = current_rows[i]
        if current_R == H:
            continue
        can_move = True
        for c in range(C, C+L):
            if c in occupied[current_R + 1]:
                can_move = False
                break
        if can_move:
            for c in range(C, C+L):
                occupied[current_R].remove(c)
            current_rows[i] += 1
            for c in range(C, C+L):
                occupied[current_R + 1].add(c)
            changed = True
for row in current_rows:
    print(row)
</code>
<stdin>4 4 4
1 2 3
3 2 2
2 1 2
2 4 1</stdin>
</tool>
\end{trajbox}

\begin{trajbox}[Tool Response]
completed. [stdout]2
4
3
4[/stdout]

<budget> Remaining turns: 1, Remaining tokens: 38194 </budget>
\end{trajbox}

\paragraph{Turn 6.} Stress-tests the simulation on
$N{=}200{,}000$; encounters a \texttt{KeyError} on random
overlapping bars but the core algorithm is correct.

\begin{trajbox}[Assistant -- Final Submission]
<think>
The simulation approach produces the correct output for the
sample input. The stress test revealed a KeyError when removing
a column, likely due to overlapping bars in the random test.
Given the time constraints, I'll submit the simulation code.
</think>

|{\footnotesize\itshape(submits the simulation code from Turn~5)}|
\end{trajbox}

\subsubsection{High-Coverage Trajectory (Wrong Answer)}

\begin{trajbox}[Assistant -- Turn 1 (python-io): Segment tree, reverse order]
<think>
... I need to find a way to model the dependencies between
bars. ... Let me consider processing bars in reverse index
order, using a segment tree to track the minimum occupied row
for each column range ...
</think>

<tool: python-io>
<code>
import sys, math

def main():
    data = sys.stdin.read().split()
    H, W, N = int(data[0]), int(data[1]), int(data[2])
    bars = []
    idx = 3
    for _ in range(N):
        R, C, L = int(data[idx]), int(data[idx+1]), int(data[idx+2])
        bars.append((R, C, L)); idx += 3
    result = [0]*N
    size = 1
    while size < W: size <<= 1
    min_row = [math.inf] * (2 * size)
    def update(l, r, val):
        l += size - 1; r += size - 1
        while l <= r:
            if l % 2 == 1: min_row[l] = min(min_row[l], val); l += 1
            if r % 2 == 0: min_row[r] = min(min_row[r], val); r -= 1
            l >>= 1; r >>= 1
    def query(l, r):
        res = math.inf; l += size - 1; r += size - 1
        while l <= r:
            if l % 2 == 1: res = min(res, min_row[l]); l += 1
            if r % 2 == 0: res = min(res, min_row[r]); r -= 1
            l >>= 1; r >>= 1
        return res
    for i in reversed(range(N)):
        R, C, L = bars[i]
        current_min = query(C, C + L - 1)
        new_R = min(H, current_min - 1) if current_min != math.inf else H
        result[i] = new_R
        update_val = new_R + 1 if new_R < H else math.inf
        update(C, C + L - 1, update_val)
    for r in result: print(r)
main()
</code>
<stdin>4 4 4
1 2 3
3 2 2
2 1 2
2 4 1</stdin>
</tool>
\end{trajbox}

\begin{trajbox}[Tool Response]
completed. [stdout]4
4
4
4[/stdout]

<budget> Remaining turns: 5, Remaining tokens: 46198 </budget>
\end{trajbox}

\paragraph{Turn 2 (\texttt{python-io}).} Fixes segment-tree
indexing to 0-based; output unchanged: \texttt{4,4,4,4}.

\begin{trajbox}[Assistant -- Turn 3 (python-stress): Assertion failure]
<think>
... The output is still incorrect. All bars are at row 4, but
the correct output should be 2, 4, 3, 4. The issue must be in
the segment tree implementation or the logic of updating ...
</think>

<tool: python-stress>
... |{\footnotesize\itshape(same reverse-order segment tree code)}|
result = solve(H, W, N, bars)
print("Result:", result)
print("Expected:", expected)
assert result == expected, f"Test failed: {result} != {expected}"
</tool>
\end{trajbox}

\begin{trajbox}[Tool Response]
completed. [stdout]Result: [4, 4, 4, 4]
Expected: [2, 4, 3, 4][/stdout]
[stderr]AssertionError: Test failed: [4, 4, 4, 4] != [2, 4, 3, 4][/stderr]
\end{trajbox}

\begin{trajbox}[Assistant -- Turn 4 (python-io): Plain array, still reverse order]
<think>
... The stress test confirms that the current approach is
incorrect. ... Let me simplify and use a plain array instead of
a segment tree, but keep the reverse processing order ...
</think>

<tool: python-io>
<code>
min_row = [H+1] * (W + 2)  # 1-based
for i in reversed(range(N)):
    R, C, L = bars[i]
    current_min = min(min_row[C : C+L])
    if current_min == H+1:
        new_R = H
    else:
        new_R = min(H, current_min - 1)
    result[i] = new_R
    for c in range(C, C+L):
        if min_row[c] > new_R:
            min_row[c] = new_R
</code>
<stdin>4 4 4
1 2 3
3 2 2
2 1 2
2 4 1</stdin>
</tool>
\end{trajbox}

\begin{trajbox}[Tool Response]
completed. [stdout]2
3
4
4[/stdout]

<budget> Remaining turns: 2, Remaining tokens: 40695 </budget>
\end{trajbox}

\paragraph{Turn 5 (\texttt{python-io}).} Changes update value
to \texttt{new\_R + 1}; output regresses to \texttt{4,4,4,4}.

\paragraph{Turn 6 (\texttt{python-io}).} Reverts to
\texttt{new\_R}; output returns to \texttt{2,3,4,4}, which is still
wrong. The model is stuck cycling between two incorrect
variants.

\begin{trajbox}[Assistant -- Final Submission]
<think>
... The output is still [2, 3, 4, 4], which is close but not
correct. Bar 2 should be at row 4, not row 3. The fundamental
issue is in how the reverse processing handles the interaction
between bars ...
</think>

|{\footnotesize\itshape(submits the reverse-order min\_row code from Turn~4; \textbf{wrong answer})}|
\end{trajbox}

\subsubsection{Interpretation}

Both models initially try the same efficient
reverse-processing idea. The low-coverage model
\emph{pivots} to brute-force simulation when it fails
(Turn~5); the high-coverage model spends all 6 tool turns
on variants of the single-pass approach, never attempting
simulation. The low-coverage model's willingness to fall
back to a correct-but-slow algorithm is the difference
between 5/5 success and 0/5.

\subsection{Case 2: Largest Palindrome Divisible by $K$ (hard/680)}
\label{app:case-study-680}

\paragraph{Problem.}
Given $n$ and $k$ ($1 \leq n \leq 10^5$, $1 \leq k \leq 9$),
find the largest $n$-digit palindrome divisible by~$k$.

\paragraph{Results.}
Low-coverage: \textbf{0/5 success} (all timeout).\quad
High-coverage: \textbf{1/5 success} (3 timeout, 1 wrong answer).

\subsubsection{Low-Coverage Trajectory (Timeout)}

\begin{trajbox}[Assistant -- Turn 1 (python-io): Brute-force decrement]
<think>
... The task is to find the largest n-digit number that is both
a palindrome and divisible by k. ... Start from the largest
palindrome and work downward until we find one divisible by k.
</think>

<tool: python-io>
<code>
class Solution:
    def largestPalindrome(self, n: int, k: int) -> str:
        m = (n + 1) // 2
        first_half = '9' * m
        while True:
            if n % 2 == 0:
                palindrome = first_half + first_half[::-1]
            else:
                palindrome = first_half + first_half[:-1][::-1]
            if self.is_divisible(palindrome, k):
                return palindrome
            new_first_half = self.decrement_first_half(first_half)
            if new_first_half is None:
                return ""
            first_half = new_first_half

    def is_divisible(self, s, k):
        mod = 0
        for c in s:
            mod = (mod * 10 + int(c)) % k
        return mod == 0

    def decrement_first_half(self, s):
        s_list = list(s)
        i = len(s_list) - 1
        carry = 1
        while i >= 0 and carry:
            digit = int(s_list[i])
            if digit - carry >= 0:
                s_list[i] = str(digit - carry)
                carry = 0
            else:
                s_list[i] = '9'
                carry = 1
            i -= 1
        if s_list[0] == '0':
            return None
        return ''.join(s_list)
</code>
<stdin>3
5</stdin>
</tool>
\end{trajbox}

\begin{trajbox}[Tool Response]
completed.

<budget> Remaining turns: 5, Remaining tokens: 51467 </budget>
\end{trajbox}

\paragraph{Turn 2.} Tests second example (\texttt{n=1,k=4});
passes with output \texttt{"8"}.

\begin{trajbox}[Assistant -- Turn 3 (python-stress): Large input test]
<think>
... Let me test with large inputs to check performance. For
n=100000, the first_half has 50000 digits ...
</think>

<tool: python-stress>
...
test_cases = [(3, 5), (1, 4), (5, 6), (2, 2), (100000, 9)]
for n, k in test_cases:
    start = time.time()
    result = Solution().largestPalindrome(n, k)
    elapsed = time.time() - start
    print(f"n={n}, k={k}: time={elapsed:.3f}s")
</tool>
\end{trajbox}

\begin{trajbox}[Tool Response]
error: EXECUTION_TIMEOUT after 60 seconds
\end{trajbox}

\begin{trajbox}[Assistant -- Final Submission]
<think>
... The stress test timed out for n=100000. The brute-force
approach is too slow for large inputs. But given the time
constraints, I'll submit this solution.
</think>

|{\footnotesize\itshape(submits brute-force decrement code; \textbf{timeout} on hidden tests)}|
\end{trajbox}

\subsubsection{High-Coverage Trajectory (Success)}

\begin{trajbox}[Assistant -- Turn 1 (python-io): Optimized remainder, still brute-force]
<think>
... Let me start by considering the structure of the palindrome.
For an n-digit number, each half-digit contributes to two
symmetric positions. I can precompute the coefficient of each
position modulo k to speed up the divisibility check ...
</think>

<tool: python-io>
<code>
class Solution:
    def largestPalindrome(self, n: int, k: int) -> str:
        if k == 1: return '9' * n
        m = (n + 1) // 2 if n % 2 else n // 2
        pow10_mod = [1] * n
        for i in range(1, n):
            pow10_mod[i] = (pow10_mod[i-1] * 10) % k
        def get_remainder(s):
            remainder = 0
            for i in range(len(s)):
                digit = int(s[i])
                if n % 2 == 0 or i < len(s) - 1:
                    coeff = (pow10_mod[n-1-i] + pow10_mod[i]) % k
                else:
                    coeff = pow10_mod[n-1-i] % k
                remainder = (remainder + digit * coeff) % k
            return remainder
        first_half = '9' * m
        while True:
            if get_remainder(first_half) == 0:
                if n % 2:
                    return first_half + first_half[:-1][::-1]
                else:
                    return first_half + first_half[::-1]
            s_list = list(first_half)
            i = len(s_list) - 1
            while i >= 0 and s_list[i] == '0':
                s_list[i] = '9'; i -= 1
            if i >= 0: s_list[i] = str(int(s_list[i]) - 1)
            first_half = ''.join(s_list)
            if first_half[0] == '0': return "0" * n
</code>
<stdin>3
5</stdin>
</tool>
\end{trajbox}

\begin{trajbox}[Tool Response]
completed.

<budget> Remaining turns: 5, Remaining tokens: 48438 </budget>
\end{trajbox}

\begin{trajbox}[Assistant -- Turn 2 (python-stress): Discovers timeout]
<think>
... Let me test with large inputs to verify performance ...
</think>

<tool: python-stress>
test_cases = [(10**5, 9), (10**5, 2), (10**5, 5)]
for n, k in test_cases:
    start = time.time()
    result = Solution().largestPalindrome(n, k)
    elapsed = time.time() - start
    print(f"n={n}, k={k} | Time: {elapsed:.3f}s")
</tool>
\end{trajbox}

\begin{trajbox}[Tool Response]
error: EXECUTION_TIMEOUT after 60 seconds

<budget> Remaining turns: 4, Remaining tokens: 46406 </budget>
\end{trajbox}

\begin{trajbox}[Assistant -- Turn 3 (python-io): DP with backward reachability |$\leftarrow$| \textbf{pivot}]
<think>
... The stress test timed out for large n=1e5. The main issue
is the decrement loop -- for large n, the first_half can be up
to 5e4 digits, and decrementing each time is O(m). We need a
fundamentally different approach. ...

Since k <= 9, the remainder modulo k has at most 9 possible
values. For each half-digit position, we can precompute its
coefficient mod k. Then use DP: compute which remainders are
reachable from each suffix of the half-string, and greedily
assign the largest valid digit at each position.
</think>

<tool: python-io>
<code>
class Solution:
    def largestPalindrome(self, n: int, k: int) -> str:
        if k == 1: return '9' * n
        m = (n + 1) // 2 if n % 2 else n // 2
        pow10_mod = [1] * n
        for i in range(1, n):
            pow10_mod[i] = (pow10_mod[i-1] * 10) % k
        coeff = []
        for i in range(m):
            if n % 2 == 0 or i < m - 1:
                c = (pow10_mod[n-1-i] + pow10_mod[i]) % k
            else:
                c = pow10_mod[n-1-i] % k
            coeff.append(c)
        # Backward DP: which remainders are reachable?
        dp_backward = [set() for _ in range(m + 1)]
        dp_backward[m].add(0)
        for i in range(m-1, -1, -1):
            current_coeff = coeff[i]
            for d in range(10):
                if i == 0 and d == 0: continue
                contrib = d * current_coeff
                for r in dp_backward[i+1]:
                    dp_backward[i].add((r + contrib) % k)
        # Greedy forward pass
        first_half = []
        current_remainder = 0
        for i in range(m):
            for d in range(9, -1, -1):
                if i == 0 and d == 0: continue
                contrib = d * coeff[i]
                target = (-(current_remainder + contrib)) % k
                if target in dp_backward[i+1]:
                    first_half.append(str(d))
                    current_remainder = (current_remainder + contrib) % k
                    break
        first_half_str = ''.join(first_half)
        if n % 2:
            return first_half_str + first_half_str[:-1][::-1]
        else:
            return first_half_str + first_half_str[::-1]
</code>
<stdin>3
5</stdin>
</tool>
\end{trajbox}

\begin{trajbox}[Tool Response]
completed.

<budget> Remaining turns: 3, Remaining tokens: 41605 </budget>
\end{trajbox}

\begin{trajbox}[Assistant -- Turn 4 (python-stress): Efficiency confirmed]
<think>
... Now let's test the DP solution with large inputs to
confirm it runs within the time limit ...
</think>

<tool: python-stress>
test_cases = [(10**5, 9), (10**5, 2), (10**5, 5)]
for n, k in test_cases:
    start = time.time()
    result = Solution().largestPalindrome(n, k)
    elapsed = time.time() - start
    print(f"n={n}, k={k} | Time: {elapsed:.3f}s | PASS")
</tool>
\end{trajbox}

\begin{trajbox}[Tool Response]
completed. [stdout]n=100000, k=9 | Time: 0.557s | PASS
n=100000, k=2 | Time: 0.158s | PASS
n=100000, k=5 | Time: 0.152s | PASS[/stdout]
[performance] Execution Time: 0.959s, Peak Memory: 55.1 MB

<budget> Remaining turns: 2, Remaining tokens: 39572 </budget>
\end{trajbox}

\paragraph{Turns 5 and 6.} Tests small examples
(\texttt{n=1,k=4}; \texttt{n=3,k=5}; \texttt{n=5,k=6}) to
verify correctness; all pass.

\begin{trajbox}[Assistant -- Final Submission]
<think>
Solution passes all tests and runs well within the time limit.
</think>

|{\footnotesize\itshape(submits the DP code from Turn~3; \textbf{passes} all hidden tests)}|
\end{trajbox}

\subsubsection{Interpretation}

Both models start with brute-force and both discover the
timeout at $n{=}10^5$. The critical difference is what happens
\emph{after}: the low-coverage model has already exhausted
its tool budget on minor variations and submits the
brute-force code; the high-coverage model invests its remaining
turns in developing a DP formulation that reduces complexity
from $O(n \cdot 10^m/k)$ to $O(m \cdot k)$.

\subsection{Discussion}
\label{app:case-study-discussion}

These two trajectories make the frontier concrete:

\begin{itemize}
\item \textbf{hard/770 (Falling Bars):}
  Both models initially try efficient reverse-processing.
  The low-coverage model \emph{pivots} to brute-force
  simulation when it fails;
  the high-coverage model stays committed to the
  efficient-but-wrong approach.

\item \textbf{hard/680 (K-Palindrome):}
  Both models start with brute-force and discover it
  times out.
  The high-coverage model \emph{pivots} to dynamic
  programming;
  the low-coverage model submits the slow code.
\end{itemize}

The pattern is symmetric: each model is more likely to
\emph{abandon a failing strategy} in the direction its
training reward incentivizes.
The low-coverage model, never penalized for timeouts
during training, readily falls back to correct-but-slow
algorithms, while the high-coverage model, trained
against full stress tests, readily develops efficient
algorithms even when correctness is uncertain.
Neither behavior dominates across all problems, which
is why interpolation and extrapolation can access
solutions that no single endpoint produces.

\section{Example Trajectories from Tool-Use Training}
\label{sec:appendix-tool-examples}

We show one representative trajectory from each
iteration of the tool-use training (Appendix~\ref{sec:appendix-tool-training}) to illustrate the tool-use patterns the
model acquires. Thinking blocks are truncated for
space; all code and tool output are shown verbatim.

\paragraph{Iteration 0: testing with
\texttt{python-io}.}
The system prompt gives the model access to
\texttt{python} and \texttt{python-io} tools with a
budget of 5 tool turns. The following trajectory
solves a phone-keypad cipher problem.

\begin{trajbox}[System Prompt (Iteration 0)]
You are an expert Python programmer and tool user. You have access to
"python" and "python-io" tools. You are constrained by a 61440 total
context window limit. You can interact with these tools through the
following tool interface for at most 5 turns.

<tool: python>
Run Python code in an isolated sandbox.
</tool>

<tool: python-io>
Run Python code with stdin input in an isolated sandbox.
Use <code>...</code> for code and <stdin>...</stdin> for input.
</tool>

You will receive a budget notice after each tool execution indicating
your remaining turns and tokens.
\end{trajbox}

\begin{trajbox}[User]
Provide a Python solution for the following competitive programming
question: Solve The Mystery

Input:
First line contains T - number of test cases.
Following T lines each contains a string of characters in [a-z] only

Output: Print a numeric string for each test case.

Constraints: 1 <= T <= 100, 1 <= Length of String <= 100

SAMPLE INPUT            SAMPLE OUTPUT
10                      539
key                     86
to                      2668765
control                 6463
nine                    82457
tails                   47
is                      46
in                      9687
your                    4263
hand                    74663
phone
\end{trajbox}

\begin{trajbox}[Assistant -- Turn 1 (python-io)]
<think>
I need to map each letter to its phone keypad digit. Looking at the
sample: k->5, e->3, y->9 gives "539". This is the standard phone
keypad mapping: abc->2, def->3, ghi->4, jkl->5, mno->6, pqrs->7,
tuv->8, wxyz->9.
</think>

<tool: python-io>
<code>
keypad = {
    'a': '2', 'b': '2', 'c': '2',
    'd': '3', 'e': '3', 'f': '3',
    'g': '4', 'h': '4', 'i': '4',
    'j': '5', 'k': '5', 'l': '5',
    'm': '6', 'n': '6', 'o': '6',
    'p': '7', 'q': '7', 'r': '7', 's': '7',
    't': '8', 'u': '8', 'v': '8',
    'w': '9', 'x': '9', 'y': '9', 'z': '9'
}
t = int(input())
for _ in range(t):
    s = input().strip()
    print(''.join(keypad[c] for c in s))
</code>
<stdin>
10
key
to
control
nine
tails
is
in
your
hand
phone
</stdin>
</tool>
\end{trajbox}

\begin{trajbox}[Tool Response]
completed. [stdout]539
86
2668765
6463
82457
47
46
9687
4263
74663[/stdout]

<budget> Remaining turns: 4, Remaining tokens: 59534 </budget>
\end{trajbox}

\begin{trajbox}[Assistant -- Turn 2 (python-io): Edge cases]
<think>
The sample output matches. Let me test edge cases: 4-letter groups
(pqrs, wxyz), single characters, and all-same-group strings.
</think>

<tool: python-io>
<code>
keypad = { ... }  # same dictionary as above
t = int(input())
for _ in range(t):
    s = input().strip()
    print(''.join(keypad[c] for c in s))
</code>
<stdin>
5
aaa
pqrs
wxyz
a
z
</stdin>
</tool>
\end{trajbox}

\begin{trajbox}[Tool Response]
completed. [stdout]222
7777
9999
2
9[/stdout]

<budget> Remaining turns: 3, Remaining tokens: 59081 </budget>
\end{trajbox}

\begin{trajbox}[Assistant -- Final Submission]
<think>
All edge cases pass. The solution handles all constraints correctly.
</think>

```python
keypad = {
    'a': '2', 'b': '2', 'c': '2',
    'd': '3', 'e': '3', 'f': '3',
    'g': '4', 'h': '4', 'i': '4',
    'j': '5', 'k': '5', 'l': '5',
    'm': '6', 'n': '6', 'o': '6',
    'p': '7', 'q': '7', 'r': '7', 's': '7',
    't': '8', 'u': '8', 'v': '8',
    'w': '9', 'x': '9', 'y': '9', 'z': '9'
}
t = int(input())
for _ in range(t):
    s = input().strip()
    print(''.join(keypad[c] for c in s))
```
\end{trajbox}

\paragraph{Iteration 1: stress-testing with
\texttt{python-stress}.}
The system prompt adds the \texttt{python-stress}
tool, which runs code with automatic timing and
memory measurement. The following trajectory solves
a prefix-sum divisibility counting problem
($N \leq 10^5$, time limit 2\,s).

\begin{trajbox}[System Prompt additions (Iteration 1)]
You have access to "python", "python-io" and "python-stress" tools.

<tool: python-stress>
Run Python code with performance testing and automatic output
suppression to save context. This tool automatically measures
execution time and memory usage.
IMPORTANT: Output is limited to the last 50 lines. Use this for
efficiency testing with large inputs. You MUST generate test cases
programmatically inside the code - do not use <stdin>.
DO NOT print large data structures - only print performance metrics.
</tool>
\end{trajbox}

\begin{trajbox}[User]
Provide a Python solution for the following competitive programming
question. Time Limit: 2.0 s. Memory Limit: 1073 MB.

There are N boxes arranged in a row from left to right. The i-th box
from the left contains A_i candies. You will take out the candies
from some consecutive boxes and distribute them evenly to M children.

Find the number of pairs (l, r) such that 1 <= l <= r <= N and
A_l + A_{l+1} + ... + A_r is a multiple of M.

Constraints: 1 <= N <= 10^5, 2 <= M <= 10^9, 1 <= A_i <= 10^9

Examples:
  3 2 / 4 1 5         -> 3
  13 17 / 29 7 ... 81  -> 6
  10 400000000 / 10^9 x10 -> 25
\end{trajbox}

\begin{trajbox}[Assistant -- Turn 1 (python-io): Parse input]
<think>
This is a classic prefix-sum modulo problem. First let me validate
the input parsing.
</think>

<tool: python-io>
<code>
n, m = map(int, input().split())
a = list(map(int, input().split()))
print(f"N: {n}, M: {m}, Array: {a}")
</code>
<stdin>
3 2
4 1 5
</stdin>
</tool>
\end{trajbox}

\begin{trajbox}[Tool Response]
completed. [stdout]N: 3, M: 2, Array: [4, 1, 5][/stdout]

<budget> Remaining turns: 4, Remaining tokens: 61155 </budget>
\end{trajbox}

\begin{trajbox}[Assistant -- Turn 2 (python): Core algorithm test]
<think>
Now test the prefix-sum-modulo + frequency-counting approach.
</think>

<tool: python>
def solve():
    n, m = 3, 2
    a = [4, 1, 5]
    count = 0
    mod_freq = {0: 1}
    prefix_sum_mod = 0
    for num in a:
        prefix_sum_mod = (prefix_sum_mod + num) % m
        if prefix_sum_mod in mod_freq:
            count += mod_freq[prefix_sum_mod]
        mod_freq[prefix_sum_mod] = mod_freq.get(
            prefix_sum_mod, 0) + 1
    print(f"Count: {count}")
    assert count == 3
solve()
</tool>
\end{trajbox}

\begin{trajbox}[Tool Response]
completed. [stdout]Count: 3[/stdout]

<budget> Remaining turns: 3, Remaining tokens: 60885 </budget>
\end{trajbox}

\begin{trajbox}[Assistant -- Turn 3 (python-io): Full solution test]
<think>
Core logic works. Now test the full solution with stdin.
</think>

<tool: python-io>
<code>
def solve():
    n, m = map(int, input().split())
    a = list(map(int, input().split()))
    count = 0
    mod_freq = {0: 1}
    prefix_sum_mod = 0
    for num in a:
        prefix_sum_mod = (prefix_sum_mod + num) % m
        if prefix_sum_mod in mod_freq:
            count += mod_freq[prefix_sum_mod]
        mod_freq[prefix_sum_mod] = mod_freq.get(
            prefix_sum_mod, 0) + 1
    print(count)
solve()
</code>
<stdin>
3 2
4 1 5
</stdin>
</tool>
\end{trajbox}

\begin{trajbox}[Tool Response]
completed. [stdout]3[/stdout]

<budget> Remaining turns: 2, Remaining tokens: 60663 </budget>
\end{trajbox}

\begin{trajbox}[Assistant -- Turn 4 (python-stress): Performance test]
<think>
Solution is correct. Now stress-test at scale to confirm O(n)
performance within the 2s time limit.
</think>

<tool: python-stress>
import random, time

def solve(n, m, arr):
    count = 0
    mod_freq = {0: 1}
    prefix_sum_mod = 0
    for num in arr:
        prefix_sum_mod = (prefix_sum_mod + num) % m
        if prefix_sum_mod in mod_freq:
            count += mod_freq[prefix_sum_mod]
        mod_freq[prefix_sum_mod] = mod_freq.get(
            prefix_sum_mod, 0) + 1
    return count

print("PERFORMANCE PROFILE")
print("=" * 60)
for size in [10**3, 10**4, 10**5]:
    arr = [random.randint(1, 10**9) for _ in range(size)]
    start = time.time()
    solve(size, 10**9, arr)
    elapsed = time.time() - start
    print(f"n={size:>8} | Time: {elapsed:>6.3f}s | PASS")
print("=" * 60)
</tool>
\end{trajbox}

\begin{trajbox}[Tool Response]
completed. [stdout]PERFORMANCE PROFILE
============================================================
n=    1000 | Time:  0.000s | PASS
n=   10000 | Time:  0.004s | PASS
n=  100000 | Time:  0.048s | PASS
============================================================[/stdout]
[performance] Execution Time: 0.277s, Peak Memory: 25.3 MB

<budget> Remaining turns: 1, Remaining tokens: 60169 </budget>
\end{trajbox}

\begin{trajbox}[Assistant -- Final Submission]
<think>
Solution passes all tests and runs well within the 2s time limit
at n=10^5. Submit with sys.stdin.read() optimization.
</think>

```python
def solve():
    import sys
    input = sys.stdin.read
    data = input().split()
    n = int(data[0])
    m = int(data[1])
    a = list(map(int, data[2:2+n]))
    count = 0
    mod_freq = {0: 1}
    prefix_sum_mod = 0
    for num in a:
        prefix_sum_mod = (prefix_sum_mod + num) % m
        if prefix_sum_mod in mod_freq:
            count += mod_freq[prefix_sum_mod]
        mod_freq[prefix_sum_mod] = mod_freq.get(
            prefix_sum_mod, 0) + 1
    print(count)
solve()
```
\end{trajbox}

\end{document}